\documentclass[11pt, a4paper, logo, onecolumn, nonumbering]{deepmind}
\usepackage{kantlipsum, lipsum}
\usepackage{booktabs}
\usepackage{siunitx}
\usepackage{array}
\usepackage{boldline}
\usepackage{xcolor}
\usepackage{wrapfig}
\usepackage{enumitem}
\newcolumntype{?}{!{\vrule width 2pt}}
\usepackage{caption} 
\newcommand{\bnote}[1]{} 

\usepackage{wrapfig}
\usepackage[normalem]{ulem}
\usepackage{hyperref}
\usepackage{amsmath}
\usepackage{graphicx}
\usepackage[export]{adjustbox}
\usepackage{float}

\usepackage{tocloft}
\usepackage{authblk}

\title{Open Problems in Cooperative AI}

\author[1]{\small Allan Dafoe}
\author[2]{\small Edward Hughes}
\author[2]{\small Yoram Bachrach}
\author[2]{\small Tantum Collins}
\author[2]{\small Kevin R. McKee}
\author[2]{\small Joel Z. Leibo}
\author[2, 3]{\small Kate Larson}
\author[2]{\small Thore Graepel}

\affil[1]{\footnotesize Centre for the Governance of AI, Future of Humanity Institute, University of Oxford}
\affil[2]{\footnotesize DeepMind}
\affil[3]{\footnotesize University of Waterloo}
\date{Current draft: December 15 2020. \\First draft: August 2019.}

\begin{document}

\begin{abstract}
Problems of cooperation---in which agents seek ways to jointly improve their welfare---are ubiquitous and important. They can be found at scales ranging from our daily routines---such as driving on highways, scheduling meetings, and working collaboratively---to our global challenges---such as peace, commerce, and pandemic preparedness. Arguably, the success of the human species is rooted in our ability to cooperate. 
Since machines powered by artificial intelligence are playing an ever greater role in our lives, it will be important to equip them with the capabilities necessary to cooperate and to foster cooperation.\\ 
   
We see an opportunity for the field of artificial intelligence to explicitly focus effort on this class of problems, which we term \textit{Cooperative AI}. The objective of this research would be to study the many aspects of the problems of cooperation and to innovate in AI to contribute to solving these problems. Central goals include building machine agents with the capabilities needed for cooperation, building tools to foster cooperation in populations of (machine and/or human) agents, and otherwise conducting AI research for insight relevant to problems of cooperation.
This research integrates ongoing work on multi-agent systems, game theory and social choice, human-machine interaction and alignment, natural-language processing, and the construction of social tools and platforms. However, Cooperative AI is not the union of these existing areas, but rather an independent bet about the productivity of specific kinds of conversations that involve these and other areas. We see opportunity to more explicitly focus on the problem of cooperation, to construct unified theory and vocabulary, and to build bridges with adjacent communities working on cooperation, including in the natural, social, and behavioural sciences. \\ 

Conversations on Cooperative AI can be organized in part in terms of the dimensions of cooperative opportunities. These include the strategic context, the extent of common versus conflicting interest, the kinds of entities who are cooperating, and whether researchers take the perspective of an individual or of a social planner.
Conversations can also be focused on key capabilities necessary for cooperation, such as understanding, communication, cooperative commitments, and cooperative institutions.  Finally, research should study the potential downsides of cooperative capabilities---such as exclusion and coercion---and how to channel cooperative capabilities to best improve human welfare. This research would connect AI research to the broader scientific enterprise 
studying the problem of cooperation, and to the broader social effort to solve cooperation problems. This conversation will continue at: \href{https://www.cooperativeAI.com}{www.cooperativeAI.com}

\end{abstract}
\maketitle 

\vspace{15mm}
\noindent Prepared for the \textit{NeurIPS 2020 Cooperative AI Workshop}\\
Current draft: December 15 2020. \\First draft: August 2019.

\newpage
\renewcommand\cftsecafterpnum{\vskip12pt}
\tableofcontents

\newpage
\section{Introduction}

Problems of cooperation---in which agents have opportunities to improve their joint welfare but are not easily able to do so---are ubiquitous and important. 
They can be found at all scales ranging from our daily routines---such as driving on highways, scheduling meetings, and working collaboratively---to our global challenges---such as peace, commerce, and pandemic preparedness. Human civilization and the success of the human species depends on our ability to cooperate. 

Advances in artificial intelligence pose increasing opportunity for AI research to promote human cooperation. AI research enables new \textit{tools} for facilitating cooperation, such as language translation, human-computer interfaces, social and political platforms, reputation systems, algorithms for group decision-making, and other deployed social mechanisms; it will be valuable to have explicit attention to what tools are needed, and what pitfalls should be avoided, to best promote cooperation.
AI \textit{agents} will play an increasingly important role in our lives, such as in self-driving vehicles, %
customer assistants, 
and personal assistants; it is important to equip AI agents with the requisite competencies to cooperate with others (humans and machines). Beyond the creation of machine tools and agents, the rapid growth of AI research presents other opportunities for advancing cooperation, such as from \textit{research insights} into social choice theory or the modeling of social systems.

The field of artificial intelligence has an opportunity to increase its attention to this class of problems, which we refer to collectively as problems in \textbf{Cooperative AI}. The goal would be to study problems of cooperation through the lens of artificial intelligence and to innovate in artificial intelligence to help solve these problems. Whereas much AI research to date has focused on improving the individual intelligence of agents and algorithms,
the time is right to also focus on improving \textbf{social intelligence}: the ability of groups to effectively cooperate to solve the problems they face. 

AI research relevant to cooperation has been taking place in many different areas, including in multi-agent systems, game theory and social choice, human-machine interaction and alignment, natural-language processing, and the construction of social tools and platforms. Our recommendation is not merely to construct an umbrella term for these areas, but rather to encourage focused research conversations, spanning these areas, focused on cooperation. We see opportunity to construct more unified theory and vocabulary related to problems of cooperation. Having done so, we think AI research will be in a better position to learn from and contribute to the broader research program on cooperation spanning the natural sciences, social sciences, and behavioural sciences. 

Our overview comes from the perspective of authors who are especially impressed by and immersed in the achievements of deep learning \cite{sejnowski2020unreasonable} and reinforcement learning \cite{sutton2020reinforcement}. From that perspective, 
it will be important to develop training environments, tasks, and domains that can provide suitable feedback for learning and in which cooperative capabilities are crucial to success, non-trivial, learnable, and measurable. Much research in multi-agent systems and human-machine interaction will focus on cooperation problems in contexts of pure common interest. This will need to be complemented by research in mixed-motives contexts, where problems of trust, deception, and commitment arise. Machine agents will often act on behalf of particular humans and will impact other humans; as a consequence, this research will need to consider how machines can adequately understand human preferences, and how best to integrate human norms and ethics into cooperative arrangements. Researchers building social tools and platforms will have other perspectives on how best to make progress on problems of cooperation, including being especially informed by real-world complexities. Areas such as trusted hardware design and cryptography may be relevant for addressing commitment problems and cryptography. Other aspects of the problem will benefit from expertise from other sciences, such as political science, law, economics, sociology, psychology, and neuroscience. We anticipate much value in explicitly connecting AI research to the broader scientific enterprise studying the problem of cooperation and to the broader effort to solve societal cooperation problems.

We recommend that ``Cooperative AI'' be given a technically precise, problem-defined scope; otherwise, there is a risk that it acquires an amorphous cloud of meaning, incorporating adjacent (clusters of) concepts such as aligned AI, trustworthy AI, and beneficial AI.  
Cooperative AI, as scoped here, refers to \textit{AI research trying to help individuals, humans and machines, to find ways to improve their joint welfare}. For any given situation and set of agents, this problem is relatively well defined and unambiguous. The \hyperref[section:scope]{Scope} section elaborates on the relationship to adjacent areas.

Conversations on Cooperative AI can be organized in part in terms of the dimensions of cooperative opportunities. These include the strategic context, the extent of common versus conflicting interest, the kinds of entities who are cooperating, and whether the researchers are focusing on the cooperative competence of individuals or taking the perspective of a social planner. Conversations can also be focused on key capabilities necessary for cooperation, including:
\vspace{-4mm}
\begin{enumerate}
    \item \textbf{Understanding} of other agents, their beliefs, incentives, and capabilities.
    \item \textbf{Communication} between agents, including building a shared language and overcoming mistrust and deception.
    \item Constructing cooperative \textbf{commitments}, so as to overcome incentives to renege on a cooperative arrangement.
    \item \textbf{Institutions}, which can provide social structure to promote cooperation, be they decentralized and informal, such as norms, or centralized and formal, such as legal systems. 
\end{enumerate}

Just as any area of research can have downsides, so is it prudent to investigate the potential downsides of research on Cooperative AI. Cooperative competence can be used to exclude others, some cooperative capabilities are closely related to coercive capabilities, and learning cooperative competence can be hard to disentangle from coercion and competition. An important aspect of this research, then, will be investigating potential downsides and studying how best to anticipate and mitigate them.

The remainder of the paper is structured as follows. 
We offer more motivation in the section \hyperref[section:why]{Why Cooperative AI?} 
We then discuss several important dimensions of \hyperref[section:opportunities]{Cooperative Opportunities}. 
The bulk of our discussion is contained in the \hyperref[section:capabilities]{Cooperative Capabilities} section, which we organize in terms of 
\hyperref[section:understanding]{Understanding}, \hyperref[section:communication]{Communication}, \hyperref[section:commitment]{Commitment}, and \hyperref[section:institutions]{Institutions}. 
We then reflect on \hyperref[section:downsides]{The Potential Downsides of Cooperative AI}
and how to mitigate them. Finally, we \hyperref[section:conclusion]{conclude}.

\section{Why Cooperative AI?}
\label{section:why}

\phantomsection
\subsection{Vignettes: Self-Driving Vehicles and COVID-19}

To ground our discussion, we introduce here two vignettes, one based on the near-term cooperation problems facing self-driving vehicles, the other about the global challenge of pandemic preparedness. 

Self-driving vehicles confront a broad portfolio of cooperation problems with respect to other drivers (human and AI). There are opportunities for joint welfare gains between drivers (or the principals in whose interests they are driving).

In order to predict the behaviour of other drivers, an AI, and others on the road, would benefit from it \textbf{understanding} the goals, beliefs, and capabilities of those other drivers. The AI, and others, would benefit from it understanding the (local) conventions of driving, such as what side of the road to drive on. It, and others, benefit from it accurately modeling the beliefs of other agents, such as whether the fast-walking pedestrian looking at their phone is aware of the car. It, and others, benefit from it understanding the intentions of other agents, such as that of a driver attempting to urgently cross several lanes of fast-flowing traffic.

An AI, and others, would benefit from improved \textbf{communication} skill and infrastructure. Will the AI understand the wave of a police officer, indicating that the officer wants the AI to go through an intersection? Can it express to other vehicles that the reason it is idling in a narrow parking lot is to wait for a car up ahead to pull out? Can it accurately, precisely, and urgently express a sudden observation of dangerous road debris to (AI) drivers behind it? Can it communicate sufficiently with another brand of driver AI to achieve the subtle coordination required to safely convoy in close proximity?

Cooperative gains are available to those able to construct credible \textbf{commitments}.
Car \textbf{A}, in busy traffic, may wish to cross several lanes; car \textbf{B} may be willing to make space for \textbf{A} to do so only if \textbf{A} commits to moving along and not staying in \textbf{B}'s lane. Or, suppose a driver is waiting for another car to pull out of a parking space, blocking other cars looking for parking spots; in principle, the driver would be willing to allow those other cars to pass, but not if one of them is going to ``defect'' by taking the newly opened parking space. Can those other drivers credibly commit to not doing so? 

Populations of drivers could be made better off by new \textbf{institutions}, which AI research could help build. What is the optimal congestion pricing to maximize human welfare, and can this be assessed and processed through AI-enabled innovations? Are there joint policies for populations of drivers which are Pareto improving to the typical equilibria from uncoordinated behaviour? If so, could a navigation app achieve this while maintaining incentive compatibility to participate in the mechanism? Can the valuable information from the video feeds of the many smart vehicles be optimally distributed, with fair pricing, safeguards for privacy, and incentive compatibility?

While our first example was meant to be grounded in soon-to-be-upon-us technical possibilities, our second is meant to illustrate the global stakes of making progress in solving cooperation problems. There are multiple problems of global cooperation with stakes in the trillions of dollars or millions of lives, including those of disarmament, avoiding nuclear war, climate change, global commerce, and pandemic preparedness. Given its timeliness, we focus on the last one. While these global problems will not immediately be the focal problems of Cooperative AI, they illustrate the magnitude of benefits that could ultimately come from substantial advances in global cooperative competence. 

\bnote{Role of AI in this example is not spelled out. We can't fix that now, but for the future would be better to have an example with a more compelling role of AI.}
The death toll from COVID-19 is in the millions, and the economic damage is estimated in the tens of trillions of dollars. Future pandemic diseases could cause similar, or greater, devastation. And yet, despite these high stakes, the world struggles to prepare adequately, in part due to cooperation problems. The gains are great from more investment in generalized vaccine development, disease surveillance and data sharing, harmonization of policy responses so as to avoid unnecessary breakdowns in supply chains, pooled reserves of supplies to support those who most need them, epidemiological research during the outbreak, and building coordinating institutions to help achieve these common interests. 

But achieving these gains is not trivial. They may require that decision makers \textbf{understand} each other sufficiently well that they can agree to a fair (enough) division of investment for a problem in which expected harm and ability to pay are unevenly distributed and hard to estimate. They may require decision makers to reliably \textbf{communicate} on sensitive issues like the existence of a disease outbreak or the state of one's medical system. They may require overcoming the \textbf{commitment} problem arising from the great pressure for countries to renege on certain agreements in a crisis, such as on  sharing of supplies and not interfering in supply chains. Lastly, building global \textbf{institutions} poses difficult problems of institutional design, and the management of difficult political realities, conflicts of interest, demands for legitimacy, and requirements for technical competence. 

As these vignettes illustrate, problems of cooperation are ubiquitous, important, and diverse, but they also share fundamental commonalities.  Problems of cooperation span scale: from inter-personal, to inter-organizational, to inter-state. They can involve two, several, or millions of agents; they exist for small stakes and global stakes; they arise amongst humans, machines, and organizations; they arise in domains with more or less well-defined interests, norms, and institutions. 

Cooperative competence requires distinct kinds of social intelligence and skills, which were critical for the success of humans.
Without a cultural inheritance of tools and skills, and a community to collaborate with, a single human cannot achieve much.  Rather, to create our modern technological and cultural wonders required a community of collaborating humans, ever growing in scale, who passed their knowledge down through the generations. Furthermore, this capacity evolved beyond a narrow kind of cooperative intelligence, capable of only solving a limited class of cooperation problems and brittle to changes, to an increasingly general cooperative intelligence capable of solving dynamic and complex problems, including interpersonal disputes, cross-border pollution, and global arms control. Further improvements in humanity's general cooperative intelligence may be critical for solving our increasingly complex global problems.
As AI systems are deployed throughout the economy, it becomes important that they be adept at participating cooperatively in our shared global civilization---which is composed of humans, organizations, and, increasingly, machine agents.

The problem of cooperation is a fundamental scientific problem for many fields, from biochemistry, to evolutionary biology, to the social sciences.
Biologists regard the history of life as the progressive formation of ever larger cooperative structures \cite{smith1997major}, from coalitions of cooperating genes \cite{smith1993origin}, to multi-cellularity requiring restraint on single-cell egoism \cite{buss1987evolution, frank2004problems}, to the emergence of complex societies featuring division of labour in the lineages of ants, termites, and primates \cite{robinson1992regulation, mueller2005evolution}.
In 2005, \textit{Science Magazine} judged the problem of ``\href{https://science.sciencemag.org/content/309/5731/93.full}{How Did Cooperative Behavior Evolve?}'' to be one of the \href{https://www.sciencemag.org/site/feature/misc/webfeat/125th/}{top 25 questions}  ``facing science over the next 25 years.'' 

The problem of cooperation is at least as important to social scientists. Economists are often
interested in when people achieve, or fail to achieve, welfare-enhancing arrangements. 
Political scientists study how people make collective decisions: the very mechanisms by which groups of people cooperate, or fail to do so. 
Within international relations and comparative politics, some of the most central and rich questions concern the causes of costly conflict, including civil war and interstate war. Prominent in sociology is the study of social order: how it emerges, persists, and evolves. 
Explicitly framing relevant research in AI as addressing this fundamental problem will help with the consilience of science, allowing relevant insights across fields to more readily flow to each other and helping to show these other fields the increasing relevance of AI research to understanding cooperation. A more explicit connection across these fields will help researchers adopt a common vocabulary and learn from their respective advances.

When considering how to build maximally beneficial AI, 
many researchers emphasize the importance of 
certain directed strands of AI research, such as on safety \cite{amodei2016concrete, russell2019human}, fairness \cite{hardt2016equality}, %
interpretability \cite{olah2018building}, and robustness \cite{russell2015research, qin2019adversarial}.
Each of these points AI research towards achieving AI with a certain set of attributes, which are thought to be on net socially beneficial, and they plausibly make for a coherent research program. We recommend that cooperation and Cooperative AI be added to this list. 

These and other areas of AI are defined by what they aspire to achieve by their goals. 
Even the field of ``artificial intelligence'' itself is defined through its aspiration to build a kind of machine intelligence that does not yet exist. 
Aspirational research programs have the advantage of prominently communicating what the point of the research is and reminding researchers of some of its long-term goals.
Consider, by contrast, labeling a cluster of this kind of research
as being on ``multi-agent AI'', ``game theory'', ``strategic interaction'': these don't communicate the social goal of the research and the pro-social bet motivating the field. %
``Cooperative AI'' more clearly points to the social purpose served by pursuing these connected clusters of research.

\subsection{Prior Work, Machine Learning, and Timeliness} 
\label{l_sect_relation_to_mas}

Cooperative AI will draw together research spanning the field of artificial intelligence, as well as many other disciplines in the natural and social sciences. Accordingly, prior work relevant to Cooperative AI is vast and will be most meaningfully summarized for particular sub-clusters of work. Reflecting the background of the majority of the authors of this paper, we briefly review prior work here with a heavy bias towards multi-agent research. 

Looking back decades, multi-agent systems research  (and before that, the distributed AI research community)  has long been interested in the interactions of intelligent agents and the arising collective behaviour~\cite{weiss1999multiagent,ferber1999multi,jennings2000agent,padgham2005developing, dresner2004multiagent,shoham2008multiagent,wooldridge2009introduction,bordini2009multi,jennings1998roadmap}. 
Early research from the 1970s and 1980s employed 
computational models for simulating such systems by combining elements of game theory, Monte Carlo simulation, evolutionary programming, and theories of emergence and complex systems~\cite{schelling1971dynamic,axelrod1981evolution,reynolds1987flocks,minsky1988society}. During the 1990s and early 2000s, the focus of research broadened from technologies for just \emph{simulating real-world environments} to \emph{architectures and methodologies for solving} inherently distributed problems such as markets, trading and auctions~\cite{sandholm1996limitations,park1999adaptive,greenwald2001autonomous,nisan2001algorithmic,conitzer2002complexity,roughgarden2010algorithmic}, security and disaster response~\cite{schurr2005future,massaguer2006multi,tambe2011security,fang2015security}, resource management~\cite{pipattanasomporn2009multi,chevaleyre2005issues}, automated argumentation, and negotiation~\cite{rosenschein1994rules,sandholm1995issues,jennings2001automated,mcburney2002desiderata,kakas2006adaptive}. More recently, there has been significant research on methods through which agents learn to communicate, collaborate, and interact with one another~\cite{claus1998dynamics,stone2000multiagent,wagner2003progress,panait2005cooperative,zhang2010multi,lazaridou2016multi,havrylov2017emergence,al2017continuous}.

Kraus~\cite{kraus1997negotiation} argued for the importance of studying mixed-motive cooperation problems and not just pure common-interest problems. She advocated an interdisciplinary approach and organized multi-agent cooperative work along several dimensions, which will recur in our paper. These are (in our words): the degree of common interest; the opportunity for building institutions; the number of agents; the kinds of agents (machines, humans); and the costs and capabilities of both communication and computation.
Panait and Luke \cite{panait2005cooperative} review work on machine learning in multi-agent cooperation, 
though mostly limited to settings of pure common interest. 
They discuss challenges from different learning dynamics and illustrate them with simple games and real-world problems. They emphasize the problem of ``teammate modeling'', which maps to the concepts we discuss in \hyperref[section:understanding]{Understanding} and \hyperref[section:communication]{Communication}. 

In the past two decades, research in multi-agent systems has exploded to cover diverse areas including %
institutions and norms~\cite{robertson2004multi,boella2006introduction,argente2006multi}; human agent interaction~\cite{lewis1998designing,guzzoni2007modeling,bradshaw2011human,sierhuis2003human}; knowledge representation, reasoning, and planning~\cite{georgeff1988communication,van2002tractable,van2008handbook}; multi-agent adaptation and learning~\cite{weiss1995adaptation,georgeff1998belief,alonso2003adaptive,
bowling2005convergence,shoham2007if,busoniu2008comprehensive}; social choice and joint decision-making~\cite{pitt2006voting,chevaleyre2007short,brandt2012computational}; strategic and economic interactions~\cite{rosenschein1994rules,nisan2001algorithmic,kraus2001strategic,parsons2002game,wellman20032001,wellman2015economic};  simulations of agent societies~\cite{minar1996swarm,artikis2001formal,neville2008presage}; and multi-agent robotics~\cite{kitano1997robocup,burgard2000collaborative,arai2002advances,gerkey2004formal}.

Despite the frameworks and methodologies that have been developed over the past decades and the great potential for multi-agent technologies, progress has been slow in some areas due to the inherent complexity of the problems~\cite{jennings1998roadmap,weiss1999multiagent,singh2007multiagent,shoham2008multiagent,wooldridge2009introduction}. In particular, it has long been noted that open, heterogeneous, and scalable multi-agent systems require learning agents, including agents who learn to adapt and cooperate with others~\cite{luck2003agent}.
In light of the recent  progress in our ability to analyze and learn from data in various forms---including image processing, recognition, and generation~\cite{krizhevsky2012imagenet,goodfellow2014generative}; voice and natural-language processing~\cite{collobert2011natural,oord2016wavenet,pennington2014glove}; and reinforcement learning~\cite{mnih2015human,silver2016mastering,berner2019dota}---it is time to reinvigorate research on Cooperative AI.\footnote{In very recent work, a statement close in spirit to this paper, produced by some of the same authors, appeared in a September 2020 announcement for the 2020 NeurIPS Workshop (reproduced in \hyperref[section:appendixA]{Appendix A}) \cite{cooperativeAIcom}. Another statement invoking similar motivation to the NeurIPS workshop and this manuscript, though focusing mostly on human-machine cooperation, was a November 2020 CCC Quadrennial Paper on ``Artificial Intelligence and Cooperation'' \cite{AIandCooperation}. A recent research agenda from the AI safety community in part emphasizes problems of cooperation \cite{critch2020ai}.}

There are other reasons why the timing is opportune. Within the domain of game-based reinforcement learning, the past years have seen tremendous progress on two-player zero-sum games, such as chess \cite{silver2018general, lai2015giraffe}, go \cite{silver2018general}, StarCraft II \cite{vinyals2019grandmaster}, and (two player) poker \cite{brownbakhtin2020}.\footnote{There has been related progress on two-team zero-sum games, such as Capture the Flag \cite{jaderberg2019human}, Dota II \cite{berner2019dota}, hide-and-seek \cite{baker2019emergent}, and Honor of Kings \cite{ye2020towards}. Within each team is a setting of pure common interest.}
Two-player zero-sum games were a productive domain for early multi-agent research as they are especially tractable: the minimax solution coincides with the Nash equilibrium and can be computed in polynomial time through a linear program, their solutions are interchangeable, and they have worst-case guarantees \cite{von2007theory}.

However, two-player zero-sum games provide no opportunity for the agents to learn how to cooperate, and it is undesirable for research to be overly focused on domains that are inherently rivalrous. 
The field of multi-agent reinforcement learning seems to be naturally reorienting towards games of pure common interest, such as Hanabi \cite{bard2020hanabi} and Overcooked \cite{carroll2019utility}, as well as team games \cite{jaderberg2019human}. There is also growth in the study of  mixed-motives settings \cite{baker2020emergent}, such as alliance politics  \cite{hughes2020learning, anthony2020learning, paquette2019press} and tax policy \cite{zheng2020ai}, which will be critical since some strategic dynamics only arise in these settings, such as issues of trust and commitment problems. 

Finally, all else equal, earlier is better in cooperation research, since much of actual cooperation depends on historically established vocabularies, norms, precedents,  protocols, and institutions: as more and more AI systems are being deployed throughout society, we do not want to unintentionally lock in sub-optimal equilibria. We want our deployed AIs to be forward compatible with future advances in cooperation.

\section{Cooperative Opportunities}
\label{section:environments}
\label{section:opportunities}

In this section, we will discuss some of the diversity in cooperative opportunities, which are situations in which agents may be able to achieve joint gains or avoid joint losses. Our discussion will consider four major dimensions that structure the character of cooperative opportunities and the associated research. (1) The degree of common versus conflicting interest between agents. (2) The kinds of agents attempting to cooperate, such as machines, humans, or organizations. (3) The perspective taken on the cooperation problem: either that of an individual trying to cooperate with others or that of a social planner facilitating the cooperative interactions of a population. (4) The scope of Cooperative AI research and, specifically, how it should relate to related fields. 

\begin{figure}[H]
\hspace*{-1.2cm}\begin{tabular}{cc}
 \includegraphics[width=90mm]{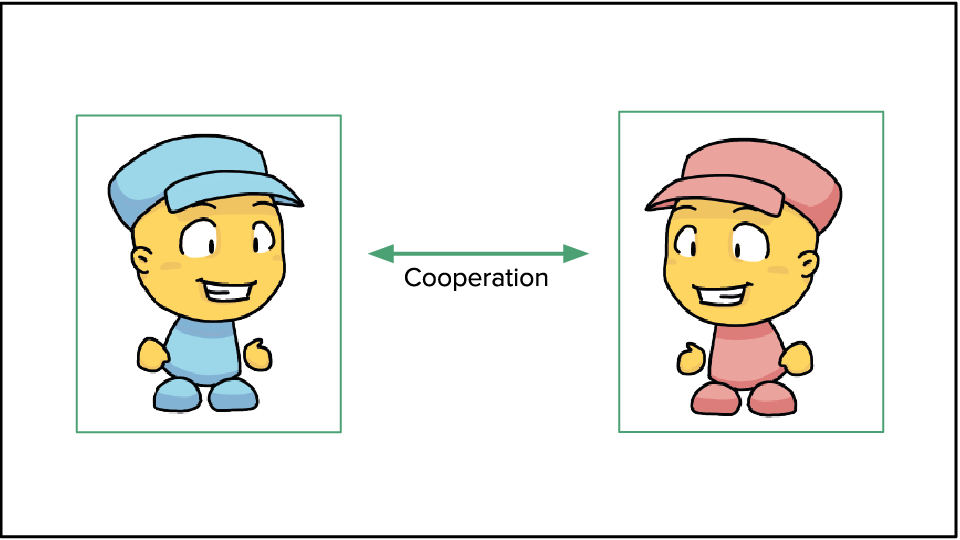} &   
 \includegraphics[width=90mm]{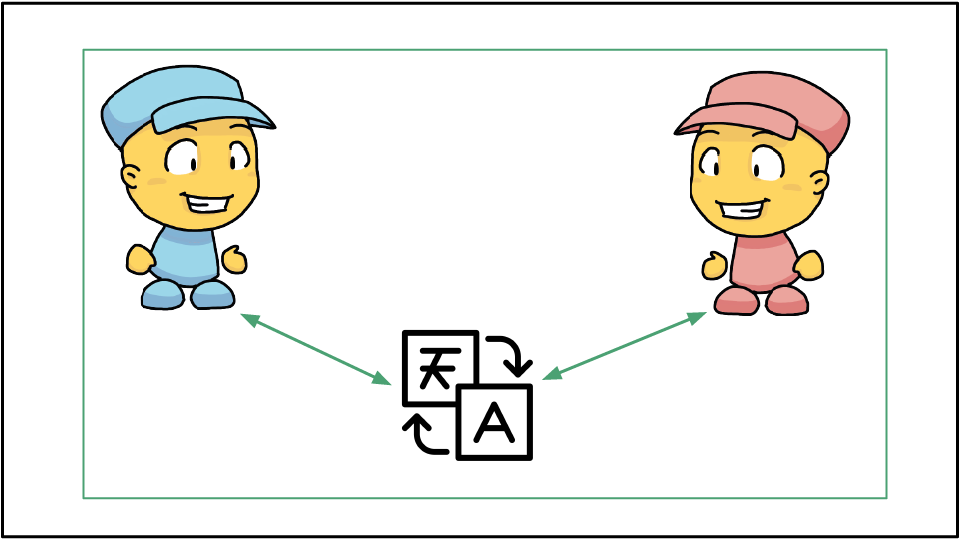} \\
 \textbf{A}: Human-Human Cooperation & 
  \textbf{B}: Cooperative Tools \\[1.5pt] \\
 \includegraphics[width=90mm]{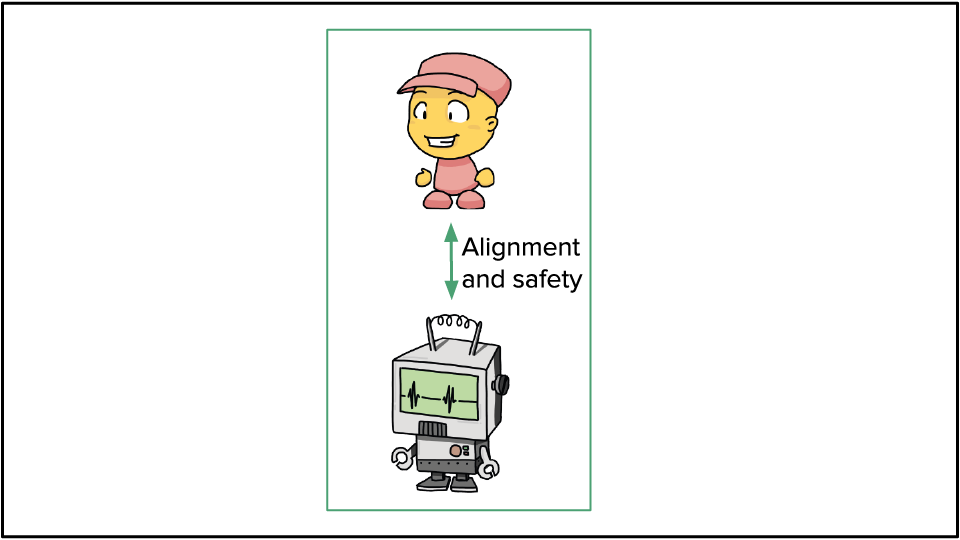} &   
 \includegraphics[width=90mm]{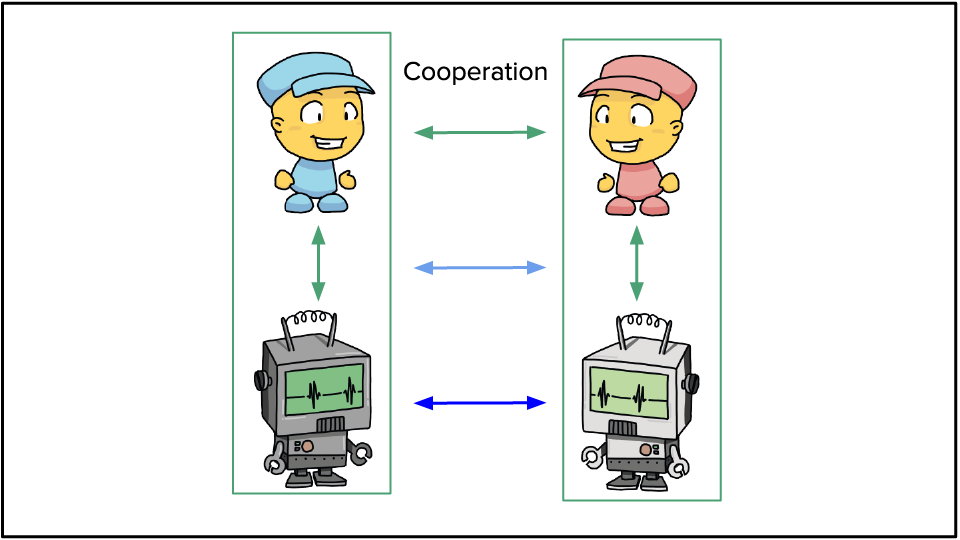} \\
  \textbf{C}: Alignment and Safety  & 
 \textbf{D}:  \{Human-AI\}-\{Human-AI\} Cooperation\\[1.5pt]  \\
  \includegraphics[width=90mm]{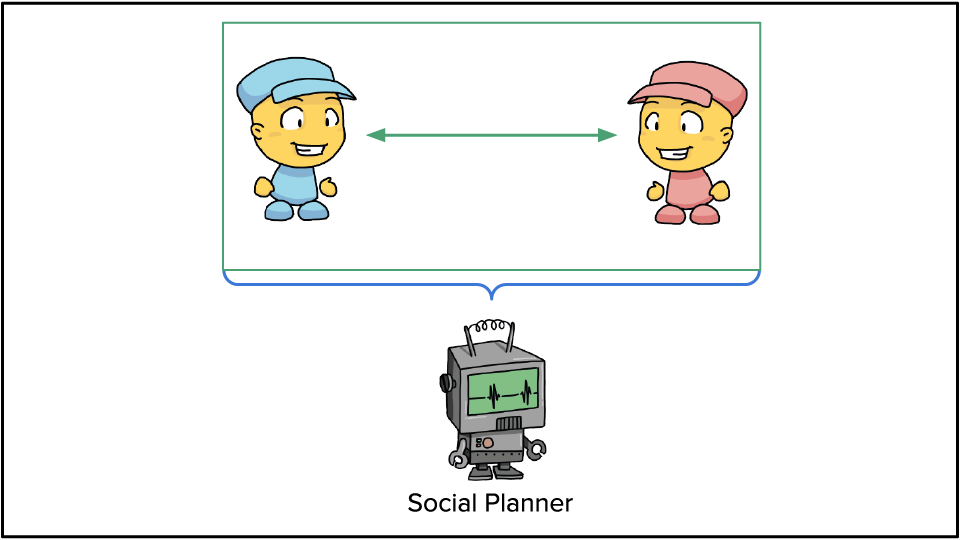} &  
  \includegraphics[width=90mm]{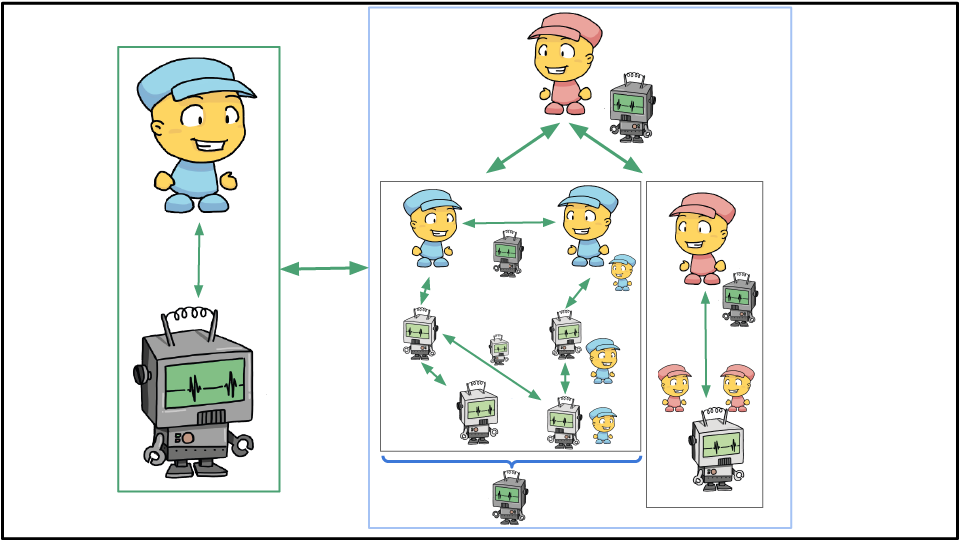} \\
  \textbf{E}: The Planner Perspective 
  & \textbf{F}: Organizations and Society \\[1.5pt]
\end{tabular}
\caption{Cooperation comes in many flavors. 
\textbf{A}: A prototypical cooperation problem between two human principals. 
\textbf{B}: AI will enable new \textit{tools} for promoting cooperation, such as language translation. %
\textbf{C}: Especially capable and autonomous AI may be better conceptualized as an agent, such as an email assistant capable of replying to many emails as well as a human assistant. Human principals need their AI agents to be safe and aligned. This relationship can be conceptualized as a cooperation game. 
The vertical dimension depicts the normative priority %
of the top agent, as in a principal-agent relationship. When the agent is aligned, it is a relationship of pure common interest. 
\textbf{D}: Combining these, the future will involve cooperative opportunities between human-AI teams. %
Advances in AI will enable the nexus of cooperation to move "down" to the AI-AI dyad (increasingly blue arrows), such as with coordination between Level V self-driving cars \cite{sae2018taxonomy}. 
\textbf{E}: AI research can take on the ``planner perspective''. Rather than focus on building Cooperative AI aids or agents for individuals, this perspective seeks to improve social \textit{infrastructure} (e.g., social media) or improve \textit{policy} to better cultivate cooperation within a population. %
\textbf{F}: The structure of interactions can of course be much more complicated, including involving organizations with complex internal structure and nested cooperative opportunities. (Thanks to Marta Garnelo for illustrations.)}
\label{figure:coopAIpanels}
\end{figure}

\subsection{Common and Conflicting Interests}

Decades of social science research have found that the dynamics of multi-agent interaction are fundamentally shaped by the extent of alignment between agents' payoffs \cite{komorita1995interpersonal, rapoport1966taxonomy, robinson2005topology, schelling1980strategy}. 

\begin{enumerate}
    \item At one edge of the space are games of \textbf{pure common interest}, in which any increase in one agent's payoffs always corresponds with an increase in the payoffs of others.
    \item In the broad middle are games with \textbf{mixed motives}, in which agent interests are, to varying extents, sometimes aligned and sometimes in conflict.
    \item At the other edge are games of \textbf{pure conflicting interest}, in which an increase in one agent's payoff is always associated with a decrease in the payoff of others.\footnote{Though note that with more than two players, it is mathematically impossible to have entirely negatively associated payoffs between all players; there must at least be many indifference relations. It is possible to construct a game with three or more players which would effectively reduce to a series of dyadic pure conflicting-interest games. However, this is only possible so long as every player has no option to intervene in the ``dyadic games'' of the others; if there was such an intervention, then there would be some common interest.}
\end{enumerate}

\begin{figure}[H]
    \includegraphics[width=\textwidth]{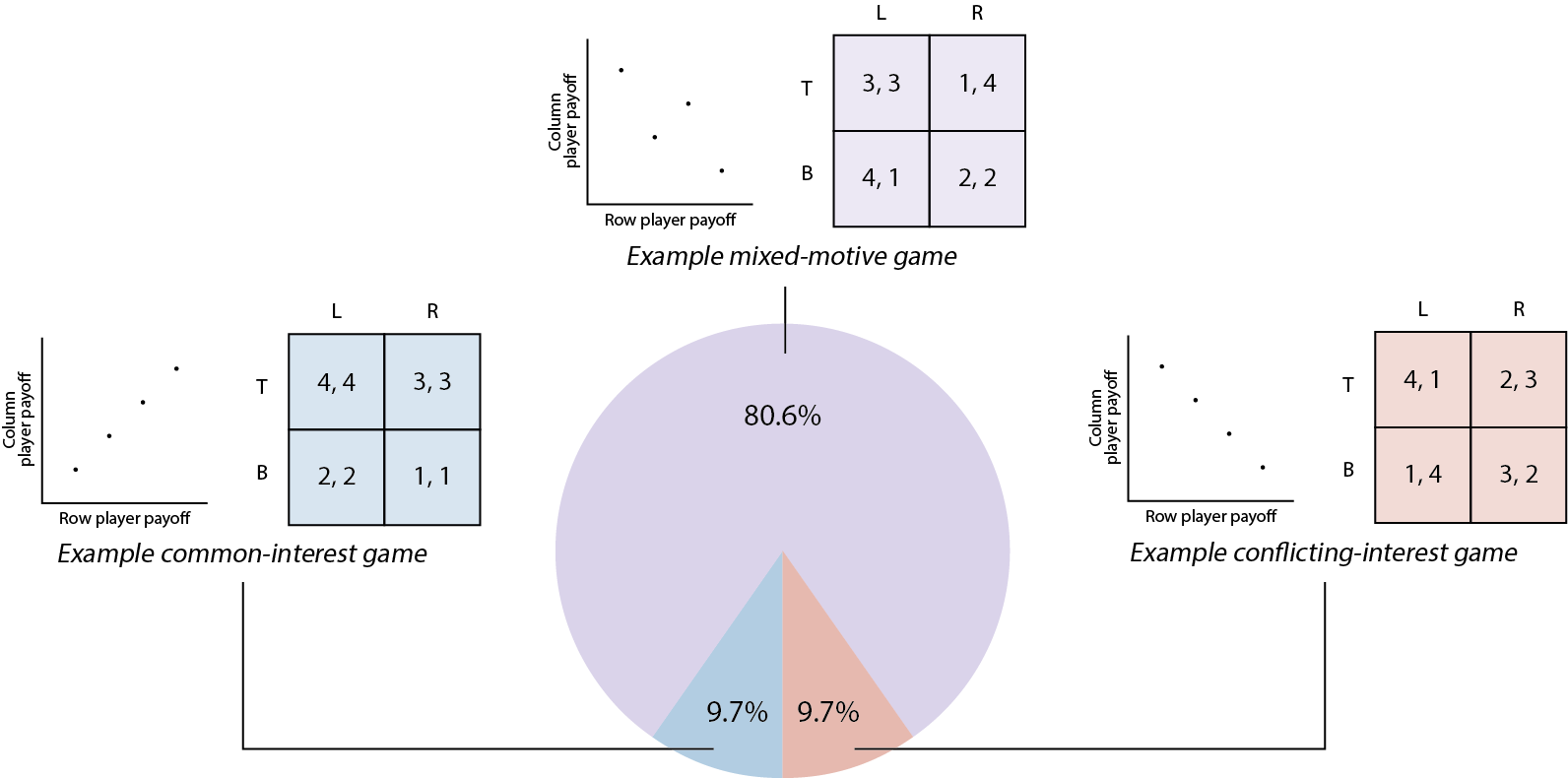}
    \caption{A simple class of multi-agent situation is a game with two players, each of which can adopt one of two possible pure strategies \cite{robinson2005topology}. By converting each player's payoffs to ranked preferences over outcomes---from the most-preferred to least-preferred outcome---we see that there are 144 distinct games. Even in this simple class of two-player games, there exists some common interest in the overwhelming majority of situations.    \label{fig:game_type_distribution} \\}
\end{figure}   
Opportunities for cooperation exist, at least in principle, in situations of pure common interest and mixed motives. It is only in situations of pure conflicting interest where cooperation is impossible. The ubiquity of cooperative opportunities can be seen by considering the small size of the space of pure conflicting-interest games. First, they are almost entirely confined to two-player games, since the introduction of a third player will typically offer at least one dyad an opportunity to cooperate, if only against the third player. Even considering only two-player games, most possible arrangements of payoffs will not be consistently inversely related. To formalize this, Figure \ref{fig:game_type_distribution} shows that in the taxonomy popularized by Robinson and Goforth \cite{robinson2005topology}, the vast majority of games are either pure common interest or mixed motive. Finally, even within the subset of games with purely conflicting interests, if the underlying utilities are not perfectly negatively correlated, then the introduction of a costly action that benefits another player (transfers utility) can introduce common interest.   
   
Thus, situations of purely conflicting interest are rare in the space of strategic games. We believe such situations are also relatively rare in the real world. However, machine learning and reinforcement learning research has focused heavily on conflicting-interest cases---and particularly on two-player, zero-sum environments. Many of the most renowned achievements of multi-agent research, for instance, have focused on pure-conflict games such as backgammon \cite{tesauro1994td}, chess \cite{campbell2002deep}, go \cite{silver2016mastering}, (two player) poker \cite{moravvcik2017deepstack, brown2019superhuman}, and StarCraft \cite{vinyals2019grandmaster}. 

Evidence of this weighting towards games of pure conflicting interest extends beyond these prominent studies. To evaluate more systematically how different fields attend to games with or without cooperative opportunities, we analyzed citation patterns. We first used keywords (such as ``chess'', ``social dilemma'', and ``coalitional'') to produce a rough proxy for whether a multi-agent paper was studying a situation of ``common interest'', ``mixed motives'', or ``conflicting interest''. We then examined the proportion of citations from papers in economics, machine learning, reinforcement learning, and multi-agent research which were directed to these different categories of papers. We found that papers in machine learning and reinforcement learning were much more likely to cite work on ``conflicting-interest'' situations (around 10--15\% of their outgoing multi-agent citations) than were economics papers or other multi-agent papers (only 2\% and 4\% of their outgoing multi-agent citations, respectively). This weighting towards conflicting-interest games suggests that there are underexplored opportunities to study mixed-motive and common-interest environments in reinforcement learning.

Work on settings of purely conflicting interests can, of course, provide useful insights also relevant to work on Cooperative AI. For example, research on poker-playing AI systems led to the development of counterfactual regret minimization \cite{zinkevich2008regret}, which has subsequently been leveraged to improve algorithmic performance in mixed-motive settings \cite{serrino2019finding}. As mentioned above, two-player zero-sum games were a productive domain for early multi-agent research as they are especially tractable: the minimax solution coincides with the Nash equilibrium and can be computed in polynomial time through a linear program, their solutions are interchangeable, and they have worst-case guarantees \cite{von2007theory}. This tractability may explain why such games have received significant research attention, despite being relatively rare in the real world and in the space of possible games. Going forward, we think the study of pure common-interest and mixed-motives games---games permitting cooperation---will be particularly fruitful.

\subsection{Who is Cooperating: Humans, Machines, Organizations}

Cooperative opportunities are critically affected by the kind of agents and entities involved in interactions. Separate research communities have coalesced around the topics of cooperation within human groups (e.g., \cite{rand2013human, tomasello2009we}), cooperation within communities of artificial agents (e.g., \cite{leibo2017multi, olfati2007consensus}), and cooperation between human and artificial partners (e.g., \cite{crandall2018cooperating, norman1994might}). Problems of cooperation between states (e.g., \cite{frieden2015world}),  firms (e.g., \cite{parkhe1993strategic}), and other entities are also prominent themes in the social and policy sciences. Within each of these categories---humans, machines, organizations---there is substantial variation in preferences and cognition. Cooperation amongst children, for example, differs from cooperation amongst adults \cite{warneken2018children}; cooperation between people in the Eastern United States diverges from cooperation between individuals in Peruvian communities \cite{henrich2004foundations}.

Though a deep body of research has studied the mechanisms underlying cooperation within human groups, settings with non-human agents are likely to possess markedly different dynamics. Machine-machine cooperation, for instance, may involve more exotic settings, higher bandwidth communication, greater strategic responsiveness, closer adherence to principles of rational decision-making, hard-to-interpret emergent language \cite{kottur2017natural}, sensitivity to narrow strategy profiles \cite{carroll2019utility}, and greater strategic complexity. Machine learning algorithms have a well-documented tendency to find unexpected or undesired solutions to problems they are posed, a critical specification challenge for AI researchers \cite{krakovna2020specification}. Cooperative opportunities involving solely artificial agents may thus generate solutions, both good and bad, that diverge from those observed in human communities.

The landscape further shifts when we consider hybrid groups containing both humans and machines. Machines developed to interact with human partners are more likely to succeed if their design incorporates processes and factors related to human behaviour---including cognitive heuristics \cite{tversky1974judgment}, social cognitive mechanisms \cite{rilling2011neuroscience}, cultural context \cite{takeuchi2000cultural}, legible motion \cite{dragan2013legibility}, and personal preferences and expectations \cite{norman1994might}. For instance, self-driving cars that performed well when interacting with other autonomous vehicles have struggled to adapt to the assertiveness of human drivers, even when this manifests in subtle ways such as inching forward at intersections \cite{schwarting2019social, richtel2015google}. 

A different set of challenges arises for AI research focused primarily on cooperation among humans (e.g., \cite{zheng2020ai}). Initial studies of such contexts suggest that the natural dynamics of human groups can be substantially altered by the presence of an artificial agent \cite{shirado2017locally, traeger2020vulnerable}. These situations will likely require researchers to adapt a new set of hybrid approaches, drawing heavily from fields including social psychology, sociology, and behavioural economics.

\subsection{The Individual Perspective and the Planner Perspective}

A third distinction relates to whose welfare one is most concerned with in a particular cooperation problem.  The \textit{individual perspective} seeks to achieve the goals of an individual in a cooperative setting, which usually involves improving the individual's cooperative capabilities (as covered in our sections on Understanding, Communication, and Commitment). The \textit{planner perspective} instead seeks to achieve some notion of social welfare for an interacting population, which usually involves intervening on population-level dynamics through policies, institutions, and centralized mechanisms (as covered in our section on Institutions).\footnote{While these different goals tend to lead to different focuses (on individual capabilities vs institutions), they need not. It is conceivable that the individual perspective leads to the problem of institution design or that the planner perspective leads to the problem of improving individual cooperative competence.} The individual perspective tends to be concerned with machine agents, but it could also look at machine aids to humans. The planner perspective tends to look at populations of humans, but it could also look at populations involving machines. 

Which perspective one takes depends in part on the problem one is trying to solve and the opportunities available. Do we have an opportunity to advise or improve the cooperative capabilities of an individual? Do we have an opportunity to influence behaviour-shaping factors, like norms, policies, mediators, or institutions?

To some extent, any cooperative situation benefits from being understood from both these perspectives. A competent social planner should understand the interests (and cooperative capabilities) of the individuals, at the least to know how best to intervene to maximally facilitate cooperation. Similarly, a maximally competent cooperative individual likely needs the ability to think about the population's cooperative opportunity as a whole to identify what group-level changes would best help bring the population (including the individual) to the Pareto frontier.\footnote{Illustrating the value of a cooperative individual being able to ``simulate the social planner'' and take the perspective of the group trying to cooperate, UN Secretary-General Dag Hammarskj\"old famously argued ``that every individual involved in international relations, particularly those working with the UN, should \emph{speak for the world}, rather than from purely national interest.'' \cite{gaudiosi2019negotiating}, emphasis ours.}

These perspectives are associated with differences in method. The individual perspective tends to involve an agent optimizing over its local environment and considering the strategic response of other agents. The planner perspective, especially for large populations, more often involves a study of equilibration and emergence, and thus more resembles a problem in statistical physics (as noted by \cite{kraus1997negotiation}): what nudges and structures would steer these emergent equilibria in desirable directions?

\subsection{Scope}
\label{section:scope}

The field of Cooperative AI, as scoped here, involves AI research which can help contribute to solving problems of cooperation. This may seem overly expansive, as it includes many different kinds of cooperation (machine-machine, human-machine, human-human, and more complex constellations) and many disparate areas (multi-agent AI research, human-machine interaction and alignment, mechanism design, and myriad tools such as language translation and collaborative productivity software). However, research fields and programs should not be thought of in this way, as merely expressing a set relationship to each other. For example, it is not especially meaningful to say that biology is just applied chemistry, and chemistry is applied physics.

Rather, research fields can be thought of as expressing a bet about where productive conversations lie. The bet of a Cooperative AI field is similar to the bet for organizing a Cooperative AI workshop: that there are productive conversations to be had here which are otherwise not happening for want of: (1) an overarching compass aligning the many disparate research threads into solving a large common problem; (2) a unified theory and vocabulary to facilitate the transfer of insights across threads; and (3) a more deliberate construction of conversations which span communities (including non-AI communities).

The problem of cooperation has been worked on in biology, game theory, economics, political science, psychology, and other fields, and there has been much productive exchange across these disciplines. The bet behind Cooperative AI is that there is value in connecting the research in AI, and opportunities with AI, to this broader conversation in a theoretically explicit and sustained way.

We will elaborate this point further. Cooperative AI emerges in part from multi-agent research, as reflected in our review of prior work. However, it is not equivalent to this field. It will emphasize different subproblems. For example, it will point away from zero-sum games and towards social dilemmas. It will also more consciously strive to develop theory which is compatible with adjacent sciences of cooperation, and to be in conversation with those sciences.

Cooperative AI emerges in part from work on human-machine interaction and AI alignment, but it is not equivalent. For example, these areas typically involve a principal-agent relationship, in which the human (principal) has normative priority over the machine (agent). AI alignment researchers are concerned with the problem of \textit{aligning} the machine agent so that its preferences are as the human intends; this is largely outside of Cooperative AI, which takes the fundamental preferences of agents as given. When alignment succeeds, the human-machine dyad then possesses pure common interest, which is an edge case in the space of cooperation problems. Absent sufficient success at alignment, these fields then invest in mechanisms of control, so that the human's preferences are otherwise dominant; such methods of control are also not the primary focus of the science of cooperation. Human-machine interaction and alignment can be interpreted as working on ``vertical'' coordination problems, where there is a clear principal; the heart of Cooperative AI concerns ``horizontal'' coordination problems, where there are multiple principals whose preferences should be taken into account. Accordingly, human-machine interaction and alignment emphasize control and alignment, and relatively underprioritize problems of bargaining, credible communication, trust, and commitment.

Conversations about Cooperative AI ought to also include work on tools and infrastructure relevant to human cooperation, though this work is typically more narrowly targeted at a specific product need. Nevertheless, a component of the Cooperative AI bet is that work on those tools would be enhanced through greater connection to the broader science of cooperation, and the science of cooperation would be similarly enhanced by learning from the work on those tools.

\section{Cooperative Capabilities}\label{section:capabilities}

\begin{wraptable}{r}{4.5cm}
\vspace{-3mm}
\begin{tabular}{lllll}
& C & D & & 
\\ \cline{2-3}
\multicolumn{1}{l|}{C} & \multicolumn{1}{l|}{1, \quad 1} & \multicolumn{1}{l|}{-0.1, 0} & & \\ \cline{2-3}
\multicolumn{1}{l|}{D} & \multicolumn{1}{l|}{0, \quad 0.5} & \multicolumn{1}{l|}{1,\quad 0} & & \\ \cline{2-3}
\end{tabular}
\caption{One-sided assurance game. (C,C) is the mutual best outcome and unique equilibrium. Given uncertainty about the other's payoffs, \textbf{Row} may still choose D.}
\label{game1}
\end{wraptable}

The above described some of the key dimensions of cooperative opportunities. We now discuss how specific strands of research can contribute by producing relevant capabilities for promoting cooperation. These are mostly cognitive skills of agents (per the individual perspective), but they also include properties of hardware and the capabilities of institutions. We organize discussion of cooperative capabilities according to the whether they address (1) understanding, (2) communication, (3) commitment, or involve (4) institutions. We illustrate this framework using simple strategic games.

At the most basic level, the decisions of agents are affected by their \textit{understanding} of the payoffs of the game and of the other player's beliefs, capabilities, and intentions. Consider a one-sided assurance game
such as depicted in Table \ref{game1}, where the players have a mutual best outcome in (C,C), but player \textbf{Row} may choose D out of fear that \textbf{Column} will play D. To make this concrete, imagine that these are researchers choosing what problems to work on. \textbf{Column} strictly prefers to work on problem ``C'', and would like to collaborate with \textbf{Row} on it. \textbf{Row} is happy working on either problem, so long as it is in collaboration with \textbf{Column}. If \textbf{Row} does not understand \textbf{Column}'s preferences, \textbf{Row} might choose D, to both of their loss. If, however, \textbf{Row} knows \textbf{Column}'s preferences, then \textbf{Row} will predict that \textbf{Column} will choose C, and thus \textbf{Row} will also choose C, making them both better off.

\begin{wraptable}{r}{4.5cm}
\begin{tabular}{lllll}
& Stag & Hare & & 
\\ \cline{2-3}
\multicolumn{1}{l|}{Stag} & \multicolumn{1}{l|}{1, \quad 1} & \multicolumn{1}{l|}{-1.2, 0} & & \\ \cline{2-3}
\multicolumn{1}{l|}{Hare} & \multicolumn{1}{l|}{0,\quad -1.2} & \multicolumn{1}{l|}{0, \quad 0} & & \\ \cline{2-3}
& & & &
\end{tabular}
\vspace{-5mm}
\caption{The Stag Hunt coordination game, where the efficient equilibrium is risk dominated.}
\label{riskdominatedcoordination}
\end{wraptable}

However, in many cooperative games, such as Stag Hunt (Table \ref{riskdominatedcoordination}), understanding of payoffs is not sufficient for cooperation because there are multiple equilibria. Here the players also need some way of coordinating their intentions, actions and beliefs to arrive at the efficient outcome. \textit{Communication} offers a solution. If \textbf{Row} can utter and be understood to be saying ``Stag'', then \textbf{Column} will believe \textbf{Row} intends to play ``Stag'', and will thus also play ``Stag''. Your announcement of ``Stag'' will be ``self-committing'', since your partner's best response will be to play stag, which implies that you should now do so too \cite{FARRELL1988209}.\footnote{Formally, the equilibrium (Hare, Hare) is not ``neologism proof'', which means that it is not robust to the invention of a word that the other player understands and is credible \cite{farrell1993meaning}. The story is a little more complicated than this, once you invoke recursive reasoning \cite{10.2307/27871271}.}

\begin{wraptable}{r}{5.5cm}
\begin{tabular}{lllll}
& Swerve & Straight & & 
\\ \cline{2-3}
\multicolumn{1}{l|}{Swerve} & \multicolumn{1}{l|}{1, \quad 1} & \multicolumn{1}{l|}{0.5, 1.5} & & \\ \cline{2-3}
\multicolumn{1}{l|}{Straight} & \multicolumn{1}{l|}{1.5,\quad 0.5} & \multicolumn{1}{l|}{0, \quad 0} & & \\ \cline{2-3}
& & & &
\end{tabular}
\vspace{-2mm}
\caption{The Chicken game, where a credible commitment device can enable equilibrium selection.}
\label{chicken}
\end{wraptable}

Communication can be complicated by conflicting incentives, as it can produce incentives to misrepresent one's beliefs or intentions \cite{Crawford1982Strategic, fearon-rationalist}. 
In the game of Chicken (Table \ref{chicken}), for instance, each player prefers the equilibrium where they drive straight and the other player swerves. And so while one player may clearly express that they intend to drive straight, the other player may feign deafness, disbelief, or confusion. When talk is cheap and incentives conflict, players may simply ignore each other;  
even with abundant opportunity to talk, two players may still fail to avoid a collision.

Where communication may fail, \textit{commitment} capabilities may help. %
In Chicken, for example, if one player can credibly commit to driving straight, then the best response for the other is to swerve, averting disaster. In the Prisoner's Dilemma---often used as the quintessential social dilemma---cooperation can be achieved if one player can make a conditional commitment: to play C if and only if the other plays C. Commitments can be constructed in various ways, such as by a physical constraint, by signing a binding contract \cite{10.2307/1289373}, by sinking costs \cite{fearon1997signaling}, by repeated play and the desire to maintain a good relationship, and by the collateral of one's broader reputation \cite{mailath2006repeated}. %

Finally, cooperation may not be achievable without supporting social structure, which we inclusively term \textit{institutions}. Institutions involve ``sets of rules'' which structure behaviour and vary in the extent to which they are formal, detailed, centralized, and intentionally designed. They may consist of \textit{conventions}---self-enforcing patterns in beliefs---such as rules about what side of the road to drive on. They may involve \textit{norms}, which further reinforce pro-social behaviour through informal sanctions. They may involve formal rules, roles, and incentives, such as we see in constitutions and governments. 

\begin{wraptable}{r}{5.5cm}
\begin{tabular}{lllll}
& C & D & & 
\\ \cline{2-3}
\multicolumn{1}{l|}{C} & \multicolumn{1}{l|}{1, \quad 1} & \multicolumn{1}{l|}{-0.5, 1.5} & & \\ \cline{2-3}
\multicolumn{1}{l|}{D} & \multicolumn{1}{l|}{1.5,\quad -0.5} & \multicolumn{1}{l|}{0, \quad 0} & & \\ \cline{2-3}
& & & &
\end{tabular}
\vspace{-2mm}
\caption{The Prisoner's Dilemma game, in which the one-shot equilibrium is mutual defection (D, D).}
\label{prisoner}
\end{wraptable}

With respect to strategic games, institutions can be interpreted as characterizing equilibrium selection, or of involving stronger interventions on the game, such as linking the game to adjacent games or inducing changes to the payoffs. For instance, in the Prisoner's Dilemma (Table \ref{prisoner}), the one-shot game has no cooperative Nash equilibrium, whereas the infinitely repeated game with discounting contains subgame-perfect equilibria that support cooperation, such as tit-for-tat \cite{axelrod1981evolution}.  Effective institutions often have the properties that they promote cooperative understanding, communication, and commitments. Fearon \cite{Fearon2020coopAI} conjectures that the problem of designing institutions---of ``changing the game''---is the harder, and more important, kind of cooperation problem.

\subsection{Understanding} 
\label{section:understanding}

In any setting, an agent would do well to understand the world: to be able to predict the (payoff relevant) consequences of their actions. In strategic settings---where outcomes depend on the actions of multiple individuals---it also helps to be able to predict the behaviour of other agents.\footnote{We use the terms \textit{behaviour}, \textit{strategies}, and \textit{policies} largely interchangeably here.} This is thus also true of cooperative opportunities: the ability to predict behaviour can be critical for achieving mutually beneficial outcomes. This was illustrated above by the one-sided assurance game (Table \ref{game1}), where improved understanding of the beliefs, preferences, or strategy of \textbf{Column} could be necessary for the players to maximize their joint welfare. These predictions may be explicit, like in model-based reinforcement learning and search, or they may be implicit in the ways a strategy is adapted to the strategies of others, such as arises from evolutionary adaptation and model-free reinforcement learning. 

We use the term \textit{understanding} to refer to when an agent adequately takes into account (1) predictions of the consequences of actions, (2) predictions of another's behaviour, or (3) contributing factors to behaviour such as another's beliefs and preferences.\footnote{This framework relates to the beliefs, desires, and intents framework for planning in a social context \cite{bratman1987intention, lee1994prs,d1997formal,huber1999jam,bellifemine2007developing}.}
Note again that though ``understanding'' connotes deliberate reasoning, we have defined it here to also include when behaviour is adapted to implicit predictions, as would arise from evolutionary selection of policies. 

Our discussion of understanding begins from a simple game-theoretic setting. We then complicate it with uncertainty, complexity, and deviations from perfect rationality. We divide our discussion into the problems of understanding the world, behaviour, preferences, and recursive beliefs. Understanding of the mental states (beliefs, goals, intentions) of another agent is also sometimes called theory of mind \cite{baron1985does}. %

\phantomsection
\subsubsection{The World}
The primary problem confronting an agent who is seeking to craft a payoff-maximizing policy is that of understanding the (payoff relevant) consequences of its actions. This problem might be entirely subsumed (in a model-free way) in learning a policy function (best policy) or a value function (estimate of the value of actions), or it could be complemented with an explicit model of the world\footnote{Also called an environment model, dynamics function, or transition function; we use these terms interchangeably.} and representation of the state of the world \cite[\S 1]{sutton2020reinforcement}. %
The heart of single-agent reinforcement learning is thus the problem of understanding the world. 

Introducing other agents enriches and complicates an agent's understanding problem in many ways, as elaborated below. One particular way involves using another agent's actions to infer that agent's \textit{private information} about the state of the world. This is relevant to the cooperation problem because sometimes the revelation of an agent's private information would be helpful for cooperation. For example, suppose an investor and an entrepreneur are considering whether to start a business; with their initial information they might each be too uncertain to make the necessary investments. However, by credibly sharing their respective private information, they may be able to confirm whether a joint venture would in fact be mutually beneficial. 

As will be discussed below, given sufficient common interest and a means of communication, this problem is easily overcome. However, if the agents lack a means of communication, then the uninformed agent may need to draw inferences in a more indirect manner. If the agents lack sufficient common interest, then the informed party's utterances cannot be trusted. In such a setting, agents may need to rely on ``costly signals''---actions which reveal information because they are too costly to fake---for achieving cooperation. The canonical example is of a student getting a good grade in an arduous course being a costly signal of certain aptitudes (\cite{spense1973job, zahavi1977reliability, barclay2013strategies}).\footnote{Some behaviours are impossible to fake no matter an agent's effort. Jervis \cite{jervis1989logic} calls these ``indices'', distinguishing them from ``signals''. Indices can be conceptualized as at the extreme end of the continuum of costly signals, where their costs are infinite for the wrong types of agents.}

AI tools could help humans jointly learn about the world in ways that would improve cooperation. For example, trusted AI advisors could help humans better understand the consequences of their actions \cite[320]{harari2016homo}.%
Other examples, such as privacy-preserving machine learning, will be discussed under Communication.

\subsubsection{Behaviour}

While understanding the environment is the full problem facing an isolated individual, in multi-agent settings an individual also benefits from anticipating the actions and responses of other agents. This is particularly true for cooperation.\footnote{We can conceptualize cooperation as a Pareto-superior action profile (a situation where players undertake a mutually beneficial set of actions) when there was some possibility of a Pareto-inferior action profile. If there was no possibility of a Pareto-inferior outcome, then there was no opportunity to cooperate (or to not cooperate).} 

To sustain a cooperative equilibrium requires some understanding of each other's \textit{strategy}---the ways that the agent will behave in response to different actions---in order to decide whether cooperative actions will be rewarding and defection unrewarding. 
This level of mutual understanding of behaviour is implicit in the concept of a Nash equilibrium, which requires that each strategy be a best response to others' strategies; each strategy implicitly takes into account a correct prediction about others' behaviour. Going further to accommodate incomplete information, the solution concept of a Bayesian Nash equilibrium, and refinements like a Perfect Bayesian equilibrium, also explicitly require that the agent's beliefs be consistent with the strategies of other players. These solution concepts thus assume a certain degree of mutual understanding of behaviour. There also exist more radically cooperative solution concepts, which require even greater mutual understanding, of superrationality \cite{hofstadter2008metamagical, fourny2015perfect} and program equilibrium \cite{tennenholtz2004program, barasz2014robust, critch2019parametric, gentry2009fully}. Likewise, there exist weaker definitions of cooperative outcomes which require greater understanding of others, such as Kaldor-Hicks efficiency \cite{10.2307/2225023, 10.2307/2224835} which only requires that the arrangement be Pareto-improving after a hypothetical transfer from the better off. %

To illustrate, consider how the strategy tit-for-tat in iterated Prisoner's Dilemma takes into account (implicit or explicit) predictions about the others' behaviour. It understands that being \textit{nice} (playing C initially) may induce cooperative reciprocation; that being \textit{forgiving} (playing C after C, even given a history of D) may allow the players to recover cooperation after a mistake; that being \textit{provocable} is a good deterrent (that playing D after D will reduce the chance of the other defecting in the future). Tit-for-tat also has the critical property that it is \textit{clear}, making it easy for others to, in turn, understand \cite[54]{axelrod1984evolution}. This understanding may be implicit, such as if the strategy emerges from an evolutionary process, or it could be explicit if it was deployed by a reasoning agent, like a game theorist after reading Axelrod.

Compared to sustaining a cooperative equilibrium, moving to a cooperative equilibrium often poses a greater challenge of understanding. Experience can no longer be relied on for evidence that the cooperative equilibrium is in fact beneficial and robust. Instead, the agents have to jointly imagine or stumble towards this new equilibrium; achieving such a shift in behaviour and expectations can be difficult. Achieving cooperation, rather than just sustaining it, thus may often require a deeper and more theoretical form of understanding.%

Understanding of behaviour can be achieved in many ways which can be arrayed on a spectrum from being more empirical to more theoretical; this distinction relates to that between model-free and model-based learning~\cite{dayan2009goal}. On the most empirical end, we have learning processes that lack any in-built ability to plan, like simple evolutionary processes, classical conditioning, and model-free learning. Some cooperative equilibria are stumbled upon from these kinds of experiential processes, such as a sports team which learns from extensive training to intuitively coordinate like a single organism. There are learning techniques which can help agents to understand each other, such as \textit{cross-training}, in which each teammate spends some time learning to perform the roles of others \cite{nikolaidis2015improved}.  At the population level, regularities often naturally arise which coordinate \textit{intentions} and behaviour; these we call \textit{conventions}. Consider the what-side-of-the-road-to-drive-on game, the solution of which in a particular jurisdiction was emergent before it was codified in law~\cite{young1996economics}. %

At the most theoretical end of the spectrum of understanding, agents may be able to reason through the vast strategy space and identify novel cooperative equilibria. Such theoretical (model-based) understanding has the advantage of permitting larger jumps in the joint strategy space, though dependence on a learned (imperfect) model brings with it additional learning costs and risks of error \cite{huys2015interplay}. 

Sufficient understanding of an agent's strategy is all that is needed for agents to identify and sustain feasible cooperative equilibria, since these are, after all, simply stable strategy profiles. However, achieving such sufficient understanding of strategy from behaviour alone is generally implausible: the strategy space for most games is computationally intractable, and strategy itself is unobservable and subject to incentives to misrepresent. Instead, agents often do well to understand the causes of others' behaviour, such as other's private information, preferences, and (recursive) beliefs themselves.

\subsubsection{Preferences}
Sometimes the critical information needed to predict behaviour to achieve cooperation is about an agent's preferences, also variously called payoffs, values, goals, desires, utility function, and reward function. This was illustrated in the assurance game in Table \ref{game1}, where \textbf{Row}'s uncertainty about \textbf{Column}'s preferences could lead to the mutually harmful outcome of (D, C). However, eliciting and learning preferences is not an easy task. Even under pure common interest, preferences may be computationally intractable; for example, humans are probably not able to express their preferences about broad phenomena in a complete and satisfactory manner \cite{bostrom2014superintelligence, russell2019human}.
Humans may not even have adequate conscious access to their own preferences \cite[138]{russell2019human}. %
Complicating this process further, in the presence of conflicting incentives, agents may have incentives to misrepresent their preferences. 

AI research on learning of preferences is growing in prominence, in part because this is regarded as a critical direction for the safety and alignment of advanced AI systems (\cite{russell2015research, christiano2017deep}). %
 Some of this research seeks to learn directly from an agent's behaviour (see recent survey on such work~\cite{albrecht2018autonomous}), where the agent is oblivious or indifferent to the learner (often called inverse reinforcement learning, or IRL) \cite{ng2000algorithms}. It is often critical to inject sufficient prior knowledge \cite{armstrong2018occam} or control \cite{amin2017repeated, sadigh2016information} to produce sufficiently useful inferences.

Some preference learning research takes place in an explicitly cooperative context, where the observee is disposed to help the observer, and may even learn to be a better teacher; this is sometimes called cooperative IRL \cite{hadfield2016cooperative} or assistance games \cite{shah2020neurips}. 
This research may involve explicit pairwise comparisons~\cite{akrour2012april,christiano2017deep,kreutzer2018reliability, stiennon2020learning}, goal state examples~\cite{bahdanau2018learning},  demonstrations~\cite{ibarz2018reward,tung2018reward,brown2019extrapolating}, or other kinds of annotations~\cite{jeon2020reward}. 

As Figure \ref{figure:coopAIpanels} illustrated, safety and alignment can be regarded as the complement to Cooperative AI for achieving broad coordination within society and avoiding outcomes that are harmful for humans. Safety and alignment address the ``vertical coordination problem'' confronting a human principal and a machine agent; Cooperative AI addresses the ``horizontal coordination problem'' facing two or more principals. In the near future, preference learning is also likely to be critical for the beneficial deployment of AI agents that interact with humans in preference-heterogeneous domains, such as with writing assistants \cite{brown2020language, leike2018scalable} and other kinds of personal assistants. 

\subsubsection{Recursive Beliefs} %
The exposition so far has largely focused on what can be called ``first-order understanding'', which avoided the recursion of how one agent's beliefs can be about the beliefs of others, which themselves may be about the beliefs of the first, and so on.  In strategic equilibria, behaviour and beliefs must be consistent and thus can involve recursive relations.
Recursive beliefs are sometimes called recursive mind-reading and are a critical skill for agents in socially complex environments. Humans, for example, have been shown to be capable of at least seven levels of recursive mind-reading \cite{o2015ease}. In AI research, recursive mind-reading has been explored in negotiation settings  \cite{Weerd2015NegotiatingWO, de2017negotiating} and is believed to be important for games like Hanabi \cite{bard2020hanabi}.%

If a proposition is known, is known to be known, and so on to a sufficiently high level, the proposition is said to be common knowledge.\footnote{Many scholars define common knowledge as requiring infinite levels of mutual knowledge, but this is likely an overly strong definition for humans. Chwe \cite[75-77]{chwe2013rational} offers several ways to relax this assumption. Instead of requiring a 100\% confident belief (``knowledge'') at each level, one can require sufficiently confident belief (say, over 90\%). Common knowledge may be achieved through a single recursive step. Perhaps $k$ levels of mutual knowledge is in practice sufficient for humans, where $k$ is 2 or 3. Humans may have various heuristics which are understood to achieve (infinite-level) common knowledge, such as eye contact.} High-order beliefs and common knowledge play important roles in social coordination, from supporting social constructs such as money \cite{harari2015sapiens, searle1995construction}
to collective action like revolution \cite{lohmann1994dynamics}. Attention to common knowledge has been productive in recent AI research \cite{DBLP:journals/corr/abs-1810-11702}%
. %

To be clear, like most cooperative capabilities, while competence at recursive mind-reading can sometimes be critical for cooperation, at other times it may undermine it. To offer a theoretical example: in a finitely repeated Prisoner's Dilemma, if the agents have mutual knowledge of the length of the game to an order greater or equal to the length of the game, then the logic of backward induction will lead them to defect immediately; whereas if they only have mutual knowledge to an order less than the length of the game, then backward induction cannot fully unravel a cooperative equilibrium. Empirically, there is also evidence that deliberative reasoning can undermine heuristic cooperative behaviour: 
participants in public goods games may contribute more when they are forced to make their decision under time pressure \cite{rand2012spontaneous, rand2014social}. This suggests that, for most people, the default intuitive strategy may be cooperation, but with further reasoning they will consider the possibility of defection.

\subsection{Communication}  
\label{section:communication}

Communication can be critical for achieving understanding and coordination. By sharing information explicitly, agents can often more effectively gain insight into one another's behaviour, intentions,
and preferences than could be gleaned implicitly from observation and regular interaction. Such an exchange of information may therefore lead to the efficient discovery of, convergence on, and maintenance of Pareto-optimal equilibria. As a simple example, two individuals who enjoy each other's company would do well to compare their calendars to find a time to meet, rather than each trying independently to infer when the other might be free. Likewise, division of labour benefits from coordination contingent on effective communication:  one agent may announce an intention to speak with a client, allowing the other to focus on updating the firm's accounts. The information exchanged may take a usefully compressed form, providing abstractions that aid cooperation. ``Let's carry this box into the lounge'' is far more efficient and flexible than a series of low-level motor instructions for several individuals.

In the simplest setting for communication, we can imagine two individuals with pure common interest and similar background knowledge about the world, with no constraint on data transfer, but each with some private information relevant to cooperation (which may include their intentions). For example, they may be playing a symmetric coordination game. 
Even here, communication is not trivial, but requires sufficient \textit{common ground} \cite{Clark1991-CLAGIC} so that an agent can interpret the other's message. With no common ground, all signals will be meaningless. 
The need for common ground will be greater for more complex messages.

In practice, most communication channels have limitations, such as in bandwidth and latency. 
Cooperation under limited bandwidth calls for efficiently compressed information transfer, and the most suitable type of compression depends on the cooperative context, such as what the agents are trying to achieve and whether the agents are humans, machines, or both. Cooperation under high latency requires that agents are able to communicate effectively despite temporal delay of messages, which may be challenging in a fast-moving setting.

Sometimes one agent has critical knowledge to impart to another agent. This knowledge may be some private information, a skill, or even a protocol for more effective communication. In such settings, the agents need to cooperate through a teacher-student relationship. Building agents who can be, or can learn to be, good teachers and good students is an active area of research. Challenges include the teacher providing a suitable curriculum and having adequate theory of mind, improved opponent modeling and shaping strategies, and greater sample efficiency in student learning.

Communication is easiest under pure common interest. When agents have conflicting interest, a host of new challenges can arise. Agents may now have to worry about being manipulated by the other's signals. Can the agent's trust each other to communicate honestly, given incentives to misrepresent? Do agent's have the ability to detect dishonesty and deception? Can agents protect their communication channel by isolating their common interests, by norms such as honesty, or with other institutional arrangements? Given the prevalence of mixed-motives settings, building artificial agents capable of cooperative communication in that context may be a critical goal.

\phantomsection
\subsubsection{Common Ground}
The first component of communication is common ground: presumed background knowledge shared by participants in an interaction \cite{stalnaker2002common}.\footnote{``Common ground'' has gained myriad definitions in fields ranging from machine learning to psychology \cite{Traum1994Computational, Pickering2004Toward}.} Such background information may take the form of shared understanding of the physical world, a shared vocabulary for linguistic interaction, or a shared representation of some relevant information. 

Some degree of common ground is necessary for meaningful communication. If the recipient of a message has no idea what the terms within a message refer to, the recipient will not be able to make sense of the message content, and the message therefore cannot guide action or improve communication. One can divide much of the effort involved in communication between the initial work of building these representations and the subsequent use thereof. Humans, for instance, spend their early years learning to ground language such that they can use terms in ways that their family and others in their community will find intelligible \cite{tomasello2013, bohn2017}. This large initial investment enables more substantive and less expensive communication later in life.

The trade-off between the fixed costs of developing common ground representations and the variable costs of using those representations to communicate makes different levels of complexity more or less optimal under different settings. For instance, scuba divers must agree on a small vocabulary of sign language where each term has a precise meaning for unambiguous safety \cite{doi:10.1177/1468797611432040}. By contrast, American Sign Language has all the ambiguity, compositionality, and open-endedness typical of spoken languages, appropriate for its use for cooperation in a much richer set of circumstances \cite{goldin2017gesture}.

Common ground will vary depending on the kinds of agents and the context of the cooperation problem.  In machine-machine communication, the common ground is not likely to be immediately human-interpretable. Hard-coded machine-machine communication \cite{AMODU2018255} already underlies the current information revolution. Such systems tend to have fixed protocols, enabling domain-specific cooperation, such as in e-commerce. Future research could enable machine-machine communication in general domains, for instance by employing learning or evolutionary algorithms. The question of how to bootstrap common ground between machines is often referred to as emergent communication \cite{wagner2003progress, sukhbaatar2016learning, foerster2016learning, DBLP:journals/corr/abs-1804-03980}. 

AI research is also facilitating the finding of common ground between humans. Machine translation is the most prominent such example: by mapping from one set of representations (e.g., German) to another (e.g., Chinese), such systems can enable rich human-human communication without the need for either party to learn an entirely new language. Improved translation, by removing barriers to communication, may lead to increased international trade \cite{brynjolfsson2019does}, higher productivity (through increased competition and efficient reallocation of resources), and a more borderless world \cite{weber20202020s}. Beyond linguistic translation, AI systems also play an important role in mapping different modalities of communication, for instance in text-to-speech \cite{ning2019review, klatt1987review} and speech-to-text systems \cite{shadiev2014review, nassif2019speech}, opening up new technological possibilities for human communication, including for people with vocal impairment.   

Perhaps the greatest challenge is the development of common ground for communication between humans and machines. In the case of linguistic communication, this capability underlies the challenge of building a useful chatbot.  
Early such natural-language generation systems, such as ELIZA \cite{10.1145/357980.357991}, were based on hard-coded rules; the current state-of-the-art \cite{brown2020language} uses large-scale unsupervised learning combined with a neural memory architecture \cite{vaswani2017attention}.
Complementary to natural language is work on building common ground in games where actions have communicative content, such as Hanabi \cite{bard2020hanabi}, no-press Diplomacy \cite{kraus1995designing}, and Overcooked \cite{carroll2019utility}. %
The communication of human preferences to machines can be conducted using a variety of linguistic and non-linguistic methods, including comparisons, demonstrations, corrections, language, proxy rewards, and direct rewards \cite{jeon2020reward}.
Other deep-learning-based approaches include imitation learning from human data \cite{carroll2019utility, paquette2019press}, policy iteration and search \cite{anthony2020learning, lerer2019improving}, exploiting symmetries \cite{hu2020otherplay}, and Bayesian reasoning \cite{foerster2018bayesian}.

\subsubsection{Bandwidth and Latency}

The bandwidth of a communication channel measures the amount of information that can be transferred over the channel in a given unit of time. To consider human-human cooperation, the versatility of human vocal cords seems to enable humans to convey more information per second than other primates \cite{ghazanfar2008evolution, fitch2018biology}. 
However, spoken language still has far lower bandwidth than our sensory and cognitive experiences, necessitating compression. The open-ended and flexible means by which human language achieves this compression is inextricably linked to our species-unique cooperation skills \cite{scott2014speaking}.  %
Considering the low-bandwidth end of the communication spectrum, one finds (long-range) communication methods used by humans such as smoke signals and maritime flags. These work well for transmitting a small set of basic messages, but would be ill-suited to negotiating a complex legal agreement. At the other end of the spectrum, fixed-protocol machine-machine communication, such as inter-server networking within data centres, can have far higher bandwidth than human language.

Research in the aforementioned emergent communication paradigm seeks to develop common ground between machines, given a limited-bandwidth communication channel. When augmented with appropriate biases \cite{eccles2019biases}, such multi-agent systems can cooperate to solve a variety of tasks, from negotiation \cite{DBLP:journals/corr/abs-1804-03980} to sequential social dilemmas \cite{DBLP:journals/corr/abs-1810-08647}, even under conditions where the bandwidth must be minimized \cite{mao2020learning}. However, the languages which emerge between agents typically do not display the classic hallmarks of human language, such as compositionality and Zipf's law \cite{kottur2017natural, DBLP:journals/corr/abs-1905-12561, lazaridou2018emergence}. Therefore, while the approach of emergent communication may work well for machine-machine systems, much work remains to be done on crossing the human-machine barrier. A related and promising research domain is human-computer interface development, with (deep learning enabled) advances in brain-computer interfaces representing an especially radical direction of advance  \cite{zhang2019survey}.

Latency refers to the delay between a message's transmission and reception. This can vary independently of bandwidth, and it shapes the degree of autonomous behavior required by each agent in a cooperative system, as well as the amount of planning and prediction necessary when crafting a message. 
Consider communication between NASA's Perseverance Rover on Mars and NASA's headquarters in the United States. Because of the latency involved in interplanetary sharing of information (between three and 23 minutes, depending on the planets' relative positions), NASA engineers cannot feasibly control the rover's actions in the same way they could if it operated locally \cite{Beaton2017Extravehicular}. Instead, Perseverance has to navigate the Red Planet with some autonomy, 
with NASA communicating high-level guidance as opposed to low-level motor direction.

The connections to cooperation are clear: latency affects actors' ability to coordinate in fast-moving circumstances. %
The cooperative importance of reducing latency is exemplified by the Moscow-Washington hotline, which was installed to accelerate communication following the near catastrophe of the Cuban Missile Crisis, during which diplomatic messages could take more than 12 hours to process.     
In general, latency shapes the type of agreements into which actors can enter, and therefore the design of communicative agents \cite{blake2003, chen2020delayaware}. The problems presented by high latency may be particularly pronounced in situations where one individual seeks to affect the learning of another, for instance by providing reward \cite{yang2020learning, lupu2020gifting} or imposing taxation \cite{zheng2020ai}, since the effects of learning may only be apparent at a later time.

\subsubsection{Teaching}

A key feature of cooperative intelligence is the ability to teach others. Teaching enables the communication of practically useful knowledge and thereby greatly increases opportunities for joint gains. %
This can be viewed as an advanced form of social learning \cite{bandura1977social, hoppitt2013social}, where not only do individuals learn by observing more capable agents, but those agents also modify their behaviour in order to elicit better learning from their students. %
 The evolution of teaching is associated with increased cumulative cultural abilities in many species \cite{thornton2010}, and stands out as an important component of the social intelligence of humans \cite{fogarty2010}, perhaps even tied to the origins of language itself \cite{laland16}. 

An important component of 
learning to teach is learning to be taught. In the evolutionary biology literature, this often goes under the moniker of observational learning, and it is closely related to what in AI is known as imitation learning. Following the deep learning revolution, the field of AI has demonstrated increasing interest in transferring knowledge from teachers---human and artificial---to student algorithms, particularly given the sizable datasets of human demonstrations (e.g., those available on video sharing platforms). Direct methods of learning from human teachers, such as imitation learning~\cite{pomerleau1989alvinn,schaal1999imitation}, have become widespread for real-world robotics research (see for example~\cite{rajeswaran2018learning,abbeel2010autonomous}). In turn, teachers can learn to select the most useful demonstrations or lessons for their students \cite{cakmak2012algorithmic}. Distillation of policies between agents can help achieve higher performance across a wide range of environments \cite{schmitt18}. A more model-free approach was recently explored \cite{DBLP:journals/corr/BorsaPMP17, woodward2020learning}, whereby a student learned to follow a teacher in a gridworld environment via curriculum-based reinforcement learning (RL). Competence in learning from others can permit faster learning of skills and superior zero-shot transfer performance \cite{ndousse2020multi}.

Learning to teach has received increasing attention in recent years \cite{da2017simultaneously, beck1998learning, omidshafiei2019learning, zimmer2014teacher}. This is particularly challenging, since teacher performance can only be evaluated after the student has learned. Therefore, the feedback signal for teacher learning may be temporally distant from the start of teaching. There are various nascent methods for addressing this problem, including meta-learning students \cite{zheng2020ai}, second-order gradient methods \cite{foerster2017learning, letcher2018stable}, and inverse reinforcement learning \cite{cakmak2012algorithmic}.

Machines that can teach humans hold much promise for society. Indeed, it may only be once high-performing agents develop the ability to instruct that we realize their full potential: super-human algorithms, such as AlphaZero \cite{silver2017mastering} in the domain of chess, go, and Shogi, would generate much more utility if humans could learn directly from their inner workings. This topic has foundations in the rapidly evolving domain of interpretable and explainable AI, which in turn has significance for technical safety. However, effective teaching goes beyond interpretability to include questions of interface design for human-computer interaction \cite{shneiderman2016designing}, the evaluation of effective pedagogy \cite{goe2008approaches}, summarization methods \cite{nallapati2016abstractive, Zhang2019HIBERT}, and knowledge distillation \cite{Yim2017Gift}. %

The notion of teaching is also relevant to considerations of equilibrium selection and the construction of welfare-improving equilibria, namely through \textit{correlated equilibria} \cite{AUMANN197467} and coarse correlated equilibria. 
Evolutionary models demonstrate that correlated equilibria can solve mixed-motive problems \cite{morsky2019evolution, LeePenagos2016LearningTC}. Furthermore, independent no-regret learning algorithms provably converge to coarse correlated equilibria \cite{hart2000simple}. Further research is warranted into the nature of communicative equilibria and how they might support cooperation.

\subsubsection{Mixed Motives}

Communication generally becomes more difficult the more agents'c preferences are in conflict. The fundamental problem facing communication under mixed-motives is the incentive to \textit{deceive}, and the consequent risk of being deceived. Under pure conflicting interest, agents have no incentive to communicate: any message that one agent would want to be heard, the other agent will not want to hear. Conflicting interest can thus destroy much of the potential of communication for achieving joint gains
 \cite{camera2011communication}. As agents' preferences are more aligned, cheap-talk communication generally increases in efficacy \cite{Crawford1982Strategic}. Alternatively, under mixed-motives, credible communication can sometimes be achieved through costly signals to overcome the incentive problems preventing honest communication, with the aforementioned canonical example being of a student's grade being a credible signal of certain aptitudes and interests (\cite{spense1973job, zahavi1977reliability, barclay2013strategies}). As artificial agents are deployed in society, scientists, policymakers, and the general public will have to grapple with complex questions about what communication norms machines should be expected to abide by, such as perhaps declaring their (machine) identity \cite{o2019google} and committing to avoid deception, and how these norms should be reinforced.

Research in mechanism design offers opportunities for building mechanisms to incentivize truthful revelation, a property known as incentive compatibility. 
AI could play an important role in devising new incentive-compatible mechanisms and in acting as a trustworthy mediator. Automated mechanism design has already achieved noteworthy results in a number of areas, including auctions, voting and matching, and assignment problems~\cite{Feng2018dl, Cai2012optimal, Cai2012algorithmic, Narisimhan2016Automated}. In many cases, function approximators can obtain (approximate) incentive compatibility in situations too complex to be tractable under closed-form methods. Given the complexity of many real-world settings, this line of research holds great promise for increasing global cooperation. 

Advances in cryptography offer other opportunities for promoting cooperative communication under mixed motives. It is increasingly possible to build (cryptographic based) information architectures which permit precise complex forms of information flow, sometimes called structured transparency \cite{Traskstructuredtransparency},  in which, for example, the owner of information can allow that information to be used for some narrow purpose while keeping it otherwise private. Successes in structured transparency can open up new opportunities for mutual gains, such as enabling privacy-preserving medical research which depends on analyzing the health data of many individuals \cite{Traskstructuredtransparency} or privacy-preserving contact tracing for pandemics \cite{shevlane2019contact}. \textit{Privacy-preserving machine learning}, which encompasses methods such as homomorphic encryption \cite{Graepel2012Confidential, gentry2009fully}, secure multi-party computation \cite{Mohassel2017secure}, and federated learning \cite{McMahan2017communication}, allows for the training of models on data without the model owner ever having access to the full, unencrypted data or the data owner ever having access to the full, unencrypted model \cite{Graepel2012Confidential, Traskstructuredtransparency}. AI research has an important role to play in advances in this kind of structured transparency.

Finally, an important mixed-motive setting for communication is negotiation. In situations lacking a formal contractual agreement protocol, ``cheap talk'' \cite{Crawford1982Strategic} can play an important role, including in human-machine interactions \cite{crandall2018cooperating}. When an agreement structure is available, automated negotiation systems may seek a cooperative outcome \cite{rosenschein1994rules,jennings2001automated,faratin2002using}. Such systems aim to reach agreements by reasoning over possible deals or iteratively making offers and modifying a deal in mutually beneficial ways~\cite{kraus1997negotiation,matos1998determining,bichler2003towards}. AI research has investigated how to design protocols and strategies for automated negotiation. Such protocols define the syntax used for communication during the negotiation, restrictions as to which messages may be sent, and the semantics of the messages~\cite{smith1980contract,chang1994speech}. Researchers have proposed many automated negotiation protocols~\cite{aknine2004extended,ito2007multi}, focusing on developing specialized negotiation languages which are expressive enough to capture key preferences of agents, but still allow for computationally efficient dealmaking~\cite{sidner1994artificial,mueller1996negotiation,wooldridge2000languages}. The advent of deep learning has opened up profitable avenues, including agents which learn to negotiate based on historical interactions \cite{oliver1996machine,narayanan2006learning,lewis2017deal}.

\subsection{Commitment}
\label{section:commitment}

The above capabilities of Understanding and Communication seek to address cooperation failures from incorrect or insufficient information. However, cooperation can fail even absent information problems. Work in social science has identified ``\textit{commitment problems}''---the inability to make credible threats or promises---as an important cause of cooperation failure. Prominent scholarship even argues that cooperation failure between rational agents requires either informational problems or commitment problems \cite{fearon-rationalist, powell2006war}.\footnote{A third cause of conflict is issue indivisibility, though this is sometimes said to still depend on the prior inability to commit to an efficient gamble which would be Pareto superior to costly conflict \cite{powell2006war}.} More broadly, a large literature has looked at the many ways that commitment problems undermine cooperation \cite{sen1985goals, north1993institutions, fearon-rationalist, bagwell1995commitment, hovi1998games, hirshleifer2000game, powell2006war, garfinkel2007economics, jackson2011reasons}.\footnote{In practice, informational problems and commitment problems are often intertwined, and solutions for them can be substitutes. It can be useful theoretically to distinguish them.} 

To illustrate, consider the Prisoner's Dilemma (Table \ref{prisoner} above), often regarded as the canonical, and most difficult, of 2-by-2 game-theoretic cooperation problems. This game involves perfect and complete information, so no amount of improved understanding or communication would help. Though the players could both be better off if they could somehow play (C, C), each has a unilateral incentive to play D, irrespective of what the other player says or intends to do. However, if one player can somehow make a \textit{conditional commitment} to play C if and only if the other player plays C, then the dilemma would be solved; the other player would now strictly prefer to also play C.

Commitment problems are ubiquitous in society. Absent the solutions society has constructed, commerce would be crippled by commitment problems. Every time a buyer and seller would like to transact, each may fear that the transaction will go awry in any number of ways; government-backed currency, credit cards, and consumer protection regulations each address some of these potential commitment problems---from a prospective buyer failing to deliver payment, to a seller delivering faulty (or no) products. %
Domestic political order depends on the ability of leaders to make credible promises. Liberal polities in particular often depend on constitutions which articulate the fundamental civic promises and social mechanisms to credibly enforce them \cite{acemoglu2012nations, north2009violence}.
When ruling elites can't make such promises, or can't trust the promises of prospective challengers, repression or civil war may be the only recourse \cite{fearon1998commitment, fearon2004some}.
 Peace among great powers itself may depend on avoiding abrupt power transitions and the commitment problems those produce \cite{powell1999shadow}.%
\bnote{Is the above too intense? Too much "political order" and war? Someone want to suggest how to lighten the mood?}

\phantomsection
\subsubsection{Commitment Solutions: Unilateral vs Multilateral; Unconditional vs Conditional}
\bnote{Integrate this reference: One device is {\it unilateral commitments}~\cite{magenau1988dual,conitzer2004computing} }
Overcoming a commitment problem often requires a {\it commitment device}, which is a device that compels one to fulfill a commitment, either through a ``soft'' change to one's incentives for taking different actions~\cite{snijders1996trust,bryan2010commitment}, such as a penalty for non-compliance, or a ``hard'' pruning of one's action space, such as implied by the commitment metaphor of burning one's boats.\footnote{Thus removing from the choice set the option of retreating of an invading force~\cite{reynolds1959burning}.}
Commitment devices may be \textit{unilateral}, where a single agent is capable of executing the commitment, or \textit{multilateral}, requiring multiple agents to consent. Commitment devices can involve an \textit{unconditional commitment} to some action or involve more sophisticated commitments \textit{conditional} on the actions of others or events. 

Unilateral unconditional commitments may be the most accessible commitment devices, since they only require that an agent have some means of shaping their own incentives or options. %
Perhaps the most common unilateral unconditional commitment is to simply take a hard-to-reverse action. These commitments are implicit in sequential games, producing less risk of agents simultaneously miscoordinating. These commitments are  more common when there is a large or salient first mover who can shift the expectations of many other players; 
examples include an "anchor tenant" in a development project or a large firm committing to a technical standard.

Unilateral \textit{conditional} commitments are typically much harder to construct, probably because they require binding oneself to a more complex pattern of behaviour. 
However, conditional commitments can be much more powerful because they can support the precise promises (or threats) that may be needed to support a cooperative venture; this is illustrated by how a conditional commitment is sufficient to overcome the Prisoner's Dilemma, whereas an unconditional commitment is not. 

Lastly, commitment devices may be multilateral, requiring multiple parties to consent to the commitment before it goes into effect. Legal contracts exemplify multilateral commitments. Most conditional commitment devices in society may be multilateral commitment devices. Though it is worth noting how a set of unilateral conditional commitments can be equivalent to a multilateral unconditional commitment. This is illustrated by the National Popular Vote Interstate Compact, which aims to replace a state-by-state first-past-the-post system with national popular voting without a change to the US Constitution. Their strategy is for US states to unilaterally commit to award all electoral votes to the presidential candidate who wins the popular vote, conditional on enough other states similarly committing  
\cite{natpopvote2020}. %

AI research can contribute in several ways. Researchers can develop languages for specifying commitment contracts and semantics for the actions taken under them~\cite{kosba2016hawk,luu2016making,frantz2016institutions}. This work will benefit by being interdisciplinary, given how the space of commitment mechanisms spans domains  such as law, politics, economics, psychology, and even physics. 
Researchers can improve our ability to reason about the strategic impacts of commitment, for example by developing algorithms for finding the optimal course of action to commit to~\cite{conitzer2006computing} or predicting how agents are likely to respond when others commit to a certain course of action~\cite{korzhyk2011stackelberg,paruchuri2008playing}. One may also examine specific domains, such as disarmament~\cite{deng2017disarmament}, to identify favorable commitment perturbations to the game that increase welfare.

\subsubsection{Devices: Reputation, Delegation, Contracts, Hardware}
\label{section:devices}
Commitment devices come in many forms, including enforcement, automated contracts, and arbitration
~\cite{boot1991credible,greif1994coordination,greif1994cultural,williams1997origins,ostrom1998behavioral,krasa2000optimal,rock2001securities,monderer2004k,mitusch2005mediation,monderer2009strong,kalai2010commitment,governatori2018legal,cong2019blockchain}. We mentioned above how agents often have some unilateral commitments available to them, if only from their ability to ``move first'', sink costs, or literally destroy some options available to them. In addition, we see several classes of other commitment devices, each of which depends on some social infrastructure: reputation, a social planner, contracts, and hardware. Each of these has distinct properties and is associated with distinct research problems. We briefly review these here and discuss some at greater length in the following section. 

First, reputation systems provide a mechanism for commitment by creating a valuable asset (reputation) which can be put up as collateral for cooperative behaviour in transient encounters. Just as a canonical solution to the Prisoner's Dilemma is to iterate the game, which then can give agents an incentive to cooperate today to preserve their reputation for being likely to cooperate tomorrow \cite{nowak1998evolution}, %
so in human society does reputation seem to undergird many cooperative achievements such as trade and debt \cite{greif1989reputation, tomz2012reputation}. AI research can assist in designing and facilitating effective reputation platforms~\cite{resnick2002trust,kornhauser1983reliance,zacharia2000trust}, thus enabling verification of agent identity~\cite{marti2003identity,gupta2003reputation}, as well as building agents who are skilled at promoting cooperative reputation systems. We discuss reputation further below. 

Another method of achieving commitment involves agents delegating decision-making power to a social planner---a trusted third party or a central authority. AI research can study such mechanisms of delegation and mediation~\cite{tennenholtz2004program,monderer2009strong} and can work to improve the efficacy of a central planner \cite{zheng2020ai}. A central authority can also provide a legal framework and enforcement, within which agents can construct multilateral commitments, namely through contracts. 

The emergence of increasingly cognitively capable algorithms, cryptographic protocols and authentication, trusted hardware~\cite{lie2003implementing,bajaj2013trusteddb}, and ``smart contracts''~\cite{christidis2016blockchains,kosba2016hawk,luu2016making,bartoletti2017empirical,governatori2018legal,cong2019blockchain, wohrer2018design} enable the delegation of increasingly sophisticated commitments to non-human entities, including commitments that are conditional on states of the world. These technologies can enable contractual commitments without requiring a central authority. These tools can also make communication more credible, such as with tamper-proof recordings from the sensors in autonomous vehicles \cite{guo2020proof}, which could provide forensic evidence in the event of an accident. %

\subsection{Institutions}
\label{section:institutions}

Achieving the requisite understanding, communication, and commitment for cooperation often requires additional social structure. Following economics and political science, we refer to this structure abstractly as \textit{institutions}.  Institutions involve a system of beliefs, norms, or rules that determine the ``rules of the game'' played by the individuals and organizations composing a collective~\cite{greif2006institutions}, shaping the actions that can be taken by individuals and the outcomes determined by these actions, resulting in stable patterns of behaviour~\cite{schotter2008economic,ordeshook1986game,knight1992institutions,austen1998social,huntington2006political,maskin2008mechanism}. \textit{Cooperative institutions} are those which support cooperative dynamics. They do so primarily by resolving coordination problems and aligning incentives to resolve social dilemmas. They may also provide structural scaffolding upon which complex inter-agent behaviours can be built~\cite{moses1995artificial}, such as by enabling agents to adopt simplifying assumptions about the behaviour of others.

For games of common interest, \textit{conventions} are patterns of expectations and behaviour which promote coordination. For mixed-motives games, these patterns can be reinforced with social reward and sanctions, which we refer to as \textit{norms}. Society can go further, allocating roles, responsibilities, power, and resources in ways designed to reproduce a pattern of desired interactions; these thicker and more formal entities are what is most commonly denoted by the term \textit{institutions}, though we also use the term in an encompassing way. 

To illustrate, consider the Prisoner's Dilemma, described in the section on Commitment and depicted in Table \ref{prisoner}. As noted, if one player is able to conditionally commit to play $C$ if and only if the other player plays $C$, then the dilemma is overcome. How can such a commitment be constructed? If the game can be made to repeat or be linked to other similar games, then cooperation may become an equilibrium; this linking of games is sometimes understood as an institution, such as is achieved in trade negotiations under the WTO or in global diplomacy within the UN.  Given a repeated game, one needs suitable expectations to support cooperation; a norm such as tit-for-tat is one such self-reinforcing norm. Cooperation can otherwise be achieved if it's possible to allocate external incentives to change the payoffs in the one-shot game or to otherwise make the conditional commitment binding; achieving these are sometimes called institutions~\cite{frieden2015world,north1993institutions}.

Institutions vary in the extent to which they are emergent vs designed, informal vs formal, and decentralized vs centralized. These properties are correlated, but not perfectly. 
Institutions may initially emerge from trial-and-error processes~\cite{ostrom1998behavioral} but then take on a more formal, designed, centralized character. For instance, in the case of file-sharing systems, participants may initially refuse to share their files with others who are not sharing. Such a rule can later be implemented formally in a peer-to-peer network or file-sharing service, where a mechanism can be introduced to limit the rate of service of participants who are not using their resources to provide service to others~\cite{golle2001incentives,lai2003incentives,legout2007clustering}. Groups may later agree on a more formal framework for making joint decisions so as to improve social welfare such as voting systems, auctions, and matching mechanisms.  We divide the following discussion into decentralized institutions and centralized institutions.

\phantomsection
\subsubsection{Decentralized Institutions and Norms}
\label{l_sect_inst_decentralized}

In decentralized institutions, there is no single central trusted authority which can make and enforce decisions on behalf of a group of agents. Instead, institutional structures will often emerge from the interactions of agents over time~\cite{ostrom1998behavioral,shoham1997emergence}, such that agents' themselves act in a way that incentivizes the desired behaviour in others (for example, through informal social punishments \cite{wiessner2005norm}). 
There is a rich literature on decentralized algorithms arising from the field of distributed computing~\cite{tanenbaum2007distributed}, which support the design and analysis of decentralized institutions.
Within multi-agent systems, many methods have been proposed which help agents to interact directly with one another, negotiate, make decisions, plan, and act jointly~\cite{smith1980contract,horling2004survey,ferber1999multi,jennings1996coordination,o1996foundations,bond2014readings,wellman1993market,vazquez2005organizing,castelfranchi1998modelling}.

One prominent way of achieving societal goals without relying on a trusted central authority is through {\it norms}.
Norms are broadly understood to be informal rules that guide the behaviour of a group or a society~\cite{sep-social-norms}.%
They constrain the behaviour of group members, often by capturing and encoding sanctions or social consequences for transgressions, and are seen as central to supporting societal coordination. One prevalent interpretation of social norms is that they can be represented as equilibria in strategic games and thus may be viewed as stable points among the group's interactions~\cite{bicchieri2006grammar,morrow2014order}.
Human groups have remarkable abilities to self-organize around social norms to overcome issues of collective action \cite{10.2307/2646923}. Indeed, the emergence of robust social norms is thought to have been a key process in the development of large-scale human civilization \cite{APICELLA2019R447, 10.2307/j.ctvc77f0d, harari2015sapiens}. It is this importance for human interaction which motivates research into how artificial intelligence can learn to recognize and follow norms.

Researchers have argued that norms can be used to organize systems of agents or influence the design of agents themselves (e.g.,~\cite{castelfranchi1999deliberative,dignum1999autonomous,vazquez2005organizing}) and 
have worked on agents who adhere to social norms,
reflecting constraints on the behaviour of agents that ensure that their individual behaviours are compatible (e.g.,~\cite{shoham1995social,conte1998autonomous,conte1995understanding,shoham1992synthesis}).
These constraints are typically imposed offline to reduce the need for negotiation and the chances of online conflict.
Furthermore, work has examined possible social norms for various environments and investigated their computational properties, both in terms of identifying predicted behaviour under various norms (for instance, in terms of the emerging equilibrium behaviour) and in identifying good norms that lead to desired behaviour (e.g.,~\cite{dignum2002desires,horling2004survey,artikis2009specifying,y2006normative,werfel2014designing,koster2020silly}). 
There has also been much work on the emergence of social norms among groups of agents (e.g.,~\cite{morris2019norm}), both in the agent-based-modelling community, where the tasks are typically abstracted matrix or network games (see for example \cite{yu2013,villatoro2013robust}), and more recently in the multi-agent reinforcement learning community (e.g., \cite{DBLP:journals/corr/abs-1902-03185, DBLP:journals/corr/abs-1803-08884, taylor2009transfer, hausknecht2015deep, pareda2017, koster2020silly}) where the state-of-the-art is temporally and spatially extended gridworld games.

This work lays a foundation for addressing important outstanding problems in Cooperative AI.
AI research could explore the space of distributed institutions that promote desirable global behaviours~\cite{Hadfield2020coopAI,criado2011open} and work to design algorithms which can predict which norms will have the best properties. 
Such algorithms already have a strong foundation to build on,
including languages for expressing societal objectives and solving them through model checking, identifying agents that are critical to achieving the global objective, or dealing with non-compliance~\cite{grossi2007designing,aagotnes2007normative,aagotnes2009power,bench2017norms}.%
Furthermore, we need to better understand how systems comprising mixtures of humans and machines devise and enforce norms, and  develop AI algorithms that are able to generalize social norms to different circumstances and co-players.

A specific decentralized dynamic for which institutions can often help is bargaining, which refers to the methods and protocols through which agents may attempt to negotiate welfare-improving arrangements. One may view a platform that enables such negotiation or bargaining as an institution. Such work includes automated negotiation systems as well as formal protocols and frameworks for multi-agent contracts, bargaining, and argumentation~\cite{smith1980contract,rosenschein1994rules,sandholm1995issues,kraus1997negotiation,jennings2001automated,mcburney2002desiderata,kakas2006adaptive,larson2001bargaining,baarslag2013evaluating}. 
Challenges in this space include the creation of formal specifications and protocols enabling interactions, computationally tractable algorithms to be used by agents, and better understanding of how to support interactions so as to yield high social welfare~\cite{kraus1997negotiation,jennings2001automated,yan2007autonomous,laengle2018twenty}.

\subsubsection{Centralized Institutions}
\label{l_sect_inst_centralized}

Centralized institutions involve an authority able to shape the 
rules and constraints on the actions of the participants.  Our understanding of these institutional structures and their properties has  been heavily influenced by social choice theory, game theory, and mechanism design.
There is  often  a  focus on the rules and axioms that should be satisfied so as to ensure desirable social outcomes are reached or to provide incentives such that agents perceive cooperation to be in their self-interest.

Social choice theory studies the aggregation of agents' preferences in support of some collective choice. Voting, for example, is a widely used class of social choice mechanism. 
Much of the research is axiomatic in nature: a set of desirable properties is proposed and then the question as to whether there exists a set of voting rules that satisfies these properties is explored. For example, Arrow's Theorem, a central result in social choice theory, states that it is generally impossible to have a non-dictatorial voting rule that also satisfies a number of reasonable properties~\cite{arrow1951social}. However, there have been significant advances in relaxing the assumptions of Arrow's Theorem and identifying and characterizing families 
of voting rules by sets of properties~\cite{brandt2016handbook}, including deepening our understanding of the impact of computation on those properties (e.g.,~\cite{bartholdi1989computational,bartholdi1992hard,faliszewski2010ai,faliszewski2010using,conitzer2002complexity,conitzer2002vote,zuckerman2009algorithms,menon2017computational}). The insights and theoretical foundations provided by social choice theory can provide guidance in the construction of cooperative institutions.

Another important strand of social choice work seeks to develop notions or protocols for \textit{fairness},  
particularly in the context of resource allocation or reward sharing.
A perceived lack of fairness in how resources, awards, or credit is shared across a group may lead to the breakdown of cooperative structures.
Such problems are common for humans in settings such as partnership or company dissolutions, divorces, dividing inheritances, or even determining how much effort to contribute towards a group project. Institutions or protocols have been developed to address these concerns, including the famous ``cut and choose'' protocol for divisible resources~\cite{brams1996fair,aziz2016discrete}, maximum Nash welfare for both divisible and indivisible resource settings~\cite{nash1950bargaining,caragiannis2019unreasonable}, and the Shapley value for group reward division~\cite{shapley1953value}.

The closely related field of \textit{mechanism design} studies when and how it might be possible to design rules or institutional frameworks so as to align the incentives of the individual agents so that it is possible to achieve a socially desirable outcome, ideally in such a way so that it is in the individual agents' best interest to truthfully share or communicate the relevant preference information.
The challenge becomes one of communication and incentives; the self-interest of agents may lead them to misreport or miscommunicate relevant information, making it impossible for the institution  to select the appropriate outcome.
While there are several impossibility results that highlight the boundary of what can be achieved for self-interested agents due to their strategic behaviour~\cite{arrow1951social,gibbard1973manipulation,satterthwaite1975strategy,green1977characterization,roberts1979characterization},
there are islands of possibility. In particular, if there is certain structure in the preferences or utility functions of agents, then mechanisms can be designed to incentivize honesty, resulting in a socially optimal outcome being selected. Examples of such mechanisms include the class of median mechanisms for when agents have single-peaked preferences~\cite{black1958theory} and the class of Groves mechanisms for when agents have quasi-linear utility functions~\cite{groves1973incentives,vickrey1961counterspeculation,clarke1971multipart}.
This latter class of mechanisms has drawn particular attention since quasi-linear utility functions naturally capture any setting where \emph{transfers} can be made between agents or between agents and the mechanism itself, where the transfers are often interpreted as payments. Auctions, widely used to allocate resources efficiently across groups of self-interested agents, are one example of mechanisms for settings with quasi-linear utility.

There remain numerous important outstanding problems that require further study as we explore the use of social choice and mechanism design to support Cooperative AI. For example, many of the foundations of social choice theory are axiomatic in nature; are these the right axioms if we consider designing institutions for collectives of humans and agents? Is it possible to combine axiomatic methods with data-driven processes~\cite{deon2020testing,armstrong2019machine}, or are there particular characterizations of social choice rules that will prove to be particularly useful for supporting cooperation and coordination (e.g.,~\cite{jiang2014diverse,caragiannis2013noisy})? How we can apply mechanism design for setting the incentive to improve social good purposes such as welfare or fairness~\cite{abebe2018mechanism}?
Finally, there is a strong interplay between \emph{understanding}, \emph{communication}, and social choice which deserves further exploration~\cite{boutilierincomplete}.

The multi-agent systems research community has long advocated for the use of mechanism design for solving coordination problems between agents~\cite{rosenschein1994rules}. With the advent of more complex agents, are there novel coordination and cooperation problems for which insights from mechanism design might prove useful, and what is the interplay between incentive structures, computation, and Cooperative AI~\cite{nisan2001algorithmic,nisan2007computationally}? For example, it has been shown that multi-robot pathfinding can be modelled as a combinatorial auction problem, where non-colliding paths are allocated to the robots (e.g.~\cite{amir2015multi}). Going forward, might it be possible to extend such ideas to navigation problems involving autonomous vehicles? There is interest in using data-driven approaches to design mechanisms for specific instances~\cite{conitzer2002complexity,sandholm2003automated}. For instance, tools from deep learning and reinforcement learning have recently been used to automatically design auctions  (e.g.~\cite{duttingFJLLP12,duetting0NPR19,tacchetti2019neural,tang2017reinforcement}). It might be useful to move beyond auctions and explore new institutional structures that best serve the goals of cooperative behaviour. 

\subsubsection{Trust and Reputation Systems}

As mentioned above, reputation provides a mechanism for aligning incentives for cooperation and for addressing commitment problems. It does so by creating a valuable asset---the reputation itself---which can then be put up as collateral to encourage cooperative behaviour. Reputation systems can exist in decentralized settings, but can often be improved by a central authority.  In typical reputation systems~\cite{resnick2002trust,marti2006taxonomy,sabater2005review,fulmer2012level,sherchan2013survey,pinyol2013computational,lu2018bars}, agents may rate each other so as to build trust between participants. These systems can be designed to reveal particular information regarding the behaviour of other agents, such as whether they have held up their part of a bargain or agreement in past transactions~\cite{fullam2005specification,jurca2009mechanisms,teacy2006travos,ramchurn2004trust}. Reputation systems are already prevalent and used by humans in e-commerce websites, such as eBay or Amazon, and information websites such as Quora or Stack Exchange~\cite{dellarocas2003digitization,leskovec2007dynamics}. While prominent and functional reputation systems are already in place in the private sector, more research is needed to understand the relative benefits of different reputation mechanisms~\cite{hazard2013macau}.

The multi-agent research community has long recognized that trust is central for supporting cooperation and coordination~\cite{castelfranchi1998principles,wang2007formal,sen2013comprehensive,cohen2019trusted}, and has used such models for maintaining, disseminating, and using information regarding the behaviour of agents~\cite{josang2009challenges,yu2013survey}. 
This research has recognized the importance of socio-economic models of trust~\cite{mayer1995integrative}, explored the relationship between trust and norms~\cite{luck2009flexible}, and used trust to support community formation~\cite{kastidou2009exchanging}.  It has also long been argued that trust will be instrumental for supporting the acceptance of robots~\cite{kaminka2013curing} and, more broadly, acceptance of and cooperation between humans and machines~\cite{melo2016people,de2017negotiating}.

\subsubsection{Cooperative AI Problems and Institutions}

While we highlighted some promising research directions above, we conclude this section with some overarching research directions we believe may advance the Cooperative AI agenda.
One promising avenue for machine learning research is institutional design, whereby (human) participants determine desiderata that the institution or mechanism should achieve and leave  the design thereof to an AI agent. This could open the door to new, innovative approaches for tackling long-established problems~\cite{zheng2020ai}. These methods could also enable institutions that take into account a richer set of features than have typically been considered in prior literature. For instance, while social choice scholarship on voting systems generally limits preference representation across alternatives to ordinal or cardinal scores, the \emph{Polis} platform aims to model user preferences by taking into account a broad set of features including opinions articulated via natural language~\cite{polis}.

It may end up that, as has happened in many domains, machine learning methods as applied to mechanism and institutional design offer major performance improvements at the cost of parsimony, interpretability, and closed-form provability---a trade-off that raises interesting  questions about what we truly value in social systems. To what extent does the liberal affinity for democratic mechanisms derive from the accountability and improved outcomes that they empirically create, and to what extent is it due to the simplicity and transparency of a system such as single-member plurality voting? The growing mistrust of election integrity in several countries suggests that the latter category has significant weight. This question is structurally similar to active debates around the implications of interpretability and bias in machine learning systems in scientific domains (when should we be satisfied with a system merely on the grounds of empirical performance and when should we push for an ``explanation'' of its decision?).

\section{The Potential Downsides of Cooperative AI}
\label{section:downsides}

All scientific and technological advances can have potential downsides, posing risks or harms. An important part of responsible research is explicit consideration of these possible downsides and exploration of strategies to mitigate the risks. We see possible downsides with Cooperative AI falling into three categories: (1) Cooperative competence itself can cause harms, such as by harming those who are excluded from the cooperating set and by undermining pro-social forms of competition (i.e., collusion). 
(2) Advances in cooperative capabilities may, as a byproduct, improve coercive capabilities (e.g., deception). 
(3) Successful cooperation often depends on coercion (e.g., pro-social punishment) and competition (e.g., rivalry as an impetus for improvement), making it hard to pull these apart. 

\phantomsection
\subsection{Exclusion and Collusion}
Advances in cooperative competence, by definition, should improve the welfare of the cooperating agents. However, enhanced cooperative competence may harm others who are excluded from the cooperating set. For example, mechanisms that promote cooperation among %
criminals---such as cryptocurrencies, the darkweb, and associated reputation systems---can be socially harmful \cite{10.1093/esr/jcx072}. %
Often, individuals collaborate to engage in ``rent seeking'': working together not to increase productivity but to transfer value from society to themselves \cite{10.2307/2117699,10.2307/j.ctt1nprdd}. Buyers or sellers in an auction can cooperate by colluding to set prices in a self-serving and welfare-harming way~\cite{Pesendorfer2000ASO}. %
In international politics, greater national cohesion and cooperation in one country can pose risks to its rivals.

``Cooperation'' can also be harmful when it undermines pro-social competition; this often manifests as collusion. Recent work in the fields of economics and law argues that the use of AI for determining prices may increase the risk of collusion between firms even without explicit communication \cite{calvano2020artificial,ezrachi2017artificial}. This would be a concerning development as competition can be a powerful mechanism for producing pro-social outcomes by incentivising effort, revealing private information, and efficiently allocating resources. It is for this reason that productive societies  forbid various kinds of anti-social ``cooperation'', such as students sharing answers during examinations, peer reviewers at a scientific journal soliciting payment from authors for a favorable review, firms coordinating on a strategy for setting prices, and policymakers soliciting personal payments.   
An open question, then, %
concerns when particular kinds of increases in cooperative capabilities lead to a net positive or net negative social outcome. What might a social planner do to incentivize the right kinds of cooperation?

\subsection{Coercive Capabilities}
Many specific capabilities that are useful for cooperation may also be useful for \textit{coercion}, defined as efforts by an actor to get something from another through threats or the use of force. %
Arbitrary improvements in coercive competence are generally not regarded as socially desirable: they may lead to an (undeserved) transfer in value from others to the more coercively competent actor in a manner that exposes society to harms, threats, and (illegitimate) uses of force. They may also lead to an increase in the use of coercion, which is generally regarded as undesirable. Whereas cooperation at least involves welfare improvements among the cooperative set, coercion can not even guarantee that.

There are many examples of capabilities which are useful for coercion as well as cooperation, and of coercive capabilities which may be learned as a byproduct of learning cooperative capabilities. While \textit{understanding} is often essential for cooperation, so is it for successful coercion; understanding a target's weaknesses and vulnerabilities confers an important advantage.\footnote{Machiavelli and Sun Tzu both counselled understanding one's adversary's capabilities and intentions, and disguising one's own.} 
In order to learn cooperative communication in mixed-motives settings, one must learn to be able to send messages interpreted by others as honest and to discern honesty from deception in others' messages; but, coercive communication benefits from the same abilities! Similarly, commitment is often essential for making credible promises, but also threats. Insights into cooperation-oriented institutional design may also be useful for promoting obedience or for manipulating institutions to serve a narrow set of interests. %

\subsection{The Role of Coercion and Competition in Cooperation}

Finally, the mechanisms and welfare implications of cooperation and coercion are often deeply intermixed, with coercive capabilities sometimes playing a critical role in cooperation. Punishment, for instance, is often critical for sustaining cooperation \cite{van2014reward, andreoni2003carrot, boyd2010coordinated, koster2020silly}. %
A prominent example is the use of legal contracts to facilitate cooperation by enabling each party to submit themselves to punishment in the event of breach of contract. 

Just as cooperative competence is not always socially beneficial, depending on who gains this competence, so is coercive competence not always socially harmful. Many believe it to be beneficial for responsible parents to have ``coercive capabilities'', such as being able to physically restrain their children from running out on the road. Similarly, it is often regarded as a requisite of a functional state for the state to possess a monopoly of violence over sub-state actors. %

Finally, one of the greatest drivers of cooperative competence has been inter-group competition. Competition facilitates learning %
by providing a smooth, motivating, and scalable curriculum \cite{Leibo2019AutocurriculaAT}. The major transitions in biological evolution and cultural evolution, which can be understood as achievements of cooperation, have all plausibly been driven by inter-group competition. Thus it may be that a valuable way to learn cooperative skills is to expose agents to strong inter-group competitive pressures.
In so doing, it may be hard to differentially train cooperative skill without also training skill in coercion and competition. %

\subsection{Understanding and Mitigating the Downsides}

In summary, there are potential downsides to developments in Cooperative AI. Acknowledging and studying these issues can help guide future research in ways that maximize positive impact and mitigate risks.

When are increases in cooperative competence socially beneficial? When do the exclusion effects or correlated increases in coercive competence outweigh the benefits of increases in cooperative competence?  As a baseline, we offer the \textit{Hypothesis that Broad Cooperative Competence is Beneficial}: large and broadly distributed increases in cooperative competence tend to be, on net, broadly welfare improving. 
We offer two theoretical arguments and an empirical argument for this hypothesis. 

Firstly, the first-order effect of greater cooperation is, by definition, to improve the welfare of those who are cooperating. It is thus only the second-order effects wherein exclusionary harms arise. It seems plausible that with broadly distributed gains in cooperative competence the positive first-order effects will, in aggregate, often dominate adverse second-order effects.
Of course, the strength of this argument will clearly depend on the social context and nature of the increase in cooperative competence; we regard investigating this as an important open question in the science of cooperation. 
Secondly, mutual gains in coercive capabilities tend to cancel each other out, like how mutual training in chess will tend to not induce a large shift in the balance of skill. To the extent, then, that research on Cooperative AI unavoidably also increases coercive skill, the hope is that those adverse impacts will largely cancel, whereas the increases in cooperative competence will be additive, if not positive complements. This argument is most true for coercive capabilities that are not destructive but merely lead to transfers of wealth between agents. Nevertheless, mutual increases in destructive coercive capabilities will also often cancel each other out through deterrence. The world has not experienced more destruction with the advent of nuclear weapons, because leaders possessing nuclear weapons have greatly moderated their aggression against each other. By contrast, cooperation and cooperative capabilities lead to positive feedback and are reinforcing; it is in one's interests to help others learn to be better cooperators. 

Finally, plausibly as a consequence of the above, historic examples of larger scale cooperative structures seem to have been more effective than smaller parasitic or rivalrous ones. The ``major transitions'' in evolution detail the systematic increase in complexity and functional differentiation in biological evolution: prokaryotes to eukaryotes, protists to multicellular organisms, individuals to colonies, primates to human society. Transitions in cultural evolution show the same trend: tribes to cities, to territorial states, to larger and more cohesive states, to globalization. %
Thus, plausibly, enhanced cooperative capabilities will on net favor larger scale, more inclusive, and broadly beneficial cooperation. 

Are there general insights about when increases in cooperative competence are most likely to have positive impacts on welfare? Might they depend on distributions of power and cooperative competence? By working with the fields of economics, governance, and institutional design, can we develop a general theory of when restrictions on certain kinds of cooperation are most socially beneficial or harmful? 

Can we identify capabilities which are disproportionately useful for (pareto-improving) cooperation, as opposed to coercion? For example, the development of skills for cheap-talk communication are plausibly cooperation-biased, since an agent can  choose to ignore a cheap-talk channel if it is on net not rewarding \cite{jaques2019social}.
Other advances in communication may be especially useful for honest revelation and not for deception, such as trusted mediators, reputation systems, trusted hardware that can verify observations, or norms against lying. 

For commitment, perhaps \textit{multilateral} commitment mechanisms, such as legal contracts, are cooperatively biased relative to unilateral commitment mechanisms?
Can we build test environments for evaluating the coercive disposition of AI agents, and decide how AIs should behave with respect to deception and threats? 
Even if competition is useful for learning cooperative capabilities, are there ways of separating the gains in cooperative and coercive capabilities and of prioritizing instilling our AIs with the former?

\section{Conclusion}
\label{section:conclusion}

Cooperation was important to the major transitions in evolution, has been foundational to the human story, and remains critical for human well-being. Problems of cooperation are also complex and hard, and they seem to scale in difficulty with the magnitude of our cooperative achievements. Cooperation is thus an attractive target for research on intelligence. 

The field of artificial intelligence has much to contribute to this research frontier. Advances in AI are providing new scientific tools for understanding social systems and for devising novel cooperative structures. Developments in AI are themselves being deployed in society as tools, infrastructure, and agents; it is imperative that this deployment be done in a way that promotes human cooperation. 

Due to the wide-ranging and deep implications that cooperation has for the human condition, research on and knowledge about cooperation are dispersed across a great number of different disciplines in the natural, engineering, and social sciences. Crucial fields include biology, sociology, social psychology, anthropology, economics, history, international relations, and computer science. As a consequence, many of the open problems raised in this article arise at the intersection of AI with those other fields. For research in Cooperative AI to succeed, it will therefore be necessary to bridge the gaps between disciplines, develop a common vocabulary on problems of cooperation, and agree on goals that can be pursued and achieved in cooperation.

As the field of AI takes increasingly confident strides in its ambition to build intelligent machine agents, it is critical to attend to the kinds of intelligence humanity most needs. Necessarily among these is cooperative intelligence.

\section{Acknowledgments}
For valuable input and discussion, the authors would like to thank Markus Anderljung, Asya Bergal, Matt Botvinik, Ryan Carey, Andrew Critch, Owen Cotton-Barratt, Nick Bostrom, Owain Evans, Ulrike Franke, Ben Garfinkel, Gillian Hadfield, Eric Horvitz, Charlotte Jander, Shane Legg, S\"{o}ren Mindermann, Luke Muehlhauser, Rohin Shah, Toby Shevlane, Peter Stone, Robert Trager, Aaron Tucker, Laura Weidinger, and especially Jan Leike, Chris Summerfield, and Toby Ord. The project also benefited from thoughtful feedback from researchers across DeepMind, and specifically the Multi-Agent team, as well as from seminars at the Centre for the Governance of AI, Future of Humanity Institute, University of Oxford. We would also like to thank Alex Lintz for excellent research assistance; Julia Cohen, Charlotte Smith, and Aliya Ahmad at DeepMind for their support; and Wes Cowley for copy editing.

 \pagebreak
 
\section{Appendix A: Cooperative AI Workshop - NeurIPS 2020}
\label{section:appendixA}

The following statement of the intellectual agenda of a 2020 NeurIPS workshop was published September 1 2020, at \href{http://www.cooperativeAI.com}{cooperativeAI.com}%
. It was circulated to potential co-organizers from June 4 2020. \\

\textbf{Aims and Focus}\\
Problems of cooperation—in which agents seek ways to jointly improve their welfare—are ubiquitous and important. They can be found at all scales ranging from our daily routines—such as highway driving, communication via shared language, division of labor, and work collaborations—to our global challenges—such as disarmament, climate change, global commerce, and pandemic preparedness. Arguably, the success of the human species is rooted in our ability to cooperate, in our social intelligence and skills. Since machines powered by artificial intelligence and machine learning are playing an ever greater role in our lives, it will be important to equip them with the skills necessary to cooperate and to foster cooperation.

We see an opportunity for the field of AI, and particularly machine learning, to explicitly focus effort on this class of problems which we term Cooperative AI. The goal of this research would be to study the many aspects of the problem of cooperation, and innovate in AI to contribute to solving these problems. Central questions include how to build machine agents with the capabilities needed for cooperation, and how advances in machine learning can help foster cooperation in populations of agents (of machines and/or humans), such as through improved mechanism design and mediation.

Such research could be organized around key capabilities necessary for cooperation, including: understanding other agents, communicating with other agents, constructing cooperative commitments, and devising and negotiating suitable bargains and institutions. In the context of machine learning, it will be important to develop training environments, tasks, and domains in which cooperative skills are crucial to success, learnable, and non-trivial. Work on the fundamental question of cooperation is by necessity interdisciplinary and will draw on a range of fields, including reinforcement learning (and inverse RL), multi-agent systems, game theory, mechanism design, social choice, language learning, and interpretability. This research may even touch upon fields like trusted hardware design and cryptography to address problems in commitment and communication.

Since artificial agents will often act on behalf of particular humans and in ways that are consequential for humans, this research will need to consider how machines can adequately learn human preferences, and how best to integrate human norms and ethics into cooperative arrangements. Research should also study the potential downsides of cooperative skills—such as exclusion, collusion, and coercion—and how to channel cooperative skills to most improve human welfare. Overall, this research would connect machine learning research to the broader scientific enterprise, in the natural sciences and social sciences, studying the problem of cooperation, and to the broader social effort to solve coordination problems.

We are planning to bring together scholars from diverse backgrounds to discuss how AI research can contribute to the field of cooperation.

\pagebreak

\bibliography{main.bib}

\newcommand{\etalchar}[1]{$^{#1}$}
\begin{thebibliography}{{\AA}vdHTW09}

\bibitem[ACN10]{abbeel2010autonomous}
Pieter Abbeel, Adam Coates, and Andrew~Y. Ng.
\newblock Autonomous helicopter aerobatics through apprenticeship learning.
\newblock {\em The International Journal of Robotics Research},
  29(13):1608--1639, 2010.
\newblock \href {http://dx.doi.org/10.1177/0278364910371999}
  {\path{doi:10.1177/0278364910371999}}.

\bibitem[AET{\etalchar{+}}20]{anthony2020learning}
Thomas Anthony, Tom Eccles, Andrea Tacchetti, János Kramár, Ian Gemp,
  Thomas~C. Hudson, Nicolas Porcel, Marc Lanctot, Julien Pérolat, Richard
  Everett, Roman Werpachowski, Satinder Singh, Thore Graepel, and Yoram
  Bachrach.
\newblock Learning to play no-press diplomacy with best response policy
  iteration.
\newblock preprint, 2020.
\newblock \href {http://arxiv.org/abs/2006.04635} {\path{arXiv:2006.04635}}.

\bibitem[AG18]{abebe2018mechanism}
Rediet Abebe and Kira Goldner.
\newblock Mechanism design for social good.
\newblock {\em AI Matters}, 4(3):27--34, 2018.
\newblock \href {http://dx.doi.org/10.1145/3284751.3284761}
  {\path{doi:10.1145/3284751.3284761}}.

\bibitem[AH81]{axelrod1981evolution}
Robert Axelrod and William~Donald Hamilton.
\newblock The evolution of cooperation.
\newblock {\em Science}, 211(4489):1390--1396, 1981.
\newblock \href {http://dx.doi.org/10.1126/science.7466396}
  {\path{doi:10.1126/science.7466396}}.

\bibitem[AHM19]{DBLP:journals/corr/abs-1902-03185}
Nicolas Anastassacos, Steve Hailes, and Mirco Musolesi.
\newblock Understanding the impact of partner choice on cooperation and social
  norms by means of multi-agent reinforcement learning.
\newblock preprint, 2019.
\newblock \href {http://arxiv.org/abs/1902.03185} {\path{arXiv:1902.03185}}.

\bibitem[AHV03]{andreoni2003carrot}
James Andreoni, William Harbaugh, and Lise Vesterlund.
\newblock The carrot or the stick: Rewards, punishments, and cooperation.
\newblock {\em American Economic Review}, 93(3):893--902, 2003.
\newblock \href {http://dx.doi.org/10.1257/000282803322157142}
  {\path{doi:10.1257/000282803322157142}}.

\bibitem[AJB06]{argente2006multi}
Estefania Argente, Vicente Julian, and Vicente Botti.
\newblock Multi-agent system development based on organizations.
\newblock {\em Electronic Notes in Theoretical Computer Science},
  150(3):55--71, 2006.
\newblock \href {http://dx.doi.org/10.1016/j.entcs.2006.03.005}
  {\path{doi:10.1016/j.entcs.2006.03.005}}.

\bibitem[AJS17]{amin2017repeated}
Kareem Amin, Nan Jiang, and Satinder Singh.
\newblock Repeated inverse reinforcement learning.
\newblock In {\em Advances in Neural Information Processing Systems},
  volume~30, pages 1815--1824, 2017.
\newblock URL:
  \url{https://proceedings.neurips.cc/paper/2017/file/8ce6790cc6a94e65f17f908f462fae85-Paper.pdf}.

\bibitem[AKK03]{alonso2003adaptive}
Eduardo Alonso, Daniel Kudenko, and Dimitar Kazakov, editors.
\newblock {\em Adaptive agents and multi-agent systems: adaptation and
  multi-agent learning}.
\newblock Springer, Berlin, 2003.

\bibitem[AL19]{armstrong2019machine}
Ben Armstrong and Kate Larson.
\newblock Machine learning to strengthen democracy.
\newblock In {\em NeurIPS Joint Workshop on AI for Social Good}, 2019.
\newblock URL:
  \url{https://aiforsocialgood.github.io/neurips2019/accepted/track1/pdfs/69_aisg_neurips2019.pdf}.

\bibitem[AM16]{aziz2016discrete}
Haris Aziz and Simon Mackenzie.
\newblock A discrete and bounded envy-free cake cutting protocol for any number
  of agents.
\newblock In {\em IEEE 57th Annual Symposium on Foundations of Computer Science
  (FOCS)}, pages 416--427. IEEE, 2016.
\newblock \href {http://dx.doi.org/10.1109/FOCS.2016.52}
  {\path{doi:10.1109/FOCS.2016.52}}.

\bibitem[AM18]{armstrong2018occam}
Stuart Armstrong and S{\"o}ren Mindermann.
\newblock Occam's razor is insufficient to infer the preferences of irrational
  agents.
\newblock In {\em Advances in Neural Information Processing Systems},
  volume~31, pages 5598--5609, 2018.
\newblock URL:
  \url{https://proceedings.neurips.cc/paper/2018/file/d89a66c7c80a29b1bdbab0f2a1a94af8-Paper.pdf}.

\bibitem[AO18]{AMODU2018255}
Oluwatosin~Ahmed Amodu and Mohamed Othman.
\newblock Machine-to-machine communication: An overview of opportunities.
\newblock {\em Computer Networks}, 145:255--276, 2018.
\newblock \href
  {http://dx.doi.org/https://doi.org/10.1016/j.comnet.2018.09.001}
  {\path{doi:https://doi.org/10.1016/j.comnet.2018.09.001}}.

\bibitem[AOS{\etalchar{+}}16]{amodei2016concrete}
Dario Amodei, Chris Olah, Jacob Steinhardt, Paul Christiano, John Schulman, and
  Dan Man{\'e}.
\newblock Concrete problems in {AI} safety.
\newblock preprint, 2016.
\newblock \href {http://arxiv.org/abs/1606.06565} {\path{arXiv:1606.06565}}.

\bibitem[AP01]{artikis2001formal}
Alexander Artikis and Jeremy Pitt.
\newblock A formal model of open agent societies.
\newblock In {\em Proceedings of the Fifth International Conference on
  Autonomous Agents}, pages 192--193, 2001.
\newblock \href {http://dx.doi.org/10.1145/375735.376108}
  {\path{doi:10.1145/375735.376108}}.

\bibitem[APP{\etalchar{+}}02]{arai2002advances}
Tamio Arai, Enrico Pagello, Lynne~E Parker, et~al.
\newblock Advances in multi-robot systems.
\newblock {\em IEEE Transactions on robotics and automation}, 18(5):655--661,
  2002.
\newblock \href {http://dx.doi.org/10.1109/TRA.2002.806024}
  {\path{doi:10.1109/TRA.2002.806024}}.

\bibitem[APS04]{aknine2004extended}
Samir Aknine, Suzanne Pinson, and Melvin~F Shakun.
\newblock An extended multi-agent negotiation protocol.
\newblock {\em Autonomous Agents and Multi-Agent Systems}, 8(1):5--45, 2004.
\newblock \href {http://dx.doi.org/0.1023/B:AGNT.0000009409.19387.f8}
  {\path{doi:0.1023/B:AGNT.0000009409.19387.f8}}.

\bibitem[AR12]{acemoglu2012nations}
Daron Acemoglu and James~A Robinson.
\newblock {\em Why nations fail: The origins of power, prosperity, and
  poverty}.
\newblock Currency, New York, 2012.

\bibitem[Arr51]{arrow1951social}
Kenneth~J Arrow.
\newblock {\em Social choice and individual values}, volume~12 of {\em Cowles
  Foundation Monograph Series}.
\newblock Yale University Press, New Haven, 1951.

\bibitem[AS18]{albrecht2018autonomous}
Stefano~V Albrecht and Peter Stone.
\newblock Autonomous agents modelling other agents: A comprehensive survey and
  open problems.
\newblock {\em Artificial Intelligence}, 258:66--95, 2018.
\newblock \href {http://dx.doi.org/10.1016/j.artint.2018.01.002}
  {\path{doi:10.1016/j.artint.2018.01.002}}.

\bibitem[AS19]{APICELLA2019R447}
Coren~L. Apicella and Joan~B. Silk.
\newblock The evolution of human cooperation.
\newblock {\em Current Biology}, 29(11):R447--R450, 2019.
\newblock \href {http://dx.doi.org/10.1016/j.cub.2019.03.036}
  {\path{doi:10.1016/j.cub.2019.03.036}}.

\bibitem[ASB98]{austen1998social}
David Austen-Smith and Jeffrey~S Banks.
\newblock Social choice theory, game theory, and positive political theory.
\newblock {\em Annual Review of Political Science}, 1:259--287, 1998.
\newblock \href {http://dx.doi.org/10.1146/annurev.polisci.1.1.259}
  {\path{doi:10.1146/annurev.polisci.1.1.259}}.

\bibitem[ASBB{\etalchar{+}}17]{al2017continuous}
Maruan Al-Shedivat, Trapit Bansal, Yuri Burda, Ilya Sutskever, Igor Mordatch,
  and Pieter Abbeel.
\newblock Continuous adaptation via meta-learning in nonstationary and
  competitive environments.
\newblock preprint, 2017.
\newblock \href {http://arxiv.org/abs/1710.03641} {\path{arXiv:1710.03641}}.

\bibitem[ASP09]{artikis2009specifying}
Alexander Artikis, Marek Sergot, and Jeremy Pitt.
\newblock Specifying norm-governed computational societies.
\newblock {\em ACM Transactions on Computational Logic}, 10(1):1--42, 2009.
\newblock \href {http://dx.doi.org/10.1145/1459010.1459011}
  {\path{doi:10.1145/1459010.1459011}}.

\bibitem[ASS12]{akrour2012april}
Riad Akrour, Marc Schoenauer, and Mich{\`e}le Sebag.
\newblock {APRIL}: Active preference learning-based reinforcement learning.
\newblock In {\em Joint European Conference on Machine Learning and Knowledge
  Discovery in Databases}, pages 116--131, 2012.
\newblock \href {http://dx.doi.org/10.1007/978-3-642-33486-3_8}
  {\path{doi:10.1007/978-3-642-33486-3_8}}.

\bibitem[ASS15]{amir2015multi}
Ofra Amir, Guni Sharon, and Roni Stern.
\newblock Multi-agent pathfinding as a combinatorial auction.
\newblock In {\em Proceedings of the AAAI Conference on Artificial
  Intelligence}, pages 2003--2009, 2015.

\bibitem[Aum74]{AUMANN197467}
Robert~J. Aumann.
\newblock Subjectivity and correlation in randomized strategies.
\newblock {\em Journal of Mathematical Economics}, 1(1):67 -- 96, 1974.
\newblock \href {http://dx.doi.org/10.1016/0304-4068(74)90037-8}
  {\path{doi:10.1016/0304-4068(74)90037-8}}.

\bibitem[{\AA}vdHTW09]{aagotnes2009power}
Thomas {\AA}gotnes, Wiebe van~der Hoek, Moshe Tennenholtz, and Michael
  Wooldridge.
\newblock Power in normative systems.
\newblock In {\em Proceedings of The 8th International Conference on Autonomous
  Agents and Multiagent Systems-Volume 1}, pages 145--152, 2009.
\newblock \href {http://dx.doi.org/10.5555/1558013.1558033}
  {\path{doi:10.5555/1558013.1558033}}.

\bibitem[{\AA}vdHW07]{aagotnes2007normative}
Thomas {\AA}gotnes, Wiebe van~der Hoek, and Michael Wooldridge.
\newblock Normative system games.
\newblock In {\em Proceedings of the 6th International Joint Conference on
  Autonomous Agents and Multiagent Systems}, pages 1--8, 2007.
\newblock \href {http://dx.doi.org/10.1145/1329125.1329284}
  {\path{doi:10.1145/1329125.1329284}}.

\bibitem[Axe84]{axelrod1984evolution}
Robert Axelrod.
\newblock {\em The evolution of cooperation}.
\newblock Basic Books, New York, 1984.

\bibitem[Bag95]{bagwell1995commitment}
Kyle Bagwell.
\newblock Commitment and observability in games.
\newblock {\em Games and Economic Behavior}, 8(2):271--280, 1995.
\newblock \href {http://dx.doi.org/10.1016/S0899-8256(05)80001-6}
  {\path{doi:10.1016/S0899-8256(05)80001-6}}.

\bibitem[Bak20]{baker2020emergent}
Bowen Baker.
\newblock Emergent reciprocity and team formation from randomized uncertain
  social preferences.
\newblock {\em Advances in Neural Information Processing Systems}, 33, 2020.
\newblock \href {http://arxiv.org/abs/2011.05373} {\path{arXiv:2011.05373}}.

\bibitem[Ban77]{bandura1977social}
A.~Bandura.
\newblock {\em Social learning theory}.
\newblock Prentice Hall, Englewood Cliffs, NJ, 1977.

\bibitem[Bar13]{barclay2013strategies}
Pat Barclay.
\newblock Strategies for cooperation in biological markets, especially for
  humans.
\newblock {\em Evolution and Human Behavior}, 34(3):164--175, 2013.
\newblock \href {http://dx.doi.org/10.1016/j.evolhumbehav.2013.02.002}
  {\path{doi:10.1016/j.evolhumbehav.2013.02.002}}.

\bibitem[BB20]{brownbakhtin2020}
Noam Brown and Anton Bakhtin.
\newblock {ReBeL}: A general game-playing ai bot that excels at poker and more.
\newblock blog, Dec 2020.
\newblock URL:
  \url{https://ai.facebook.com/blog/rebel-a-general-game-playing-ai-bot-that-excels-at-poker-and-more/}.

\bibitem[BBC{\etalchar{+}}19]{berner2019dota}
Christopher Berner, Greg Brockman, Brooke Chan, Vicki Cheung, Przemys{\l}aw
  D{\k{e}}biak, Christy Dennison, David Farhi, Quirin Fischer, Shariq Hashme,
  Chris Hesse, et~al.
\newblock Dota 2 with large scale deep reinforcement learning.
\newblock preprint, 2019.
\newblock \href {http://arxiv.org/abs/1912.06680} {\path{arXiv:1912.06680}}.

\bibitem[BBDS08]{busoniu2008comprehensive}
Lucian Busoniu, Robert Babuska, and Bart De~Schutter.
\newblock A comprehensive survey of multiagent reinforcement learning.
\newblock {\em IEEE Transactions on Systems, Man, and Cybernetics, Part C
  (Applications and Reviews)}, 38(2):156--172, 2008.
\newblock \href {http://dx.doi.org/10.1109/TSMCC.2007.913919}
  {\path{doi:10.1109/TSMCC.2007.913919}}.

\bibitem[BCA{\etalchar{+}}17]{Beaton2017Extravehicular}
Kara~H. Beaton, Steven~P. Chappell, Andrew F.~J. Abercromby, Matthew~J. Miller,
  Shannon~Kobs Nawotniak, Scott~S. Hughes, Allyson Brady, and Darlene S.~S.
  Lim.
\newblock Extravehicular activity operations concepts under communication
  latency and bandwidth constraints.
\newblock {\em IEEE Aerospace Conference}, pages 1--20, 2017.
\newblock \href {http://dx.doi.org/10.1109/AERO.2017.7943570}
  {\path{doi:10.1109/AERO.2017.7943570}}.

\bibitem[BCE12]{brandt2012computational}
Felix Brandt, Vincent Conitzer, and Ulle Endriss.
\newblock Computational social choice.
\newblock In Gerhard Weiss, editor, {\em Multiagent Systems}, pages 213--283.
  MIT Press, Cambridge, MA, 2012.

\bibitem[BCE{\etalchar{+}}16]{brandt2016handbook}
Felix Brandt, Vincent Conitzer, Ulle Endriss, J{\'e}r{\^o}me Lang, and Ariel~D
  Procaccia, editors.
\newblock {\em Handbook of computational social choice}.
\newblock Cambridge University Press, New York, 2016.

\bibitem[BCF{\etalchar{+}}14]{barasz2014robust}
Mihaly Barasz, Paul Christiano, Benja Fallenstein, Marcello Herreshoff, Patrick
  LaVictoire, and Eliezer Yudkowsky.
\newblock Robust cooperation in the prisoner's dilemma: Program equilibrium via
  provability logic.
\newblock preprint, 2014.
\newblock \href {http://arxiv.org/abs/1401.5577} {\path{arXiv:1401.5577}}.

\bibitem[BCG07]{bellifemine2007developing}
Fabio~Luigi Bellifemine, Giovanni Caire, and Dominic Greenwood.
\newblock {\em Developing multi-agent systems with JADE}.
\newblock Number~7 in Wiley Series in Agent Technology. John Wiley \& Sons,
  West Sussex, 2007.

\bibitem[BCLF85]{baron1985does}
Simon Baron-Cohen, Alan~M Leslie, and Uta Frith.
\newblock Does the autistic child have a “theory of mind”.
\newblock {\em Cognition}, 21(1):37--46, 1985.
\newblock \href {http://dx.doi.org/10.1016/0010-0277(85)90022-8}
  {\path{doi:10.1016/0010-0277(85)90022-8}}.

\bibitem[BCM17]{bench2017norms}
Trevor Bench-Capon and Sanjay Modgil.
\newblock Norms and value based reasoning: justifying compliance and violation.
\newblock {\em Artificial Intelligence and Law}, 25:29--64, 2017.
\newblock \href {http://dx.doi.org/10.1007/s10506-017-9194-9}
  {\path{doi:10.1007/s10506-017-9194-9}}.

\bibitem[BDDS09]{bordini2009multi}
Rafael~H Bordini, Mehdi Dastani, J{\"u}rgen Dix, and Amal El~Fallah
  Seghrouchni, editors.
\newblock {\em Multi-Agent Programming}.
\newblock Springer, New York, 2009.

\bibitem[BDVG{\etalchar{+}}19]{AIandCooperation}
Elisa Bertino, Finale Doshi-Velez, Maria Gini, Daniel Lopresti, and David
  Parkes.
\newblock Artificial intelligence \& cooperation.
\newblock Technical report, Computing Community Consortium, November 2019.
\newblock URL:
  \url{https://cra.org/ccc/resources/ccc-led-whitepapers/#2020-quadrennial-papers}.

\bibitem[Bec98]{beck1998learning}
Joseph Beck.
\newblock Learning to teach with a reinforcement learning agent.
\newblock In {\em Proceedings of the Fifteenth National/Tenth Conference on
  Artificial Intelligence/Innovative Applications of Artificial Intelligence},
  page 1185, 1998.

\bibitem[BFC{\etalchar{+}}20]{bard2020hanabi}
Nolan Bard, Jakob~N Foerster, Sarath Chandar, Neil Burch, Marc Lanctot,
  H~Francis Song, Emilio Parisotto, Vincent Dumoulin, Subhodeep Moitra, Edward
  Hughes, et~al.
\newblock The {H}anabi challenge: A new frontier for {AI} research.
\newblock {\em Artificial Intelligence}, 280:103216, 2020.
\newblock \href {http://dx.doi.org/10.1016/j.artint.2019.103216}
  {\path{doi:10.1016/j.artint.2019.103216}}.

\bibitem[BFG{\etalchar{+}}13]{baarslag2013evaluating}
Tim Baarslag, Katsuhide Fujita, Enrico~H Gerding, Koen Hindriks, Takayuki Ito,
  Nicholas~R Jennings, Catholijn Jonker, Sarit Kraus, Raz Lin, Valentin Robu,
  and Colin~R Williams.
\newblock Evaluating practical negotiating agents: Results and analysis of the
  2011 international competition.
\newblock {\em Artificial Intelligence}, 198:73--103, 2013.
\newblock \href {http://dx.doi.org/10.1016/j.artint.2012.09.004}
  {\path{doi:10.1016/j.artint.2012.09.004}}.

\bibitem[BFJ11]{bradshaw2011human}
Jeffrey~M Bradshaw, Paul~J Feltovich, and Matthew Johnson.
\newblock Human-agent interaction.
\newblock In {\em Handbook of Human-Machine Interaction}, pages 283--302.
  Ashgate, Surrey, 2011.

\bibitem[BG14]{bond2014readings}
Alan~H Bond and Les Gasser, editors.
\newblock {\em Readings in Distributed Artificial Intelligence}.
\newblock Morgan Kaufmann, San Mateo, CA, 2014.

\bibitem[BGB10]{boyd2010coordinated}
Robert Boyd, Herbert Gintis, and Samuel Bowles.
\newblock Coordinated punishment of defectors sustains cooperation and can
  proliferate when rare.
\newblock {\em Science}, 328(5978):617--620, 2010.
\newblock \href {http://dx.doi.org/10.1126/science.1183665}
  {\path{doi:10.1126/science.1183665}}.

\bibitem[BGNN19]{brown2019extrapolating}
Daniel Brown, Wonjoon Goo, Prabhat Nagarajan, and Scott Niekum.
\newblock Extrapolating beyond suboptimal demonstrations via inverse
  reinforcement learning from observations.
\newblock In {\em Proceedings of the 36th International Conference on Machine
  Learning}, pages 783--792, 2019.

\bibitem[BHL{\etalchar{+}}19a]{bahdanau2018learning}
Dzmitry Bahdanau, Felix Hill, Jan Leike, Edward Hughes, Arian Hosseini,
  Pushmeet Kohli, and Edward Grefenstette.
\newblock Learning to understand goal specifications by modelling reward.
\newblock In {\em International Conference on Learning Representations}, 2019.
\newblock URL: \url{https://openreview.net/forum?id=H1xsSjC9Ym}.

\bibitem[BHL19b]{brynjolfsson2019does}
Erik Brynjolfsson, Xiang Hui, and Meng Liu.
\newblock Does machine translation affect international trade? evidence from a
  large digital platform.
\newblock {\em Management Science}, 65(12):5449--5456, 2019.
\newblock \href {http://dx.doi.org/10.1287/mnsc.2019.3388}
  {\path{doi:10.1287/mnsc.2019.3388}}.

\bibitem[Bic06]{bicchieri2006grammar}
Cristina Bicchieri.
\newblock {\em The grammar of society: The nature and dynamics of social
  norms}.
\newblock Cambridge University Press, Cambridge, UK, 2006.

\bibitem[BITT92]{bartholdi1992hard}
John~J Bartholdi~III, Craig~A Tovey, and Michael~A Trick.
\newblock How hard is it to control an election?
\newblock {\em Mathematical and Computer Modelling}, 16(8-9):27--40, 1992.
\newblock \href {http://dx.doi.org/10.1016/0895-7177(92)90085-Y}
  {\path{doi:10.1016/0895-7177(92)90085-Y}}.

\bibitem[BK17]{bohn2017}
Manuel Bohn and Bahar Köymen.
\newblock Common ground and development.
\newblock {\em Child Development Perspectives}, 12:104--108, 2017.
\newblock \href {http://dx.doi.org/10.1111/cdep.12269}
  {\path{doi:10.1111/cdep.12269}}.

\bibitem[BKM{\etalchar{+}}19]{baker2019emergent}
Bowen Baker, Ingmar Kanitscheider, Todor Markov, Yi~Wu, Glenn Powell, Bob
  McGrew, and Igor Mordatch.
\newblock Emergent tool use from multi-agent autocurricula.
\newblock preprint, 2019.
\newblock \href {http://arxiv.org/abs/1909.07528} {\path{arXiv:1909.07528}}.

\bibitem[BKN10]{bryan2010commitment}
Gharad Bryan, Dean Karlan, and Scott Nelson.
\newblock Commitment devices.
\newblock {\em Annual Review of Economics}, 2:671--698, 2010.
\newblock \href {http://dx.doi.org/10.1146/annurev.economics.102308.124324}
  {\path{doi:10.1146/annurev.economics.102308.124324}}.

\bibitem[BKS03]{bichler2003towards}
Martin Bichler, Gregory Kersten, and Stefan Strecker.
\newblock Towards a structured design of electronic negotiations.
\newblock {\em Group Decision and Negotiation}, 12(4):311--335, 2003.
\newblock \href {http://dx.doi.org/10.1023/A:1024867820235}
  {\path{doi:10.1023/A:1024867820235}}.

\bibitem[Bla58]{black1958theory}
Duncan Black.
\newblock {\em The theory of committees and elections}.
\newblock Springer, 1958.

\bibitem[Bla03]{blake2003}
M.~Brian Blake.
\newblock Coordinating multiple agents for workflow-oriented process
  orchestration.
\newblock {\em Information Systems and e-Business Management}, 1:387--404, 11
  2003.
\newblock \href {http://dx.doi.org/10.1007/s10257-003-0023-1}
  {\path{doi:10.1007/s10257-003-0023-1}}.

\bibitem[BMF{\etalchar{+}}00]{burgard2000collaborative}
Wolfram Burgard, Mark Moors, Dieter Fox, Reid Simmons, and Sebastian Thrun.
\newblock Collaborative multi-robot exploration.
\newblock In {\em Proceedings 2000 ICRA. Millennium Conference. IEEE
  International Conference on Robotics and Automation. Symposia Proceedings},
  volume~1, pages 476--481. IEEE, 2000.
\newblock \href {http://dx.doi.org/10.1109/ROBOT.2000.844100}
  {\path{doi:10.1109/ROBOT.2000.844100}}.

\bibitem[BMR{\etalchar{+}}20]{brown2020language}
Tom~B. Brown, Benjamin Mann, Nick Ryder, Melanie Subbiah, Jared Kaplan,
  Prafulla Dhariwal, Arvind Neelakantan, Pranav Shyam, Girish Sastry, Amanda
  Askell, Sandhini Agarwal, Ariel Herbert-Voss, Gretchen Krueger, Tom Henighan,
  Rewon Child, Aditya Ramesh, Daniel~M. Ziegler, Jeffrey Wu, Clemens Winter,
  Christopher Hesse, Mark Chen, Eric Sigler, Mateusz Litwin, Scott Gray,
  Benjamin Chess, Jack Clark, Christopher Berner, Sam McCandlish, Alec Radford,
  Ilya Sutskever, and Dario Amodei.
\newblock Language models are few-shot learners.
\newblock preprint, 2020.
\newblock \href {http://arxiv.org/abs/2005.14165} {\path{arXiv:2005.14165}}.

\bibitem[BMS18]{sep-social-norms}
Cristina Bicchieri, Ryan Muldoon, and Alessandro Sontuoso.
\newblock Social norms.
\newblock In Edward~N. Zalta, editor, {\em The Stanford Encyclopedia of
  Philosophy}. Metaphysics Research Lab, Stanford University, {W}inter 2018
  edition, 2018.
\newblock URL:
  \url{https://plato.stanford.edu/archives/win2018/entries/social-norms/}.

\bibitem[Bos14]{bostrom2014superintelligence}
Nick Bostrom.
\newblock {\em Superintelligence}.
\newblock Oxford University Press, Oxford, 2014.

\bibitem[Bow05]{bowling2005convergence}
Michael Bowling.
\newblock Convergence and no-regret in multiagent learning.
\newblock In {\em Advances in Neural Information Processing Systems},
  volume~17, pages 209--216, 2005.
\newblock URL:
  \url{https://proceedings.neurips.cc/paper/2004/file/88fee0421317424e4469f33a48f50cb0-Paper.pdf}.

\bibitem[BP17]{bartoletti2017empirical}
Massimo Bartoletti and Livio Pompianu.
\newblock An empirical analysis of smart contracts: platforms, applications,
  and design patterns.
\newblock In {\em International Conference on Financial Cryptography and Data
  Security}, pages 494--509. Springer, 2017.
\newblock \href {http://dx.doi.org/10.1007/978-3-319-70278-0_31}
  {\path{doi:10.1007/978-3-319-70278-0_31}}.

\bibitem[BPMP17]{DBLP:journals/corr/BorsaPMP17}
Diana Borsa, Bilal Piot, R{\'{e}}mi Munos, and Olivier Pietquin.
\newblock Observational learning by reinforcement learning.
\newblock preprint, 2017.
\newblock \href {http://arxiv.org/abs/1706.06617} {\path{arXiv:1706.06617}}.

\bibitem[BR16]{boutilierincomplete}
Craig Boutilier and Jeffrey~S. Rosenschein.
\newblock Incomplete information with communication in voting.
\newblock In Felix Brandt, Vincent Conitzer, Ulle Endriss, J\'{e}r\^{o}me Lang,
  and Ariel~D. Procaccia, editors, {\em Handbook on Computational Social
  Choice}, chapter~10, pages 223--257. Cambridge University Press, New York,
  2016.

\bibitem[Bra87]{bratman1987intention}
Michael Bratman.
\newblock {\em Intention, plans, and practical reason}.
\newblock Number~10 in The David Hume Series. Harvard University Press,
  Cambridge, MA, 1987.

\bibitem[BS13]{bajaj2013trusteddb}
Sumeet Bajaj and Radu Sion.
\newblock Trusted{DB}: A trusted hardware-based database with privacy and data
  confidentiality.
\newblock {\em IEEE Transactions on Knowledge and Data Engineering},
  26(3):752--765, 2013.
\newblock \href {http://dx.doi.org/10.1109/TKDE.2013.38}
  {\path{doi:10.1109/TKDE.2013.38}}.

\bibitem[BS19]{brown2019superhuman}
Noam Brown and Tuomas Sandholm.
\newblock Superhuman {AI} for multiplayer poker.
\newblock {\em Science}, 365(6456):885--890, 2019.
\newblock \href {http://dx.doi.org/10.1126/science.aay2400}
  {\path{doi:10.1126/science.aay2400}}.

\bibitem[BT96]{brams1996fair}
Steven~J Brams and Alan~D Taylor.
\newblock {\em Fair Division: From cake-cutting to dispute resolution}.
\newblock Cambridge University Press, Cambridge, UK, 1996.

\bibitem[BTT89]{bartholdi1989computational}
John~J Bartholdi, Craig~A Tovey, and Michael~A Trick.
\newblock The computational difficulty of manipulating an election.
\newblock {\em Social Choice and Welfare}, 6:227--241, 1989.
\newblock \href {http://dx.doi.org/10.1007/BF00295861}
  {\path{doi:10.1007/BF00295861}}.

\bibitem[BTU91]{boot1991credible}
Arnoud~WA Boot, Anjan~V Thakor, and Gregory~F Udell.
\newblock Credible commitments, contract enforcement problems and banks:
  Intermediation as credibility assurance.
\newblock {\em Journal of Banking \& Finance}, 15(3):605--632, 1991.
\newblock \href {http://dx.doi.org/10.1016/0378-4266(91)90088-4}
  {\path{doi:10.1016/0378-4266(91)90088-4}}.

\bibitem[Bus87]{buss1987evolution}
Leo~W Buss.
\newblock {\em The evolution of individuality}.
\newblock Princeton University Press, Princeton, NJ, 1987.

\bibitem[BVDTV06]{boella2006introduction}
Guido Boella, Leendert Van Der~Torre, and Harko Verhagen.
\newblock Introduction to normative multiagent systems.
\newblock {\em Computational \& Mathematical Organization Theory}, 12:71--79,
  2006.
\newblock \href {http://dx.doi.org/10.1007/s10588-006-9537-7}
  {\path{doi:10.1007/s10588-006-9537-7}}.

\bibitem[CAB11]{criado2011open}
Natalia Criado, Estefania Argente, and V~Botti.
\newblock Open issues for normative multi-agent systems.
\newblock {\em AI Communications}, 24(3):233--264, 2011.
\newblock URL: \url{10.3233/AIC-2011-0502}.

\bibitem[Cas98]{castelfranchi1998modelling}
Cristiano Castelfranchi.
\newblock Modelling social action for {AI} agents.
\newblock {\em Artificial Intelligence}, 103(1-2):157--182, 1998.
\newblock \href {http://dx.doi.org/10.1016/S0004-3702(98)00056-3}
  {\path{doi:10.1016/S0004-3702(98)00056-3}}.

\bibitem[CB91]{Clark1991-CLAGIC}
Herbert~H. Clark and Susan~E. Brennan.
\newblock Grounding in communication.
\newblock In Lauren~B. Resnick, John~M. Levine, and Stephanie~D. Teasley,
  editors, {\em Perspectives on Socially Shared Cognition}, pages 127--149.
  American Psychological Association, 1991.
\newblock \href {http://dx.doi.org/10.1037/10096-006}
  {\path{doi:10.1037/10096-006}}.

\bibitem[CB98]{claus1998dynamics}
Caroline Claus and Craig Boutilier.
\newblock The dynamics of reinforcement learning in cooperative multiagent
  systems.
\newblock In {\em Proceedings of the Fifteenth National/Tenth Conference on
  Artificial Intelligence/Innovative Applications of Artificial Intelligence},
  pages 746--752, 1998.

\bibitem[CC95]{conte1995understanding}
Rosaria Conte and Cristiano Castelfranchi.
\newblock Understanding the functions of norms in social groups through
  simulation.
\newblock {\em Artificial societies: The computer simulation of social life},
  1:252--267, 1995.

\bibitem[CCB11]{camera2011communication}
Gabriele Camera, Marco Casari, and Maria Bigoni.
\newblock Communication, commitment, and deception in social dilemmas:
  experimental evidence.
\newblock Quaderni - working paper dse no. 751, University of Bologna, 2011.
\newblock \href {http://dx.doi.org/10.2139/ssrn.1854132}
  {\path{doi:10.2139/ssrn.1854132}}.

\bibitem[CCD98]{conte1998autonomous}
Rosaria Conte, Cristiano Castelfranchi, and Frank Dignum.
\newblock Autonomous norm acceptance.
\newblock In {\em International Workshop on Agent Theories, Architectures, and
  Languages}, pages 99--112. Springer, 1998.
\newblock \href {http://dx.doi.org/10.1007/3-540-49057-4_7}
  {\path{doi:10.1007/3-540-49057-4_7}}.

\bibitem[CCDP20]{calvano2020artificial}
Emilio Calvano, Giacomo Calzolari, Vincenzo Denicolo, and Sergio Pastorello.
\newblock Artificial intelligence, algorithmic pricing, and collusion.
\newblock {\em American Economic Review}, 110(10):3267--97, 2020.
\newblock \href {http://dx.doi.org/10.1257/aer.20190623}
  {\path{doi:10.1257/aer.20190623}}.

\bibitem[CD16]{christidis2016blockchains}
Konstantinos Christidis and Michael Devetsikiotis.
\newblock Blockchains and smart contracts for the internet of things.
\newblock {\em {IEEE} Access}, 4:2292--2303, 2016.
\newblock \href {http://dx.doi.org/10.1109/ACCESS.2016.2566339}
  {\path{doi:10.1109/ACCESS.2016.2566339}}.

\bibitem[CDE{\etalchar{+}}06]{chevaleyre2005issues}
Yann Chevaleyre, Paul~E Dunne, Ulle Endriss, J{\'e}r{\^o}me Lang, Michel
  Lema{\^i}tre, Nicolas Maudet, Julian Padget, Steve Phelps, Juan~A
  Rodr{\'\i}gues-Aguilar, and Paulo Sousa.
\newblock Issues in multiagent resource allocation.
\newblock {\em Informatica}, 30(1):3--31, 2006.

\bibitem[CDJT99]{castelfranchi1999deliberative}
Cristiano Castelfranchi, Frank Dignum, Catholijn~M Jonker, and Jan Treur.
\newblock Deliberative normative agents: Principles and architecture.
\newblock In {\em International Workshop on Agent Theories, Architectures, and
  Languages}, pages 364--378. Springer, 1999.
\newblock \href {http://dx.doi.org/10.1007/10719619_27}
  {\path{doi:10.1007/10719619_27}}.

\bibitem[CDW12a]{Cai2012algorithmic}
Yang Cai, Constantinos Daskalakis, and S.~Matthew Weinberg.
\newblock An algorithmic characterization of multi-dimensional mechanisms.
\newblock In {\em Proceedings of the 44th ACM Symposium on Theory of
  Computing}, page 459–478, 2012.
\newblock \href {http://dx.doi.org/10.1145/2213977.2214021}
  {\path{doi:10.1145/2213977.2214021}}.

\bibitem[CDW12b]{Cai2012optimal}
Yang Cai, Constantinos Daskalakis, and S.~Matthew Weinberg.
\newblock Optimal multi-dimensional mechanism design: Reducing revenue to
  welfare maximization.
\newblock In {\em Proceedings of the 53rd IEEE Symposium on Foundations of
  Computer Science}, pages 130--139, 2012.
\newblock \href {http://dx.doi.org/10.1109/FOCS.2012.88}
  {\path{doi:10.1109/FOCS.2012.88}}.

\bibitem[CELM07]{chevaleyre2007short}
Yann Chevaleyre, Ulle Endriss, J{\'e}r{\^o}me Lang, and Nicolas Maudet.
\newblock A short introduction to computational social choice.
\newblock In {\em International Conference on Current Trends in Theory and
  Practice of Computer Science}, pages 51--69. Springer, 2007.
\newblock \href {http://dx.doi.org/10.1007/978-3-540-69507-3_4}
  {\path{doi:10.1007/978-3-540-69507-3_4}}.

\bibitem[CF98]{castelfranchi1998principles}
Cristiano Castelfranchi and Rino Falcone.
\newblock Principles of trust for {MAS}: Cognitive anatomy, social importance,
  and quantification.
\newblock In {\em Proceedings of the International Conference on Multi Agent
  Systems (ICMAS'98)}, pages 72--79. IEEE, 1998.
\newblock \href {http://dx.doi.org/0.1109/ICMAS.1998.699034}
  {\path{doi:0.1109/ICMAS.1998.699034}}.

\bibitem[CH19]{cong2019blockchain}
Lin~William Cong and Zhiguo He.
\newblock Blockchain disruption and smart contracts.
\newblock {\em The Review of Financial Studies}, 32(5):1754--1797, 2019.
\newblock \href {http://dx.doi.org/10.1093/rfs/hhz007}
  {\path{doi:10.1093/rfs/hhz007}}.

\bibitem[CHJH02]{campbell2002deep}
Murray Campbell, A.~Joseph Hoane~Jr, and Feng-hsiung Hsu.
\newblock Deep blue.
\newblock {\em Artificial Intelligence}, 134(1-2):57--83, January 2002.
\newblock \href {http://dx.doi.org/10.1016/S0004-3702(01)00129-1}
  {\path{doi:10.1016/S0004-3702(01)00129-1}}.

\bibitem[Chw13]{chwe2013rational}
Michael Suk-Young Chwe.
\newblock {\em Rational ritual: Culture, coordination, and common knowledge}.
\newblock Princeton University Press, Princeton, NJ, 2013.

\bibitem[CK20]{critch2020ai}
Andrew Critch and David Krueger.
\newblock Ai research considerations for human existential safety ({ARCHES}).
\newblock preprint, 2020.
\newblock \href {http://arxiv.org/abs/2006.04948} {\path{arXiv:2006.04948}}.

\bibitem[CKDB19]{DBLP:journals/corr/abs-1905-12561}
Rahma Chaabouni, Eugene Kharitonov, Emmanuel Dupoux, and Marco Baroni.
\newblock Anti-efficient encoding in emergent communication.
\newblock preprint, 2019.
\newblock \href {http://arxiv.org/abs/1905.12561} {\path{arXiv:1905.12561}}.

\bibitem[CKM{\etalchar{+}}19]{caragiannis2019unreasonable}
Ioannis Caragiannis, David Kurokawa, Herv{\'e} Moulin, Ariel~D Procaccia,
  Nisarg Shah, and Junxing Wang.
\newblock The unreasonable fairness of maximum {N}ash welfare.
\newblock {\em ACM Transactions on Economics and Computation (TEAC)},
  7(3):1--32, 2019.
\newblock \href {http://dx.doi.org/10.1145/3355902}
  {\path{doi:10.1145/3355902}}.

\bibitem[CL12]{cakmak2012algorithmic}
Maya Cakmak and Manuel Lopes.
\newblock Algorithmic and human teaching of sequential decision tasks.
\newblock In {\em Proceedings of the Twenty-Sixth AAAI Conference on Artificial
  Intelligence}, page 1536–1542, 2012.
\newblock \href {http://dx.doi.org/10.5555/2900929.2900946}
  {\path{doi:10.5555/2900929.2900946}}.

\bibitem[Cla71]{clarke1971multipart}
Edward~H Clarke.
\newblock Multipart pricing of public goods.
\newblock {\em Public choice}, 11(1):17--33, 1971.
\newblock \href {http://dx.doi.org/10.1007/BF01726210}
  {\path{doi:10.1007/BF01726210}}.

\bibitem[CLB{\etalchar{+}}17]{christiano2017deep}
Paul~F Christiano, Jan Leike, Tom Brown, Miljan Martic, Shane Legg, and Dario
  Amodei.
\newblock Deep reinforcement learning from human preferences.
\newblock In {\em Advances in Neural Information Processing Systems},
  volume~30, pages 4299--4307, 2017.
\newblock URL:
  \url{https://proceedings.neurips.cc/paper/2017/file/d5e2c0adad503c91f91df240d0cd4e49-Paper.pdf}.

\bibitem[CLL{\etalchar{+}}18]{DBLP:journals/corr/abs-1804-03980}
Kris Cao, Angeliki Lazaridou, Marc Lanctot, Joel~Z. Leibo, Karl Tuyls, and
  Stephen Clark.
\newblock Emergent communication through negotiation.
\newblock preprint, 2018.
\newblock \href {http://arxiv.org/abs/1804.03980} {\path{arXiv:1804.03980}}.

\bibitem[COIO{\etalchar{+}}18]{crandall2018cooperating}
Jacob~W Crandall, Mayada Oudah, Fatimah Ishowo-Oloko, Sherief Abdallah,
  Jean-Fran{\c{c}}ois Bonnefon, Manuel Cebrian, Azim Shariff, Michael~A
  Goodrich, and Iyad Rahwan.
\newblock Cooperating with machines.
\newblock {\em Nature Communications}, 9:233, 2018.
\newblock \href {http://dx.doi.org/10.1038/s41467-017-02597-8}
  {\path{doi:10.1038/s41467-017-02597-8}}.

\bibitem[{Coo}19]{cooperativeAIcom}
{Cooperative AI Workshop}.
\newblock Cooperative {AI} workshop, {NeurIPS} 2020, September 2019.
\newblock URL: \url{https://www.cooperativeAI.com}.

\bibitem[CPS13]{caragiannis2013noisy}
Ioannis Caragiannis, Ariel~D Procaccia, and Nisarg Shah.
\newblock When do noisy votes reveal the truth?
\newblock In {\em Proceedings of the Fourteenth {ACM} Conference on Electronic
  Commerce}, pages 143--160, 2013.
\newblock \href {http://dx.doi.org/10.1145/2482540.2482570}
  {\path{doi:10.1145/2482540.2482570}}.

\bibitem[Cri19]{critch2019parametric}
Andrew Critch.
\newblock A parametric, resource-bounded generalization of {L}\"{o}b’s
  theorem, and a robust cooperation criterion for open-source game theory.
\newblock {\em The Journal of Symbolic Logic}, 84(4):1368--1381, 2019.
\newblock \href {http://dx.doi.org/10.1017/jsl.2017.42}
  {\path{doi:10.1017/jsl.2017.42}}.

\bibitem[CS82]{Crawford1982Strategic}
Vincent~P. Crawford and Joel Sobel.
\newblock Strategic information transmission.
\newblock {\em Econometrics}, 50(6):1431--1451, 1982.
\newblock \href {http://dx.doi.org/10.2307/1913390}
  {\path{doi:10.2307/1913390}}.

\bibitem[CS02a]{conitzer2002complexity}
Vincent Conitzer and Tuomas Sandholm.
\newblock Complexity of manipulating elections with few candidates.
\newblock In {\em Eighteenth National Conference on Artificial Intelligence},
  pages 314--319, 2002.

\bibitem[CS02b]{conitzer2002vote}
Vincent Conitzer and Tuomas Sandholm.
\newblock Vote elicitation: Complexity and strategy-proofness.
\newblock In {\em Eighteenth National Conference on Artificial Intelligence},
  pages 392--397, 2002.

\bibitem[CS06]{conitzer2006computing}
Vincent Conitzer and Tuomas Sandholm.
\newblock Computing the optimal strategy to commit to.
\newblock In {\em Proceedings of the 7th ACM Conference on Electronic
  Commerce}, pages 82--90, 2006.
\newblock \href {http://dx.doi.org/10.1145/1134707.1134717}
  {\path{doi:10.1145/1134707.1134717}}.

\bibitem[CSH{\etalchar{+}}19]{carroll2019utility}
Micah Carroll, Rohin Shah, Mark~K Ho, Thomas~L Griffiths, Sanjit~A Seshia,
  Pieter Abbeel, and Anca Dragan.
\newblock On the utility of learning about humans for human-{AI} coordination.
\newblock In 33, editor, {\em Advances in Neural Information Processing
  Systems}, pages 5175--5186, 2019.
\newblock URL:
  \url{https://papers.nips.cc/paper/2019/file/f5b1b89d98b7286673128a5fb112cb9a-Paper.pdf}.

\bibitem[CSLC19]{cohen2019trusted}
Robin Cohen, Mike Schaekermann, Sihao Liu, and Michael Cormier.
\newblock Trusted {AI} and the contribution of trust modeling in multiagent
  systems.
\newblock In {\em Proceedings of the 18th International Conference on
  Autonomous Agents and MultiAgent Systems}, pages 1644--1648, 2019.

\bibitem[CW94]{chang1994speech}
Man~Kit Chang and Carson~C Woo.
\newblock A speech-act-based negotiation protocol: design, implementation, and
  test use.
\newblock {\em ACM Transactions on Information Systems (TOIS)}, 12(4):360--382,
  1994.
\newblock \href {http://dx.doi.org/10.1145/185462.185477}
  {\path{doi:10.1145/185462.185477}}.

\bibitem[CWB{\etalchar{+}}11]{collobert2011natural}
Ronan Collobert, Jason Weston, L{\'e}on Bottou, Michael Karlen, Koray
  Kavukcuoglu, and Pavel Kuksa.
\newblock Natural language processing (almost) from scratch.
\newblock {\em Journal of {M}achine {L}earning {R}esearch}, 12:2493--2537,
  2011.
\newblock URL: \url{http://jmlr.org/papers/v12/collobert11a.html}.

\bibitem[CXL{\etalchar{+}}20]{chen2020delayaware}
Baiming Chen, Mengdi Xu, Zuxin Liu, Liang Li, and Ding Zhao.
\newblock Delay-aware multi-agent reinforcement learning for cooperative and
  competitive environments.
\newblock preprint, 2020.
\newblock \href {http://arxiv.org/abs/2005.05441} {\path{arXiv:2005.05441}}.

\bibitem[Day09]{dayan2009goal}
Peter Dayan.
\newblock Goal-directed control and its antipodes.
\newblock {\em Neural Networks}, 22(3):213--219, 2009.
\newblock \href {http://dx.doi.org/10.1016/j.neunet.2009.03.004}
  {\path{doi:10.1016/j.neunet.2009.03.004}}.

\bibitem[DC17]{deng2017disarmament}
Yuan Deng and Vincent Conitzer.
\newblock Disarmament games.
\newblock In {\em Proceedings of the Thirty-First AAAI Conference on Artificial
  Intelligence}, pages 473--479, 2017.
\newblock \href {http://dx.doi.org/10.5555/3298239.3298310}
  {\path{doi:10.5555/3298239.3298310}}.

\bibitem[Del03]{dellarocas2003digitization}
Chrysanthos Dellarocas.
\newblock The digitization of word of mouth: Promise and challenges of online
  feedback mechanisms.
\newblock {\em Management Science}, 49(10):1407--1424, 2003.
\newblock \href {http://dx.doi.org/10.1287/mnsc.49.10.1407.17308}
  {\path{doi:10.1287/mnsc.49.10.1407.17308}}.

\bibitem[DFJ{\etalchar{+}}12]{duttingFJLLP12}
Paul D{\"{u}}tting, Felix~A. Fischer, Pichayut Jirapinyo, John~K. Lai, Benjamin
  Lubin, and David~C. Parkes.
\newblock Payment rules through discriminant-based classifiers.
\newblock In Boi Faltings, Kevin Leyton{-}Brown, and Panos Ipeirotis, editors,
  {\em Proceedings of the 13th {ACM} Conference on Electronic Commerce}, pages
  477--494. {ACM}, 2012.
\newblock \href {http://dx.doi.org/10.1145/2229012.2229048}
  {\path{doi:10.1145/2229012.2229048}}.

\bibitem[DFN{\etalchar{+}}19]{duetting0NPR19}
Paul Duetting, Zhe Feng, Harikrishna Narasimhan, David~C. Parkes, and
  Sai~Srivatsa Ravindranath.
\newblock Optimal auctions through deep learning.
\newblock In Kamalika Chaudhuri and Ruslan Salakhutdinov, editors, {\em
  Proceedings of the 36th International Conference on Machine Learning},
  volume~97 of {\em Proceedings of Machine Learning Research}, pages
  1706--1715. {PMLR}, 2019.
\newblock URL: \url{http://proceedings.mlr.press/v97/duetting19a.html}.

\bibitem[Dig99]{dignum1999autonomous}
Frank Dignum.
\newblock Autonomous agents with norms.
\newblock {\em Artificial intelligence and Law}, 7:69--79, 1999.
\newblock \href {http://dx.doi.org/10.1023/A:1008315530323}
  {\path{doi:10.1023/A:1008315530323}}.

\bibitem[dKLW97]{d1997formal}
Mark d'Inverno, David Kinny, Michael Luck, and Michael Wooldridge.
\newblock A formal specification of {dMARS}.
\newblock In {\em International Workshop on Agent Theories, Architectures, and
  Languages}, pages 155--176. Springer, 1997.

\bibitem[DKS02]{dignum2002desires}
Frank Dignum, David Kinny, and Liz Sonenberg.
\newblock From desires, obligations and norms to goals.
\newblock {\em Cognitive Science Quarterly}, 2(3-4):407--430, 2002.

\bibitem[dL20]{deon2020testing}
Greg d'Eon and Kate Larson.
\newblock Testing axioms against human reward divisions in cooperative games.
\newblock In {\em Proceedings of the 19th International Conference on
  Autonomous Agents and MultiAgent Systems}, pages 312--320, 2020.
\newblock \href {http://dx.doi.org/10.5555/3398761.3398802}
  {\path{doi:10.5555/3398761.3398802}}.

\bibitem[DLS13]{dragan2013legibility}
Anca~D Dragan, Kenton~CT Lee, and Siddhartha~S Srinivasa.
\newblock Legibility and predictability of robot motion.
\newblock In {\em 8th ACM/IEEE International Conference on Human-Robot
  Interaction}, pages 301--308. IEEE, 2013.
\newblock \href {http://dx.doi.org/10.1109/HRI.2013.6483603}
  {\path{doi:10.1109/HRI.2013.6483603}}.

\bibitem[DS04]{dresner2004multiagent}
Kurt Dresner and Peter Stone.
\newblock Multiagent traffic management: A reservation-based intersection
  control mechanism.
\newblock In {\em Proceedings of the Third International Joint Conference on
  Autonomous Agents and Multiagent Systems-Volume 2}, pages 530--537, 2004.
\newblock \href {http://dx.doi.org/10.1109/AAMAS.2004.10121}
  {\path{doi:10.1109/AAMAS.2004.10121}}.

\bibitem[DSGC17]{da2017simultaneously}
Felipe~Leno Da~Silva, Ruben Glatt, and Anna Helena~Reali Costa.
\newblock Simultaneously learning and advising in multiagent reinforcement
  learning.
\newblock In {\em Proceedings of the 16th Conference on Autonomous Agents and
  Multiagent Systems}, pages 1100--1108, 2017.

\bibitem[dWVV15]{Weerd2015NegotiatingWO}
Harmen de~Weerd, Rineke Verbrugge, and Bart Verheij.
\newblock Negotiating with other minds: the role of recursive theory of mind in
  negotiation with incomplete information.
\newblock {\em Autonomous Agents and Multi-Agent Systems}, 31:250--287, 2015.
\newblock \href {http://dx.doi.org/10.1007/s10458-015-9317-1}
  {\path{doi:10.1007/s10458-015-9317-1}}.

\bibitem[dWVV17]{de2017negotiating}
Harmen de~Weerd, Rineke Verbrugge, and Bart Verheij.
\newblock Negotiating with other minds: the role of recursive theory of mind in
  negotiation with incomplete information.
\newblock {\em Autonomous Agents and Multi-Agent Systems}, 31(2):250--287,
  2017.
\newblock \href {http://dx.doi.org/10.1007/s10458-015-9317-1}
  {\path{doi:10.1007/s10458-015-9317-1}}.

\bibitem[EBL{\etalchar{+}}19]{eccles2019biases}
Tom Eccles, Yoram Bachrach, Guy Lever, Angeliki Lazaridou, and Thore Graepel.
\newblock Biases for emergent communication in multi-agent reinforcement
  learning, 2019.
\newblock \href {http://arxiv.org/abs/1912.05676} {\path{arXiv:1912.05676}}.

\bibitem[EO10]{10.2307/27871271}
Tore Ellingsen and Robert \"{O}stling.
\newblock When does communication improve coordination?
\newblock {\em The American Economic Review}, 100(4):1695--1724, 2010.
\newblock \href {http://dx.doi.org/10.1257/aer.100.4.1695}
  {\path{doi:10.1257/aer.100.4.1695}}.

\bibitem[ES17]{ezrachi2017artificial}
Ariel Ezrachi and Maurice~E Stucke.
\newblock Artificial intelligence \& collusion: When computers inhibit
  competition.
\newblock {\em University of Illinois Law Review}, pages 1775--1810, 2017.
\newblock URL:
  \url{https://illinoislawreview.org/wp-content/uploads/2017/10/Ezrachi-Stucke.pdf}.

\bibitem[FADFW16]{foerster2016learning}
Jakob Foerster, Ioannis~Alexandros Assael, Nando De~Freitas, and Shimon
  Whiteson.
\newblock Learning to communicate with deep multi-agent reinforcement learning.
\newblock In {\em Advances in Neural Information Processing Systems}, pages
  2137--2145, 2016.
\newblock URL:
  \url{https://proceedings.neurips.cc/paper/2016/file/c7635bfd99248a2cdef8249ef7bfbef4-Paper.pdf}.

\bibitem[Far88]{FARRELL1988209}
Joseph Farrell.
\newblock Communication, coordination and {N}ash equilibrium.
\newblock {\em Economics Letters}, 27(3):209--214, 1988.
\newblock \href {http://dx.doi.org/10.1016/0165-1765(88)90172-3}
  {\path{doi:10.1016/0165-1765(88)90172-3}}.

\bibitem[Far93]{farrell1993meaning}
Joseph Farrell.
\newblock Meaning and credibility in cheap-talk games.
\newblock {\em Games and Economic Behavior}, 5(4):514--531, 1993.
\newblock \href {http://dx.doi.org/10.1006/game.1993.1029}
  {\path{doi:10.1006/game.1993.1029}}.

\bibitem[FCAS{\etalchar{+}}17]{foerster2017learning}
Jakob~N Foerster, Richard~Y Chen, Maruan Al-Shedivat, Shimon Whiteson, Pieter
  Abbeel, and Igor Mordatch.
\newblock Learning with opponent-learning awareness.
\newblock preprint, 2017.
\newblock \href {http://arxiv.org/abs/1709.04326} {\path{arXiv:1709.04326}}.

\bibitem[FdWF{\etalchar{+}}18]{DBLP:journals/corr/abs-1810-11702}
Jakob~N. Foerster, Christian A.~Schr{\"{o}}der de~Witt, Gregory Farquhar,
  Philip H.~S. Torr, Wendelin Boehmer, and Shimon Whiteson.
\newblock Multi-agent common knowledge reinforcement learning.
\newblock preprint, 2018.
\newblock \href {http://arxiv.org/abs/1810.11702} {\path{arXiv:1810.11702}}.

\bibitem[Fea95]{fearon-rationalist}
James~D. Fearon.
\newblock Rationalist explanations for war.
\newblock {\em International Organization}, 49(3):379--414, 1995.
\newblock \href {http://dx.doi.org/10.1017/S0020818300033324}
  {\path{doi:10.1017/S0020818300033324}}.

\bibitem[Fea97]{fearon1997signaling}
James~D Fearon.
\newblock Signaling foreign policy interests: Tying hands versus sinking costs.
\newblock {\em Journal of Conflict Resolution}, 41(1):68--90, 1997.
\newblock \href {http://dx.doi.org/10.1177/0022002797041001004}
  {\path{doi:10.1177/0022002797041001004}}.

\bibitem[Fea98]{fearon1998commitment}
James~D Fearon.
\newblock Commitment problems and the spread of ethnic conflict.
\newblock In David~A. Lake and Donald Rothchild, editors, {\em The
  international spread of ethnic conflict}. Princeton University Press,
  Princeton, NJ, 1998.

\bibitem[Fea04]{fearon2004some}
James~D Fearon.
\newblock Why do some civil wars last so much longer than others?
\newblock {\em Journal of Peace Research}, 41(3):275--301, 2004.
\newblock \href {http://dx.doi.org/10.1177/0022343304043770}
  {\path{doi:10.1177/0022343304043770}}.

\bibitem[Fea20]{Fearon2020coopAI}
James~D. Fearon.
\newblock Two kinds of cooperative {AI} challenges: Game play and game design,
  2020.
\newblock \textit{Cooperative AI Workshop at NeurIPS 2020}.
\newblock URL:
  \url{https://slideslive.com/38938227/two-kinds-of-cooperative-ai-challenges-game-play-and-game-design}.

\bibitem[Fer99]{ferber1999multi}
Jacques Ferber.
\newblock {\em Multi-agent systems: an introduction to distributed artificial
  intelligence}, volume~1.
\newblock Addison-Wesley Reading, 1999.

\bibitem[FG12]{fulmer2012level}
C~Ashley Fulmer and Michele~J Gelfand.
\newblock At what level (and in whom) we trust: Trust across multiple
  organizational levels.
\newblock {\em Journal of Management}, 38(4):1167--1230, 2012.
\newblock \href {http://dx.doi.org/10.1177/0149206312439327}
  {\path{doi:10.1177/0149206312439327}}.

\bibitem[FHH10]{faliszewski2010using}
Piotr Faliszewski, Edith Hemaspaandra, and Lane~A Hemaspaandra.
\newblock Using complexity to protect elections.
\newblock {\em Communications of the ACM}, 53(11):74--82, 2010.
\newblock \href {http://dx.doi.org/10.1145/1839676.1839696}
  {\path{doi:10.1145/1839676.1839696}}.

\bibitem[Fit18]{fitch2018biology}
W~Tecumseh Fitch.
\newblock The biology and evolution of speech: a comparative analysis.
\newblock {\em Annual Review of Linguistics}, 4:255--279, 2018.
\newblock \href {http://dx.doi.org/10.1146/annurev-linguistics-011817-045748}
  {\path{doi:10.1146/annurev-linguistics-011817-045748}}.

\bibitem[FKM{\etalchar{+}}05]{fullam2005specification}
Karen~K Fullam, Tomas~B Klos, Guillaume Muller, Jordi Sabater, Andreas
  Schlosser, Zvi Topol, K~Suzanne Barber, Jeffrey~S Rosenschein, Laurent
  Vercouter, and Marco Voss.
\newblock A specification of the agent reputation and trust ({ART}) testbed:
  experimentation and competition for trust in agent societies.
\newblock In {\em Proceedings of the Fourth International Joint Conference on
  Autonomous Agents and Multiagent Systems}, pages 512--518, 2005.
\newblock \href {http://dx.doi.org/10.1145/1082473.1082551}
  {\path{doi:10.1145/1082473.1082551}}.

\bibitem[FL15]{frieden2015world}
Jeffry~A Frieden and David~A Lake.
\newblock {\em World Politics: Interests, Interactions, Institutions: Third
  International Student Edition}.
\newblock WW Norton \& Company, New York, 2015.

\bibitem[FN04]{frank2004problems}
Steven~A Frank and Martin~A Nowak.
\newblock Problems of somatic mutation and cancer.
\newblock {\em Bioessays}, 26(3):291--299, 2004.
\newblock \href {http://dx.doi.org/10.1002/bies.20000}
  {\path{doi:10.1002/bies.20000}}.

\bibitem[FN16]{frantz2016institutions}
Christopher~K Frantz and Mariusz Nowostawski.
\newblock From institutions to code: Towards automated generation of smart
  contracts.
\newblock In {\em 2016 IEEE 1st International Workshops on Foundations and
  Applications of Self* Systems (FAS* W)}, pages 210--215, 2016.
\newblock \href {http://dx.doi.org/0.1109/FAS-W.2016.53}
  {\path{doi:0.1109/FAS-W.2016.53}}.

\bibitem[FNP18]{Feng2018dl}
Zhe Feng, Harikrishna Narisimhan, and David~C. Parkes.
\newblock Deep learning for revenue-optimal auctions with budgets.
\newblock In {\em Proceedings of the 17th International Conference on
  Autonomous Agents and Multiagent Systems (AAMAS 2018)}, 2018.
\newblock \href {http://dx.doi.org/10.5555/3237383.3237439}
  {\path{doi:10.5555/3237383.3237439}}.

\bibitem[FP10]{faliszewski2010ai}
Piotr Faliszewski and Ariel~D Procaccia.
\newblock {AI}’s war on manipulation: Are we winning?
\newblock {\em AI Magazine}, 31(4):53--64, 2010.
\newblock \href {http://dx.doi.org/10.1609/aimag.v31i4.2314}
  {\path{doi:10.1609/aimag.v31i4.2314}}.

\bibitem[FRD15]{fourny2015perfect}
Ghislain Fourny, Stéphane Reiche, and Jean-Pierre Dupuy.
\newblock Perfect prediction equilibrium.
\newblock preprint, 2015.
\newblock \href {http://arxiv.org/abs/1409.6172} {\path{arXiv:1409.6172}}.

\bibitem[FSH{\etalchar{+}}18]{foerster2018bayesian}
Jakob~N Foerster, Francis Song, Edward Hughes, Neil Burch, Iain Dunning, Shimon
  Whiteson, Matthew Botvinick, and Michael Bowling.
\newblock Bayesian action decoder for deep multi-agent reinforcement learning.
\newblock preprint, 2018.
\newblock \href {http://arxiv.org/abs/1811.01458} {\path{arXiv:1811.01458}}.

\bibitem[FSJ02]{faratin2002using}
Peyman Faratin, Carles Sierra, and Nicholas~R Jennings.
\newblock Using similarity criteria to make issue trade-offs in automated
  negotiations.
\newblock {\em Artificial Intelligence}, 142(2):205--237, 2002.
\newblock \href {http://dx.doi.org/10.1016/S0004-3702(02)00290-4}
  {\path{doi:10.1016/S0004-3702(02)00290-4}}.

\bibitem[FSL11]{fogarty2010}
L~Fogarty, P~Strimling, and K.~N. Laland.
\newblock The evolution of teaching.
\newblock {\em Evolution}, 65:2760--2770, 10 2011.
\newblock \href {http://dx.doi.org/10.1111/j.1558-5646.2011.01370.x}
  {\path{doi:10.1111/j.1558-5646.2011.01370.x}}.

\bibitem[FST15]{fang2015security}
Fei Fang, Peter Stone, and Milind Tambe.
\newblock When security games go green: Designing defender strategies to
  prevent poaching and illegal fishing.
\newblock In {\em Proceedings of the 24th International Conference on
  Artificial Intelligence}, pages 2589--2595, 2015.
\newblock \href {http://dx.doi.org/0.5555/2832581.2832611}
  {\path{doi:0.5555/2832581.2832611}}.

\bibitem[GBC07]{guzzoni2007modeling}
Didier Guzzoni, Charles Baur, and Adam Cheyer.
\newblock Modeling human-agent interaction with active ontologies.
\newblock In {\em AAAI Spring Symposium: Interaction Challenges for Intelligent
  Assistants}, pages 52--59, 2007.
\newblock URL:
  \url{https://www.aaai.org/Library/Symposia/Spring/2007/ss07-04-009.php}.

\bibitem[GBL08]{goe2008approaches}
Laura Goe, Courtney Bell, and Olivia Little.
\newblock Approaches to evaluating teacher effectiveness: A research synthesis.
\newblock Technical report, National Comprehensive Center for Teacher Quality,
  2008.
\newblock URL: \url{https://eric.ed.gov/?id=ED521228}.

\bibitem[Gen09]{gentry2009fully}
Craig Gentry.
\newblock {\em A fully homomorphic encryption scheme}.
\newblock PhD thesis, Stanford University, 2009.
\newblock URL: \url{https://crypto.stanford.edu/craig/craig-thesis.pdf}.

\bibitem[Geo88]{georgeff1988communication}
Michael Georgeff.
\newblock Communication and interaction in multi-agent planning.
\newblock In Alan~H. Bond and Les Gasser, editors, {\em Readings in Distributed
  Artificial Intelligence}, pages 200--204. Elsevier, 1988.

\bibitem[Gib73]{gibbard1973manipulation}
Allan Gibbard.
\newblock Manipulation of voting schemes: a general result.
\newblock {\em Econometrica: journal of the Econometric Society}, pages
  587--601, 1973.

\bibitem[GIM{\etalchar{+}}18]{governatori2018legal}
Guido Governatori, Florian Idelberger, Zoran Milosevic, Regis Riveret, Giovanni
  Sartor, and Xiwei Xu.
\newblock On legal contracts, imperative and declarative smart contracts, and
  blockchain systems.
\newblock {\em Artificial Intelligence and Law}, 26:377--409, 2018.
\newblock \href {http://dx.doi.org/10.1007/s10506-018-9223-3}
  {\path{doi:10.1007/s10506-018-9223-3}}.

\bibitem[GJA03]{gupta2003reputation}
Minaxi Gupta, Paul Judge, and Mostafa Ammar.
\newblock A reputation system for peer-to-peer networks.
\newblock In {\em Proceedings of the 13th international Workshop on Network and
  Operating Systems Support for Digital Audio and Video}, pages 144--152, 2003.
\newblock \href {http://dx.doi.org/10.1145/776322.776346}
  {\path{doi:10.1145/776322.776346}}.

\bibitem[GL77]{green1977characterization}
Jerry Green and Jean-Jacques Laffont.
\newblock Characterization of satisfactory mechanisms for the revelation of
  preferences for public goods.
\newblock {\em Econometrica}, 45(2):427--438, 1977.
\newblock \href {http://dx.doi.org/10.2307/1911219}
  {\path{doi:10.2307/1911219}}.

\bibitem[GLBML01]{golle2001incentives}
Philippe Golle, Kevin Leyton-Brown, Ilya Mironov, and Mark Lillibridge.
\newblock Incentives for sharing in peer-to-peer networks.
\newblock In {\em International Workshop on Electronic Commerce}, pages 75--87.
  Springer, 2001.
\newblock \href {http://dx.doi.org/10.1007/3-540-45598-1_9}
  {\path{doi:10.1007/3-540-45598-1_9}}.

\bibitem[GLN12]{Graepel2012Confidential}
Thore Graepel, Kristin Lauter, and Michael Naehrig.
\newblock {ML} confidential: Machine learning on encrypted data.
\newblock In {\em Information Security and Cryptology – ICISC 2012}, number
  7839 in LNCS, 2012.
\newblock \href {http://dx.doi.org/10.1007/978-3-642-37682-5_1}
  {\path{doi:10.1007/978-3-642-37682-5_1}}.

\bibitem[GLNS20]{guo2020proof}
Hao Guo, Wanxin Li, Mark Nejad, and Chien-Chung Shen.
\newblock Proof-of-event recording system for autonomous vehicles: A
  blockchain-based solution.
\newblock {\em IEEE Access}, 8:182776--182786, 2020.
\newblock \href {http://dx.doi.org/10.1109/ACCESS.2020.3029512}
  {\path{doi:10.1109/ACCESS.2020.3029512}}.

\bibitem[GM04]{gerkey2004formal}
Brian~P Gerkey and Maja~J Matari{\'c}.
\newblock A formal analysis and taxonomy of task allocation in multi-robot
  systems.
\newblock {\em The International Journal of Robotics Research}, 23(9):939--954,
  2004.
\newblock \href {http://dx.doi.org/10.1177/0278364904045564}
  {\path{doi:10.1177/0278364904045564}}.

\bibitem[GMB17]{goldin2017gesture}
Susan Goldin-Meadow and Diane Brentari.
\newblock Gesture, sign, and language: The coming of age of sign language and
  gesture studies.
\newblock {\em Behavioral and Brain Sciences}, 40:e46, 2017.
\newblock \href {http://dx.doi.org/10.1017/S0140525X15001247}
  {\path{doi:10.1017/S0140525X15001247}}.

\bibitem[GMW94]{greif1994coordination}
Avner Greif, Paul Milgrom, and Barry~R Weingast.
\newblock Coordination, commitment, and enforcement: The case of the merchant
  guild.
\newblock {\em Journal of Political Economy}, 102(4):745--776, 1994.
\newblock \href {http://dx.doi.org/10.1086/261953} {\path{doi:10.1086/261953}}.

\bibitem[GPAM{\etalchar{+}}14]{goodfellow2014generative}
Ian Goodfellow, Jean Pouget-Abadie, Mehdi Mirza, Bing Xu, David Warde-Farley,
  Sherjil Ozair, Aaron Courville, and Yoshua Bengio.
\newblock Generative adversarial nets.
\newblock In {\em Advances in Neural Information Processing Systems},
  volume~27, pages 2672--2680, 2014.
\newblock URL:
  \url{https://proceedings.neurips.cc/paper/2014/file/5ca3e9b122f61f8f06494c97b1afccf3-Paper.pdf}.

\bibitem[GPP{\etalchar{+}}98]{georgeff1998belief}
Michael Georgeff, Barney Pell, Martha Pollack, Milind Tambe, and Michael
  Wooldridge.
\newblock The belief-desire-intention model of agency.
\newblock In {\em International Workshop on Agent Theories, Architectures, and
  Languages}, pages 1--10. Springer, 1998.
\newblock \href {http://dx.doi.org/10.1007/3-540-49057-4_1}
  {\path{doi:10.1007/3-540-49057-4_1}}.

\bibitem[GR08]{ghazanfar2008evolution}
Asif~A Ghazanfar and Drew Rendall.
\newblock Evolution of human vocal production.
\newblock {\em Current Biology}, 18(11):R457--R460, 2008.
\newblock URL:
  \url{https://www.cell.com/current-biology/pdf/S0960-9822(08)00371-0.pdf}.

\bibitem[Gre89]{greif1989reputation}
Avner Greif.
\newblock Reputation and coalitions in medieval trade: evidence on the
  {M}aghribi traders.
\newblock {\em The Journal of Economic History}, 49(4):857--882, 1989.
\newblock \href {http://dx.doi.org/10.1017/S0022050700009475}
  {\path{doi:10.1017/S0022050700009475}}.

\bibitem[Gre94]{greif1994cultural}
Avner Greif.
\newblock Cultural beliefs and the organization of society: A historical and
  theoretical reflection on collectivist and individualist societies.
\newblock {\em Journal of Political Economy}, 102(5):912--950, 1994.
\newblock \href {http://dx.doi.org/10.1086/261959} {\path{doi:10.1086/261959}}.

\bibitem[Gre06]{greif2006institutions}
Avner Greif.
\newblock {\em Institutions and the path to the modern economy: Lessons from
  medieval trade}.
\newblock Cambridge University Press, New York, 2006.

\bibitem[Gro73]{groves1973incentives}
Theodore Groves.
\newblock Incentives in teams.
\newblock {\em Econometrica}, 41(4):617--631, 1973.
\newblock \href {http://dx.doi.org/10.2307/1914085}
  {\path{doi:10.2307/1914085}}.

\bibitem[Gro07]{grossi2007designing}
Davide Grossi.
\newblock {\em Designing invisible handcuffs: Formal investigations in
  institutions and organizations for multi-agent systems}.
\newblock PhD thesis, Utrecht University, 2007.
\newblock URL: \url{https://dspace.library.uu.nl/handle/1874/22838}.

\bibitem[GRW19]{gaudiosi2019negotiating}
Rebecca~W Gaudiosi, Jimena~Leiva Roesch, and Ye-Min Wu.
\newblock {\em Negotiating at the United Nations: A Practitioner's Guide}.
\newblock Routledge, London, 2019.

\bibitem[GS01]{greenwald2001autonomous}
Amy Greenwald and Peter Stone.
\newblock Autonomous bidding agents in the trading agent competition.
\newblock {\em IEEE Internet Computing}, 5(2):52--60, 2001.
\newblock \href {http://dx.doi.org/10.1109/4236.914648}
  {\path{doi:10.1109/4236.914648}}.

\bibitem[GS07]{garfinkel2007economics}
Michelle~R Garfinkel and Stergios Skaperdas.
\newblock Economics of conflict: An overview.
\newblock {\em Handbook of Defense Economics}, 2:649--709, 2007.
\newblock \href {http://dx.doi.org/10.1016/S1574-0013(06)02022-9}
  {\path{doi:10.1016/S1574-0013(06)02022-9}}.

\bibitem[H{\etalchar{+}}00]{hirshleifer2000game}
Jack Hirshleifer et~al.
\newblock Game-theoretic interpretations of commitment.
\newblock Technical report, UCLA Department of Economics, 2000.
\newblock URL: \url{http://www.econ.ucla.edu/workingpapers/wp799.pdf}.

\bibitem[Had20]{Hadfield2020coopAI}
Gillian Hadfield.
\newblock The normative infrastructure of cooperation, 2020.
\newblock \textit{Cooperative AI Workshop at NeurIPS 2020}.
\newblock URL:
  \url{https://slideslive.com/38938226/the-normative-infrastructure-of-cooperation}.

\bibitem[HAE{\etalchar{+}}20]{hughes2020learning}
Edward Hughes, Thomas~W. Anthony, Tom Eccles, Joel~Z. Leibo, David Balduzzi,
  and Yoram Bachrach.
\newblock Learning to resolve alliance dilemmas in many-player zero-sum games.
\newblock preprint, 2020.
\newblock \href {http://arxiv.org/abs/2003.00799} {\path{arXiv:2003.00799}}.

\bibitem[Har15]{harari2015sapiens}
Yuval~Noah Harari.
\newblock {\em Sapiens: A Brief History of Humankind}.
\newblock Harper, New York, 2015.

\bibitem[Har16]{harari2016homo}
Yuval~Noah Harari.
\newblock {\em Homo Deus: A brief history of tomorrow}.
\newblock Random House, New York, 2016.

\bibitem[HBB{\etalchar{+}}04]{henrich2004foundations}
Joseph~Patrick Henrich, Robert Boyd, Samuel Bowles, Ernst Fehr, Colin Camerer,
  and Herbert Gintis, editors.
\newblock {\em Foundations of human sociality: Economic experiments and
  ethnographic evidence from fifteen small-scale societies}.
\newblock Oxford University Press, Oxford, 2004.
\newblock \href {http://dx.doi.org/10.1093/0199262055.001.0001}
  {\path{doi:10.1093/0199262055.001.0001}}.

\bibitem[Hen16]{10.2307/j.ctvc77f0d}
Joseph Henrich.
\newblock {\em The Secret of Our Success: How Culture Is Driving Human
  Evolution, Domesticating Our Species, and Making Us Smarter}.
\newblock Princeton University Press, Princeton, NJ, 2016.
\newblock \href {http://dx.doi.org/10.2307/j.ctvc77f0d}
  {\path{doi:10.2307/j.ctvc77f0d}}.

\bibitem[Hic39]{10.2307/2225023}
J.~R. Hicks.
\newblock The foundations of welfare economics.
\newblock {\em The Economic Journal}, 49(196):696--712, 1939.
\newblock \href {http://dx.doi.org/10.2307/2225023}
  {\path{doi:10.2307/2225023}}.

\bibitem[HL04]{horling2004survey}
Bryan Horling and Victor Lesser.
\newblock A survey of multi-agent organizational paradigms.
\newblock {\em The Knowledge Engineering Review}, 19(4):281--316, 2004.
\newblock \href {http://dx.doi.org/10.1017/S0269888905000317}
  {\path{doi:10.1017/S0269888905000317}}.

\bibitem[HL13]{hoppitt2013social}
William Hoppitt and Kevin~N Laland.
\newblock {\em Social learning: an introduction to mechanisms, methods, and
  models}.
\newblock Princeton University Press, Princeton, NJ, 2013.

\bibitem[HLF{\etalchar{+}}15]{huys2015interplay}
Quentin J~M Huys, N{\'\i}all Lally, Paul Faulkner, Neir Eshel, Erich Seifritz,
  Samuel~J Gershman, Peter Dayan, and Jonathan~P Roiser.
\newblock Interplay of approximate planning strategies.
\newblock {\em Proceedings of the National Academy of Sciences},
  112(10):3098--3103, 2015.
\newblock \href {http://dx.doi.org/10.1073/pnas.1414219112}
  {\path{doi:10.1073/pnas.1414219112}}.

\bibitem[HLP{\etalchar{+}}18]{DBLP:journals/corr/abs-1803-08884}
Edward Hughes, Joel~Z. Leibo, Matthew~G. Philips, Karl Tuyls, Edgar~A.
  Du{\'{e}}{\~{n}}ez{-}Guzm{\'{a}}n, Antonio~Garc{\'{\i}}a Casta{\~{n}}eda,
  Iain Dunning, Tina Zhu, Kevin~R. McKee, Raphael Koster, Heather Roff, and
  Thore Graepel.
\newblock Inequity aversion resolves intertemporal social dilemmas.
\newblock preprint, 2018.
\newblock \href {http://arxiv.org/abs/1803.08884} {\path{arXiv:1803.08884}}.

\bibitem[HLPF20]{hu2020otherplay}
Hengyuan Hu, Adam Lerer, Alex Peysakhovich, and Jakob Foerster.
\newblock "{O}ther-play" for zero-shot coordination.
\newblock preprint, 2020.
\newblock \href {http://arxiv.org/abs/2003.02979} {\path{arXiv:2003.02979}}.

\bibitem[HMC00]{hart2000simple}
Sergiu Hart and Andreu Mas-Colell.
\newblock A simple adaptive procedure leading to correlated equilibrium.
\newblock {\em Econometrica}, 68(5):1127--1150, 2000.
\newblock \href {http://dx.doi.org/10.1111/1468-0262.00153}
  {\path{doi:10.1111/1468-0262.00153}}.

\bibitem[HMRAD16]{hadfield2016cooperative}
Dylan Hadfield-Menell, Stuart~J Russell, Pieter Abbeel, and Anca Dragan.
\newblock Cooperative inverse reinforcement learning.
\newblock In {\em Advances in Neural Information Processing Systems},
  volume~29, pages 3909--3917, 2016.
\newblock URL:
  \url{https://proceedings.neurips.cc/paper/2016/file/c3395dd46c34fa7fd8d729d8cf88b7a8-Paper.pdf}.

\bibitem[Hof08]{hofstadter2008metamagical}
Douglas~R Hofstadter.
\newblock {\em Metamagical themas: Questing for the essence of mind and
  pattern}.
\newblock Basic Books, New York, 2008.

\bibitem[Hov98]{hovi1998games}
Jon Hovi.
\newblock {\em Games, threats, and treaties: understanding commitments in
  international relations}.
\newblock Pinter Pub Limited, 1998.

\bibitem[HPS16]{hardt2016equality}
Moritz Hardt, Eric Price, and Nathan Srebro.
\newblock Equality of opportunity in supervised learning.
\newblock In {\em Advances in Neural Information Processing Systems},
  volume~29, pages 3315--3323, 2016.
\newblock URL:
  \url{https://proceedings.neurips.cc/paper/2016/file/9d2682367c3935defcb1f9e247a97c0d-Paper.pdf}.

\bibitem[HS13]{hazard2013macau}
Christopher~J Hazard and Munindar~P Singh.
\newblock Macau: A basis for evaluating reputation systems.
\newblock In {\em Twenty-Third International Joint Conference on Artificial
  Intelligence (IJCAI 2013)}, 2013.
\newblock \href {http://dx.doi.org/10.5555/2540128.2540158}
  {\path{doi:10.5555/2540128.2540158}}.

\bibitem[HS15]{hausknecht2015deep}
Matthew Hausknecht and Peter Stone.
\newblock Deep recurrent {Q}-learning for partially observable {MDP}s.
\newblock preprint, 2015.
\newblock \href {http://arxiv.org/abs/1507.06527} {\path{arXiv:1507.06527}}.

\bibitem[HT17]{havrylov2017emergence}
Serhii Havrylov and Ivan Titov.
\newblock Emergence of language with multi-agent games: Learning to communicate
  with sequences of symbols.
\newblock In {\em Advances in Neural Information Processing Systems},
  volume~30, pages 2149--2159, 2017.
\newblock URL:
  \url{https://papers.nips.cc/paper/2017/file/70222949cc0db89ab32c9969754d4758-Paper.pdf}.

\bibitem[Hub99]{huber1999jam}
Marcus~J Huber.
\newblock {JAM}: A {BDI}-theoretic mobile agent architecture.
\newblock In {\em Proceedings of the Third Annual Conference on Autonomous
  Agents}, pages 236--243, 1999.
\newblock \href {http://dx.doi.org/10.1145/301136.301202}
  {\path{doi:10.1145/301136.301202}}.

\bibitem[Hun06]{huntington2006political}
Samuel~P Huntington.
\newblock {\em Political order in changing societies}.
\newblock Yale University Press, New Haven, 2006.

\bibitem[IHK07]{ito2007multi}
Takayuki Ito, Hiromitsu Hattori, and Mark Klein.
\newblock Multi-issue negotiation protocol for agents: Exploring nonlinear
  utility spaces.
\newblock In {\em Proceedings of the 20th International Joint Conference on
  Artificial Intelligence}, pages 1347--1352, 2007.
\newblock URL: \url{https://www.ijcai.org/Proceedings/07/Papers/217.pdf}.

\bibitem[ILP{\etalchar{+}}18]{ibarz2018reward}
Borja Ibarz, Jan Leike, Tobias Pohlen, Geoffrey Irving, Shane Legg, and Dario
  Amodei.
\newblock Reward learning from human preferences and demonstrations in {A}tari.
\newblock In {\em Advances in Neural Information Processing Systems},
  volume~31, pages 8011--8023, 2018.
\newblock URL:
  \url{https://papers.nips.cc/paper/2018/file/8cbe9ce23f42628c98f80fa0fac8b19a-Paper.pdf}.

\bibitem[JCD{\etalchar{+}}19]{jaderberg2019human}
Max Jaderberg, Wojciech~M Czarnecki, Iain Dunning, Luke Marris, Guy Lever,
  Antonio~Garcia Castañeda, Charles Beattie, Neil~C Rabinowitz, Ari~S Morcos,
  Avraham Ruderman, Nicolas Sonnerat, Tim Green, Louise Deason, Joel~Z. Leibo,
  DAvid Silver, Demis Hassabis, Koray Kavukcuoglu, and Thore Graepel.
\newblock Human-level performance in 3d multiplayer games with population-based
  reinforcement learning.
\newblock {\em Science}, 364(6443):859--865, 2019.
\newblock \href {http://dx.doi.org/10.1126/science.aau6249}
  {\path{doi:10.1126/science.aau6249}}.

\bibitem[Jen96]{jennings1996coordination}
Nicholas~R Jennings.
\newblock Coordination techniques for distributed artificial intelligence.
\newblock In Greg M~P O'Hare and Nicholas~R Jennings, editors, {\em Foundations
  of Distributed Artificial Intelligence}, pages 187--210. Wiley, New York,
  1996.

\bibitem[Jen00]{jennings2000agent}
Nicholas~R Jennings.
\newblock On agent-based software engineering.
\newblock {\em Artificial intelligence}, 117(2):277--296, 2000.
\newblock \href {http://dx.doi.org/10.1016/S0004-3702(99)00107-1}
  {\path{doi:10.1016/S0004-3702(99)00107-1}}.

\bibitem[Jer89]{jervis1989logic}
Robert Jervis.
\newblock {\em The logic of images in international relations}.
\newblock Columbia University Press, New York, 1989.

\bibitem[JF09]{jurca2009mechanisms}
Radu Jurca and Boi Faltings.
\newblock Mechanisms for making crowds truthful.
\newblock {\em Journal of Artificial Intelligence Research}, 34:209--253, 2009.
\newblock \href {http://dx.doi.org/10.1613/jair.2621}
  {\path{doi:10.1613/jair.2621}}.

\bibitem[JFL{\etalchar{+}}01]{jennings2001automated}
Nicholas~R Jennings, Peyman Faratin, Alessio~R Lomuscio, Simon Parsons, Carles
  Sierra, and Michael Wooldridge.
\newblock Automated negotiation: prospects, methods and challenges.
\newblock {\em Group Decision and Negotiation}, 10(2):199--215, 2001.
\newblock \href {http://dx.doi.org/10.1023/A:1008746126376}
  {\path{doi:10.1023/A:1008746126376}}.

\bibitem[JG09]{josang2009challenges}
Audun J{\o}sang and Jennifer Golbeck.
\newblock Challenges for robust trust and reputation systems.
\newblock In {\em Proceedings of the 5th International Workshop on Security and
  Trust Management}, page~52, 2009.

\bibitem[JLH{\etalchar{+}}18]{DBLP:journals/corr/abs-1810-08647}
Natasha Jaques, Angeliki Lazaridou, Edward Hughes, {\c{C}}aglar
  G{\"{u}}l{\c{c}}ehre, Pedro~A. Ortega, DJ~Strouse, Joel~Z. Leibo, and Nando
  de~Freitas.
\newblock Intrinsic social motivation via causal influence in multi-agent {RL}.
\newblock preprint, 2018.
\newblock \href {http://arxiv.org/abs/1810.08647} {\path{arXiv:1810.08647}}.

\bibitem[JLH{\etalchar{+}}19]{jaques2019social}
Natasha Jaques, Angeliki Lazaridou, Edward Hughes, Caglar Gulcehre, Pedro
  Ortega, DJ~Strouse, Joel~Z Leibo, and Nando De~Freitas.
\newblock Social influence as intrinsic motivation for multi-agent deep
  reinforcement learning.
\newblock In {\em Proceedings of the 36th International Conference on Machine
  Learning,}, pages 3040--3049, 2019.
\newblock URL: \url{http://proceedings.mlr.press/v97/jaques19a.html}.

\bibitem[JM11]{jackson2011reasons}
Matthew~O Jackson and Massimo Morelli.
\newblock The reasons for wars: an updated survey.
\newblock In Christopher~J coyne and Rachel~L Mathers, editors, {\em The
  handbook on the political economy of war}, pages 34--57. Edward Elgar,
  Cheltenham, UK, 2011.

\bibitem[JMD20]{jeon2020reward}
Hong~Jun Jeon, Smitha Milli, and Anca~D Dragan.
\newblock Reward-rational (implicit) choice: A unifying formalism for reward
  learning.
\newblock {\em Advances in Neural Information Processing Systems}, 33, 2020.
\newblock URL:
  \url{https://papers.nips.cc/paper/2020/file/2f10c1578a0706e06b6d7db6f0b4a6af-Paper.pdf}.

\bibitem[JMP{\etalchar{+}}14]{jiang2014diverse}
Albert~Xin Jiang, Leandro~Soriano Marcolino, Ariel~D Procaccia, Tuomas
  Sandholm, Nisarg Shah, and Milind Tambe.
\newblock Diverse randomized agents vote to win.
\newblock In {\em Advances in Neural Information Processing Systems},
  volume~27, pages 2573--2581, 2014.
\newblock URL:
  \url{https://proceedings.neurips.cc/paper/2014/file/8edd72158ccd2a879f79cb2538568fdc-Paper.pdf}.

\bibitem[JSW98]{jennings1998roadmap}
Nicholas~R Jennings, Katia Sycara, and Michael Wooldridge.
\newblock A roadmap of agent research and development.
\newblock {\em Autonomous Agents and Multi-Agent Systems}, 1(1):7--38, 1998.
\newblock \href {http://dx.doi.org/10.1023/A:1010090405266}
  {\path{doi:10.1023/A:1010090405266}}.

\bibitem[KAK{\etalchar{+}}97]{kitano1997robocup}
Hiroaki Kitano, Minoru Asada, Yasuo Kuniyoshi, Itsuki Noda, and Eiichi Osawa.
\newblock Robocup: The robot world cup initiative.
\newblock In {\em Proceedings of the First International Conference on
  Autonomous Agents}, pages 340--347, 1997.
\newblock \href {http://dx.doi.org/10.1145/267658.267738}
  {\path{doi:10.1145/267658.267738}}.

\bibitem[Kal39]{10.2307/2224835}
Nicholas Kaldor.
\newblock Welfare propositions of economics and interpersonal comparisons of
  utility.
\newblock {\em The Economic Journal}, 49(195):549--552, 1939.
\newblock \href {http://dx.doi.org/10.2307/2224835}
  {\path{doi:10.2307/2224835}}.

\bibitem[Kam13]{kaminka2013curing}
Gal~A Kaminka.
\newblock Curing robot autism: a challenge.
\newblock In {\em Proceedings of the 2013 International Conference on
  Autonomous Agents and Multiagent Systems (AAMAS)}, pages 801--804, 2013.
\newblock \href {http://dx.doi.org/10.5555/2484920.2485047}
  {\path{doi:10.5555/2484920.2485047}}.

\bibitem[Kat90]{10.2307/1289373}
Avery Katz.
\newblock The strategic structure of offer and acceptance: Game theory and the
  law of contract formation.
\newblock {\em Michigan Law Review}, 89(2):215--295, 1990.
\newblock \href {http://dx.doi.org/10.2307/1289373}
  {\path{doi:10.2307/1289373}}.

\bibitem[KHMHL20]{koster2020silly}
Raphael K{\"o}ster, Dylan Hadfield-Menell, Gillian~K Hadfield, and Joel~Z
  Leibo.
\newblock Silly rules improve the capacity of agents to learn stable
  enforcement and compliance behaviors.
\newblock preprint, 2020.
\newblock \href {http://arxiv.org/abs/2001.09318} {\path{arXiv:2001.09318}}.

\bibitem[KKLS10]{kalai2010commitment}
Adam~Tauman Kalai, Ehud Kalai, Ehud Lehrer, and Dov Samet.
\newblock A commitment folk theorem.
\newblock {\em Games and Economic Behavior}, 69(1):127--137, 2010.
\newblock \href {http://dx.doi.org/10.1016/j.geb.2009.09.008}
  {\path{doi:10.1016/j.geb.2009.09.008}}.

\bibitem[KL95]{kraus1995designing}
Sarit Kraus and Daniel Lehmann.
\newblock Designing and building a negotiating automated agent.
\newblock {\em Computational Intelligence}, 11(1):132--171, 1995.
\newblock \href {http://dx.doi.org/10.1111/j.1467-8640.1995.tb00026.x}
  {\path{doi:10.1111/j.1467-8640.1995.tb00026.x}}.

\bibitem[Kla87]{klatt1987review}
Dennis~H Klatt.
\newblock Review of text-to-speech conversion for {E}nglish.
\newblock {\em The Journal of the Acoustical Society of America},
  82(3):737--793, 1987.
\newblock \href {http://dx.doi.org/10.1121/1.395275}
  {\path{doi:10.1121/1.395275}}.

\bibitem[KLC09]{kastidou2009exchanging}
Georgia Kastidou, Kate Larson, and Robin Cohen.
\newblock Exchanging reputation information between communities: A
  payment-function approach.
\newblock In {\em Proceedings of the 21st International Joint Conference on
  Artificial Intelligence}, pages 195--200, 2009.

\bibitem[KM06]{kakas2006adaptive}
Antonis Kakas and Pavlos Moraitis.
\newblock Adaptive agent negotiation via argumentation.
\newblock In {\em Proceedings of the Fifth International Joint Conference on
  Autonomous Agents and Multiagent Systems}, pages 384--391, 2006.
\newblock \href {http://dx.doi.org/10.1145/1160633.1160701}
  {\path{doi:10.1145/1160633.1160701}}.

\bibitem[KMLB17]{kottur2017natural}
Satwik Kottur, Jos{\'e}~MF Moura, Stefan Lee, and Dhruv Batra.
\newblock Natural language does not emerge'naturally'in multi-agent dialog.
\newblock preprint, 2017.
\newblock \href {http://arxiv.org/abs/1706.08502} {\path{arXiv:1706.08502}}.

\bibitem[KMS{\etalchar{+}}16]{kosba2016hawk}
Ahmed Kosba, Andrew Miller, Elaine Shi, Zikai Wen, and Charalampos Papamanthou.
\newblock Hawk: The blockchain model of cryptography and privacy-preserving
  smart contracts.
\newblock In {\em IEEE Symposium on Security and Privacy}, pages 839--858.
  IEEE, 2016.
\newblock \href {http://dx.doi.org/10.1109/SP.2016.55}
  {\path{doi:10.1109/SP.2016.55}}.

\bibitem[Kni92]{knight1992institutions}
Jack Knight.
\newblock {\em Institutions and social conflict}.
\newblock Cambridge University Press, Cambridge, UK, 1992.

\bibitem[Kor83]{kornhauser1983reliance}
Lewis~A Kornhauser.
\newblock Reliance, reputation, and breach of contract.
\newblock {\em The Journal of Law and Economics}, 26(3):691--706, 1983.
\newblock \href {http://dx.doi.org/10.1086/467054} {\path{doi:10.1086/467054}}.

\bibitem[KP95]{komorita1995interpersonal}
Samuel~S Komorita and Craig~D Parks.
\newblock Interpersonal relations: Mixed-motive interaction.
\newblock {\em Annual Review of Psychology}, 46(1):183--207, 1995.
\newblock \href {http://dx.doi.org/10.1146/annurev.ps.46.020195.001151}
  {\path{doi:10.1146/annurev.ps.46.020195.001151}}.

\bibitem[Kra97]{kraus1997negotiation}
Sarit Kraus.
\newblock Negotiation and cooperation in multi-agent environments.
\newblock {\em Artificial intelligence}, 94(1-2):79--97, 1997.
\newblock \href {http://dx.doi.org/10.1016/S0004-3702(97)00025-8}
  {\path{doi:10.1016/S0004-3702(97)00025-8}}.

\bibitem[Kra01]{kraus2001strategic}
Sarit Kraus.
\newblock {\em Strategic negotiation in multiagent environments}.
\newblock MIT Press, Cambridge, MA, 2001.

\bibitem[KSH12]{krizhevsky2012imagenet}
Alex Krizhevsky, Ilya Sutskever, and Geoffrey~E Hinton.
\newblock Image{N}et classification with deep convolutional neural networks.
\newblock In {\em Advances in Neural Information Processing Systems},
  volume~25, pages 1097--1105, 2012.
\newblock URL:
  \url{https://proceedings.neurips.cc/paper/2012/file/c399862d3b9d6b76c8436e924a68c45b-Paper.pdf}.

\bibitem[KUM{\etalchar{+}}]{krakovna2020specification}
Victoria Krakovna, Jonathan Uesato, Vladimir Mikulik, Matthew Rahtz, Tom
  Everitt, Ramana Kumar, Zac Kenton, Jan Leike, and Shane Legg.
\newblock Specification gaming: the flip side of ai ingenuity, 2020.
\newblock blog.
\newblock URL:
  \url{https://deepmind.com/blog/article/Specification-gaming-the-flip-side-of-AI-ingenuity}.

\bibitem[KUR18]{kreutzer2018reliability}
Julia Kreutzer, Joshua Uyheng, and Stefan Riezler.
\newblock Reliability and learnability of human bandit feedback for
  sequence-to-sequence reinforcement learning.
\newblock preprint, 2018.
\newblock \href {http://arxiv.org/abs/1805.10627} {\path{arXiv:1805.10627}}.

\bibitem[KV00]{krasa2000optimal}
Stefan Krasa and Anne~P Villamil.
\newblock Optimal contracts when enforcement is a decision variable.
\newblock {\em Econometrica}, 68(1):119--134, 2000.
\newblock \href {http://dx.doi.org/10.1111/1468-0262.00095}
  {\path{doi:10.1111/1468-0262.00095}}.

\bibitem[KYK{\etalchar{+}}11]{korzhyk2011stackelberg}
Dmytro Korzhyk, Zhengyu Yin, Christopher Kiekintveld, Vincent Conitzer, and
  Milind Tambe.
\newblock Stackelberg vs. {N}ash in security games: An extended investigation
  of interchangeability, equivalence, and uniqueness.
\newblock {\em Journal of Artificial Intelligence Research}, 41:297--327, 2011.
\newblock \href {http://dx.doi.org/10.1613/jair.3269}
  {\path{doi:10.1613/jair.3269}}.

\bibitem[LAH07]{leskovec2007dynamics}
Jure Leskovec, Lada~A Adamic, and Bernardo~A Huberman.
\newblock The dynamics of viral marketing.
\newblock {\em ACM Transactions on the Web}, 1(1):5--es, 2007.
\newblock \href {http://dx.doi.org/0.1145/1232722.1232727}
  {\path{doi:0.1145/1232722.1232727}}.

\bibitem[Lai15]{lai2015giraffe}
Matthew Lai.
\newblock Giraffe: Using deep reinforcement learning to play chess.
\newblock preprint, 2015.
\newblock \href {http://arxiv.org/abs/1509.01549} {\path{arXiv:1509.01549}}.

\bibitem[Lal17]{laland16}
Kevin~N. Laland.
\newblock The origins of language in teaching.
\newblock {\em Psychonomic Bulletin \& Review}, 24:225--231, 07 2017.
\newblock \href {http://dx.doi.org/10.3758/s13423-016-1077-7}
  {\path{doi:10.3758/s13423-016-1077-7}}.

\bibitem[LBK{\etalchar{+}}09]{luck2009flexible}
Michael Luck, Lina Barakat, Jeroen Keppens, Samhar Mahmoud, Simon Miles, Nir
  Oren, Matthew Shaw, and Adel Taweel.
\newblock Flexible behaviour regulation in agent based systems.
\newblock In {\em International Workshop on Collaborative Agents, Research and
  Development}, pages 99--113. Springer, 2009.
\newblock \href {http://dx.doi.org/10.1007/978-3-642-22427-0_8}
  {\path{doi:10.1007/978-3-642-22427-0_8}}.

\bibitem[LCO{\etalchar{+}}16]{luu2016making}
Loi Luu, Duc-Hiep Chu, Hrishi Olickel, Prateek Saxena, and Aquinas Hobor.
\newblock Making smart contracts smarter.
\newblock In {\em Proceedings of the 2016 ACM SIGSAC Conference on Computer and
  Communications Security}, pages 254--269, 2016.
\newblock \href {http://dx.doi.org/10.1145/2976749.2978309}
  {\path{doi:10.1145/2976749.2978309}}.

\bibitem[LCT13]{tomasello2013}
Kristin Liebal, Malinda Carpenter, and Michael Tomasello.
\newblock Young children's understanding of cultural common ground.
\newblock {\em The British Journal of Developmental Psychology}, 31(1):88--96,
  2013.
\newblock \href {http://dx.doi.org/10.1111/j.2044-835X.2012.02080.x}
  {\path{doi:10.1111/j.2044-835X.2012.02080.x}}.

\bibitem[Lew98]{lewis1998designing}
Michael Lewis.
\newblock Designing for human-agent interaction.
\newblock {\em AI Magazine}, 19(2):67--78, 1998.
\newblock \href {http://dx.doi.org/10.1609/aimag.v19i2.1369}
  {\path{doi:10.1609/aimag.v19i2.1369}}.

\bibitem[LFB{\etalchar{+}}18]{letcher2018stable}
Alistair Letcher, Jakob Foerster, David Balduzzi, Tim Rockt{\"a}schel, and
  Shimon Whiteson.
\newblock Stable opponent shaping in differentiable games.
\newblock preprint, 2018.
\newblock \href {http://arxiv.org/abs/1811.08469} {\path{arXiv:1811.08469}}.

\bibitem[LFSC03]{lai2003incentives}
Kevin Lai, Michal Feldman, Ion Stoica, and John Chuang.
\newblock Incentives for cooperation in peer-to-peer networks.
\newblock In {\em Workshop on Economics of Peer-to-Peer Systems}, pages
  1243--1248, 2003.
\newblock URL:
  \url{https://groups.ischool.berkeley.edu/archive/p2pecon/papers/s1-lai.pdf}.

\bibitem[LHDK94]{lee1994prs}
Jaeho Lee, Marcus~J Huber, Edmund~H Durfee, and Patrick~G Kenny.
\newblock Um-prs: An implementation of the procedural reasoning system for
  multirobot applications.
\newblock In {\em AIAA/NASA Conference on Intelligent Robots in Field, Factory,
  Service, and Space}, pages 842--849, 1994.

\bibitem[LHFB19]{lerer2019improving}
Adam Lerer, Hengyuan Hu, Jakob Foerster, and Noam Brown.
\newblock Improving policies via search in cooperative partially observable
  games.
\newblock preprint, 2019.
\newblock \href {http://arxiv.org/abs/1912.02318} {\path{arXiv:1912.02318}}.

\bibitem[LHLG19]{Leibo2019AutocurriculaAT}
Joel~Z. Leibo, E.~Hughes, Marc Lanctot, and Thore Graepel.
\newblock Autocurricula and the emergence of innovation from social
  interaction: A manifesto for multi-agent intelligence research.
\newblock preprint, 2019.
\newblock \href {http://arxiv.org/abs/abs/1903.00742}
  {\path{arXiv:abs/1903.00742}}.

\bibitem[LHTC18]{lazaridou2018emergence}
Angeliki Lazaridou, Karl~Moritz Hermann, Karl Tuyls, and Stephen Clark.
\newblock Emergence of linguistic communication from referential games with
  symbolic and pixel input.
\newblock In {\em International Conference on Learning Representations}, 2018.
\newblock URL: \url{https://openreview.net/forum?id=HJGv1Z-AW}.

\bibitem[LKE{\etalchar{+}}18]{leike2018scalable}
Jan Leike, David Krueger, Tom Everitt, Miljan Martic, Vishal Maini, and Shane
  Legg.
\newblock Scalable agent alignment via reward modeling: a research direction.
\newblock preprint, 2018.
\newblock \href {http://arxiv.org/abs/1811.07871} {\path{arXiv:1811.07871}}.

\bibitem[LLKZ07]{legout2007clustering}
Arnaud Legout, Nikitas Liogkas, Eddie Kohler, and Lixia Zhang.
\newblock Clustering and sharing incentives in {B}it{T}orrent systems.
\newblock {\em ACM SIGMETRICS Performance Evaluation Review}, 35(1):301--312,
  2007.
\newblock \href {http://dx.doi.org/10.1145/1269899.1254919}
  {\path{doi:10.1145/1269899.1254919}}.

\bibitem[LMMZ18]{laengle2018twenty}
Sigifredo Laengle, Nikunja~Mohan Modak, Jose~M Merigo, and Gustavo Zurita.
\newblock Twenty-five years of group decision and negotiation: a bibliometric
  overview.
\newblock {\em Group Decision and Negotiation}, 27:505--542, 2018.
\newblock \href {http://dx.doi.org/10.1007/s10726-018-9582-x}
  {\path{doi:10.1007/s10726-018-9582-x}}.

\bibitem[LMP03]{luck2003agent}
Michael Luck, Peter McBurney, and Chris Preist.
\newblock {\em Agent technology: Enabling next generation computing (A roadmap
  for agent based computing)}.
\newblock AgentLink, 2003.
\newblock URL: \url{http://eprints.soton.ac.uk/id/eprint/257309}.

\bibitem[Loh94]{lohmann1994dynamics}
Susanne Lohmann.
\newblock The dynamics of informational cascades: The {M}onday demonstrations
  in {L}eipzig, {E}ast {G}ermany, 1989-91.
\newblock {\em World Politics}, 47(1):42, 1994.
\newblock \href {http://dx.doi.org/10.2307/2950679}
  {\path{doi:10.2307/2950679}}.

\bibitem[LP16]{LeePenagos2016LearningTC}
Alejandro Lee-Penagos.
\newblock Learning to coordinate: Co-evolution and correlated equilibrium.
\newblock Technical report, Centre for Decision Research \& Experimental
  Economics, 2016.
\newblock URL: \url{https://econpapers.repec.org/RePEc:not:notcdx:2016-11}.

\bibitem[LP20]{lupu2020gifting}
Andrei Lupu and Doina Precup.
\newblock Gifting in multi-agent reinforcement learning.
\newblock In {\em Proceedings of the 19th International Conference on
  Autonomous Agents and MultiAgent Systems}, pages 789--797, 2020.
\newblock \href {http://dx.doi.org/10.5555/3398761.3398855}
  {\path{doi:10.5555/3398761.3398855}}.

\bibitem[LPB16]{lazaridou2016multi}
Angeliki Lazaridou, Alexander Peysakhovich, and Marco Baroni.
\newblock Multi-agent cooperation and the emergence of (natural) language.
\newblock preprint, 2016.
\newblock \href {http://arxiv.org/abs/1612.07182} {\path{arXiv:1612.07182}}.

\bibitem[LS01]{larson2001bargaining}
Kate Larson and Tuomas Sandholm.
\newblock Bargaining with limited computation: Deliberation equilibrium.
\newblock {\em Artificial Intelligence}, 132(2):183--217, 2001.
\newblock \href {http://dx.doi.org/10.1016/S0004-3702(01)00132-1}
  {\path{doi:10.1016/S0004-3702(01)00132-1}}.

\bibitem[LTH03]{lie2003implementing}
David Lie, Chandramohan~A Thekkath, and Mark Horowitz.
\newblock Implementing an untrusted operating system on trusted hardware.
\newblock In {\em Proceedings of the Nineteenth ACM Symposium on Operating
  Systems Principles}, pages 178--192, 2003.
\newblock \href {http://dx.doi.org/10.1145/945445.945463}
  {\path{doi:10.1145/945445.945463}}.

\bibitem[LWQL18]{lu2018bars}
Zhaojun Lu, Qian Wang, Gang Qu, and Zhenglin Liu.
\newblock {BARS}: a blockchain-based anonymous reputation system for trust
  management in {VANET}s.
\newblock In {\em 17th IEEE International Conference On Trust, Security And
  Privacy In Computing And Communications/12th IEEE International Conference On
  Big Data Science And Engineering (TrustCom/BigDataSE)}, pages 98--103. IEEE,
  2018.
\newblock \href {http://dx.doi.org/10.1109/TrustCom/BigDataSE.2018.00025}
  {\path{doi:10.1109/TrustCom/BigDataSE.2018.00025}}.

\bibitem[LYD{\etalchar{+}}17]{lewis2017deal}
Mike Lewis, Denis Yarats, Yann~N Dauphin, Devi Parikh, and Dhruv Batra.
\newblock Deal or no deal? end-to-end learning for negotiation dialogues.
\newblock preprint, 2017.
\newblock \href {http://arxiv.org/abs/1706.05125} {\path{arXiv:1706.05125}}.

\bibitem[LZL{\etalchar{+}}17]{leibo2017multi}
Joel~Z. Leibo, Vinicius Zambaldi, Marc Lanctot, Janusz Marecki, and Thore
  Graepel.
\newblock Multi-agent reinforcement learning in sequential social dilemmas.
\newblock In {\em Proceedings of the 16th International Conference on
  Autonomous Agents and MultiAgent Systems}, 2017.
\newblock \href {http://dx.doi.org/10.5555/3091125.3091194}
  {\path{doi:10.5555/3091125.3091194}}.

\bibitem[MA19]{morsky2019evolution}
Bryce Morsky and Erol Ak{\c{c}}ay.
\newblock Evolution of social norms and correlated equilibria.
\newblock {\em Proceedings of the National Academy of Sciences},
  116(18):8834--8839, 2019.
\newblock \href {http://dx.doi.org/10.1073/pnas.1817095116}
  {\path{doi:10.1073/pnas.1817095116}}.

\bibitem[Mas08]{maskin2008mechanism}
Eric~S Maskin.
\newblock Mechanism design: How to implement social goals.
\newblock {\em American Economic Review}, 98(3):567--76, 2008.
\newblock \href {http://dx.doi.org/10.1257/aer.98.3.567}
  {\path{doi:10.1257/aer.98.3.567}}.

\bibitem[MBLA96]{minar1996swarm}
Nelson Minar, Roger Burkhart, Chris Langton, and Manor Askenazi.
\newblock The swarm simulation system: A toolkit for building multi-agent
  simulations.
\newblock Technical report, Sante Fe University, 1996.
\newblock URL:
  \url{https://www.santafe.edu/research/results/working-papers/the-swarm-simulation-system-a-toolkit-for-building}.

\bibitem[MBMV06]{massaguer2006multi}
Daniel Massaguer, Vidhya Balasubramanian, Sharad Mehrotra, and Nalini
  Venkatasubramanian.
\newblock Multi-agent simulation of disaster response.
\newblock In {\em Proceedings of the First International Workshop on Agent
  Technology for Disaster Management}, pages 124--130, 2006.

\bibitem[MDS95]{mayer1995integrative}
Roger~C Mayer, James~H Davis, and F~David Schoorman.
\newblock An integrative model of organizational trust.
\newblock {\em Academy of Management Review}, 20(3):709--734, 1995.
\newblock \href {http://dx.doi.org/10.5465/amr.1995.9508080335}
  {\path{doi:10.5465/amr.1995.9508080335}}.

\bibitem[Mer11]{doi:10.1177/1468797611432040}
Stephanie Merchant.
\newblock Negotiating underwater space: The sensorium, the body and the
  practice of scuba-diving.
\newblock {\em Tourist Studies}, 11(3):215--234, 2011.
\newblock \href {http://dx.doi.org/10.1177/1468797611432040}
  {\path{doi:10.1177/1468797611432040}}.

\bibitem[MGA{\etalchar{+}}05]{mueller2005evolution}
Ulrich~G Mueller, Nicole~M Gerardo, Duur~K Aanen, Diana~L Six, and Ted~R
  Schultz.
\newblock The evolution of agriculture in insects.
\newblock {\em Annual Review of Ecology, Evolution, and Systematics},
  36:563--595, 2005.
\newblock \href {http://dx.doi.org/10.1146/annurev.ecolsys.36.102003.152626}
  {\path{doi:10.1146/annurev.ecolsys.36.102003.152626}}.

\bibitem[MGM03]{marti2003identity}
Sergio Marti and Hector Garcia-Molina.
\newblock Identity crisis: anonymity vs reputation in {P2P} systems.
\newblock In {\em Proceedings Third International Conference on Peer-to-Peer
  Computing (P2P2003)}, pages 134--141. IEEE, 2003.
\newblock \href {http://dx.doi.org/10.1109/PTP.2003.1231513}
  {\path{doi:10.1109/PTP.2003.1231513}}.

\bibitem[MGM06]{marti2006taxonomy}
Sergio Marti and Hector Garcia-Molina.
\newblock Taxonomy of trust: Categorizing {P2P} reputation systems.
\newblock {\em Computer Networks}, 50(4):472--484, 2006.
\newblock \href {http://dx.doi.org/10.1016/j.comnet.2005.07.011}
  {\path{doi:10.1016/j.comnet.2005.07.011}}.

\bibitem[Min88]{minsky1988society}
Marvin Minsky.
\newblock {\em Society of mind}.
\newblock Simon and Schuster, New York, 1988.

\bibitem[MKS{\etalchar{+}}15]{mnih2015human}
Volodymyr Mnih, Koray Kavukcuoglu, David Silver, Andrei~A Rusu, Joel Veness,
  Marc~G Bellemare, Alex Graves, Martin Riedmiller, Andreas~K Fidjeland, Georg
  Ostrovski, Stig Petersen, Charles Beattie, Amir Sadik, Ioannis Antonoglou,
  Helen King, Dharshan Kumaran, Daan Wierstra, Shane Legg, and Demis Hassabis.
\newblock Human-level control through deep reinforcement learning.
\newblock {\em Nature}, 518:529--533, 2015.
\newblock \href {http://dx.doi.org/doi.org/10.1038/nature14236}
  {\path{doi:doi.org/10.1038/nature14236}}.

\bibitem[ML17]{menon2017computational}
Vijay Menon and Kate Larson.
\newblock Computational aspects of strategic behaviour in elections with
  top-truncated ballots.
\newblock {\em Autonomous Agents and Multi-Agent Systems}, 31:1506--1547, 2017.
\newblock \href {http://dx.doi.org/10.1007/s10458-017-9369-5}
  {\path{doi:10.1007/s10458-017-9369-5}}.

\bibitem[MMDVP19]{morris2019norm}
Andreasa Morris-Martin, Marina De~Vos, and Julian Padget.
\newblock Norm emergence in multiagent systems: a viewpoint paper.
\newblock {\em Autonomous Agents and Multi-Agent Systems}, 33:706--749, 2019.
\newblock \href {http://dx.doi.org/10.1007/s10458-019-09422-0}
  {\path{doi:10.1007/s10458-019-09422-0}}.

\bibitem[MMG16]{melo2016people}
Celso~De Melo, Stacy Marsella, and Jonathan Gratch.
\newblock People do not feel guilty about exploiting machines.
\newblock {\em ACM Transactions on Computer-Human Interaction}, 23(2):1--17,
  2016.
\newblock \href {http://dx.doi.org/10.1145/2890495}
  {\path{doi:10.1145/2890495}}.

\bibitem[MMR{\etalchar{+}}17]{McMahan2017communication}
H.~Brendan McMahan, Eider Moore, Daniel Ramage, Seth Hampson, and Blaise
  Aguera~y Arcas.
\newblock Communication-efficient learning of deep networks from decentralized
  data.
\newblock In {\em Proceedings of the 20th International Conference on
  Artificial Intelligence and Statistics}, volume~54 of {\em Proceedings of
  Machine Learning Research}, pages 1273--1282, 2017.
\newblock URL: \url{http://proceedings.mlr.press/v54/mcmahan17a.html}.

\bibitem[Mor14]{morrow2014order}
James~D Morrow.
\newblock {\em Order within anarchy: The laws of war as an international
  institution}.
\newblock Cambridge University Press, New York, 2014.

\bibitem[MPW02]{mcburney2002desiderata}
Peter McBurney, Simon Parsons, and Michael Wooldridge.
\newblock Desiderata for agent argumentation protocols.
\newblock In {\em Proceedings of the First International Joint Conference on
  Autonomous Agents and Multiagent Systems: part 1}, pages 402--409, 2002.
\newblock \href {http://dx.doi.org/10.1145/544741.544836}
  {\path{doi:10.1145/544741.544836}}.

\bibitem[MS05]{mitusch2005mediation}
Kay Mitusch and Roland Strausz.
\newblock Mediation in situations of conflict and limited commitment.
\newblock {\em Journal of Law, Economics, and Organization}, 21(2):467--500,
  2005.
\newblock \href {http://dx.doi.org/10.1093/jleo/ewi018}
  {\path{doi:10.1093/jleo/ewi018}}.

\bibitem[MS06]{mailath2006repeated}
George~J Mailath and Larry Samuelson.
\newblock {\em Repeated games and reputations: long-run relationships}.
\newblock Oxford University Press, Oxford, 2006.
\newblock \href {http://dx.doi.org/10.1093/acprof:oso/9780195300796.001.0001}
  {\path{doi:10.1093/acprof:oso/9780195300796.001.0001}}.

\bibitem[MSB{\etalchar{+}}17]{moravvcik2017deepstack}
Matej Morav{\v{c}}{\'\i}k, Martin Schmid, Neil Burch, Viliam Lis{\`y}, Dustin
  Morrill, Nolan Bard, Trevor Davis, Kevin Waugh, Michael Johanson, and Michael
  Bowling.
\newblock Deepstack: Expert-level artificial intelligence in heads-up no-limit
  poker.
\newblock {\em Science}, 356(6337):508--513, 2017.
\newblock \href {http://dx.doi.org/10.1126/science.aam6960}
  {\path{doi:10.1126/science.aam6960}}.

\bibitem[MSJ98]{matos1998determining}
Noyda Matos, Carles Sierra, and Nicholas~R Jennings.
\newblock Determining successful negotiation strategies: An evolutionary
  approach.
\newblock In {\em Proceedings International Conference on Multi Agent Systems},
  pages 182--189. IEEE, 1998.
\newblock \href {http://dx.doi.org/10.1109/ICMAS.1998.699048}
  {\path{doi:10.1109/ICMAS.1998.699048}}.

\bibitem[MSV93]{10.2307/2117699}
Kevin~M. Murphy, Andrei Shleifer, and Robert~W. Vishny.
\newblock Why is rent-seeking so costly to growth?
\newblock {\em The American Economic Review}, 83(2):409--414, 1993.
\newblock URL: \url{http://www.jstor.org/stable/2117699}.

\bibitem[MT95]{moses1995artificial}
Yoram Moses and Moshe Tennenholtz.
\newblock Artificial social systems.
\newblock {\em Computers and Artificial Intelligence}, 14:533--562, 1995.

\bibitem[MT04]{monderer2004k}
Dov Monderer and Moshe Tennenholtz.
\newblock {K}-implementation.
\newblock {\em Journal of Artificial Intelligence Research}, 21:37--62, 2004.
\newblock \href {http://dx.doi.org/doi.org/10.1613/jair.1231}
  {\path{doi:doi.org/10.1613/jair.1231}}.

\bibitem[MT09]{monderer2009strong}
Dov Monderer and Moshe Tennenholtz.
\newblock Strong mediated equilibrium.
\newblock {\em Artificial Intelligence}, 173(1):180--195, 2009.
\newblock \href {http://dx.doi.org/10.1016/j.artint.2008.10.005}
  {\path{doi:10.1016/j.artint.2008.10.005}}.

\bibitem[Mue96]{mueller1996negotiation}
H~Juergen Mueller.
\newblock Negotiation principles.
\newblock In Greg M~P O'Hare and Nicholas~R Jennings, editors, {\em Foundations
  of Distributed Artificial Intelligence}, pages 211--230. John Wiley \& Sons,
  New York, 1996.

\bibitem[MZ17]{Mohassel2017secure}
Payman Mohassel and Yupeng Zhang.
\newblock {SecureML}: A system for scalable privacy-preserving machine
  learning.
\newblock In {\em IEEE Symposium on Security and Privacy}, pages 19--38. IEEE,
  2017.
\newblock \href {http://dx.doi.org/10.1109/SP.2017.12}
  {\path{doi:10.1109/SP.2017.12}}.

\bibitem[MZX{\etalchar{+}}20]{mao2020learning}
Hangyu Mao, Zhengchao Zhang, Zhen Xiao, Zhibo Gong, and Yan Ni.
\newblock Learning agent communication under limited bandwidth by message
  pruning.
\newblock In {\em Proceedings of the AAAI Conference on Artificial
  Intelligence}, volume~34, pages 5142--5149, 2020.
\newblock \href {http://dx.doi.org/10.1609/aaai.v34i04.5957}
  {\path{doi:10.1609/aaai.v34i04.5957}}.

\bibitem[NAP16]{Narisimhan2016Automated}
Harikrishna Narasimhan, Shivani Agarwal, and David~C. Parkes.
\newblock Automated mechanism design without money via machine learning.
\newblock In {\em Proceedings of the 25th International Joint Conference on
  Artificial Intelligence}, page 433–439, 2016.
\newblock URL: \url{https://www.ijcai.org/Abstract/16/068}.

\bibitem[{Nat}20]{natpopvote2020}
{National Popular Vote Inc.}
\newblock Agreement among the states to elect the president by national popular
  vote, 2020.
\newblock URL:
  \url{https://www.nationalpopularvote.com/sites/default/files/1-pager-npv-v203-2020-11-27.pdf}.

\bibitem[NELJ20]{ndousse2020multi}
Kamal Ndousse, Douglas Eck, Sergey Levine, and Natasha Jaques.
\newblock Learning social learning, 2020.
\newblock NeurIPS Workshop on Cooperative AI.

\bibitem[NHW{\etalchar{+}}19]{ning2019review}
Yishuang Ning, Sheng He, Zhiyong Wu, Chunxiao Xing, and Liang-Jie Zhang.
\newblock A review of deep learning based speech synthesis.
\newblock {\em Applied Sciences}, 9(19):4050, 2019.
\newblock \href {http://dx.doi.org/10.3390/app9194050}
  {\path{doi:10.3390/app9194050}}.

\bibitem[NJ50]{nash1950bargaining}
John~F Nash~Jr.
\newblock The bargaining problem.
\newblock {\em Econometrica: Journal of the econometric society},
  18(2):155--162, 1950.
\newblock \href {http://dx.doi.org/10.2307/1907266}
  {\path{doi:10.2307/1907266}}.

\bibitem[NJ06]{narayanan2006learning}
Vidya Narayanan and Nicholas~R Jennings.
\newblock Learning to negotiate optimally in non-stationary environments.
\newblock In {\em International Workshop on Cooperative Information Agents},
  pages 288--300. Springer, 2006.
\newblock \href {http://dx.doi.org/10.1007/11839354_2}
  {\path{doi:10.1007/11839354_2}}.

\bibitem[NLRS15]{nikolaidis2015improved}
Stefanos Nikolaidis, Przemyslaw Lasota, Ramya Ramakrishnan, and Julie Shah.
\newblock Improved human--robot team performance through cross-training, an
  approach inspired by human team training practices.
\newblock {\em The International Journal of Robotics Research},
  34(14):1711--1730, 2015.
\newblock \href {http://dx.doi.org/10.1177/0278364915609673}
  {\path{doi:10.1177/0278364915609673}}.

\bibitem[Nor93]{north1993institutions}
Douglass~C North.
\newblock Institutions and credible commitment.
\newblock {\em Journal of Institutional and Theoretical Economics
  (JITE)/Zeitschrift f{\"u}r die gesamte Staatswissenschaft}, 149(1):11--23,
  1993.
\newblock URL: \url{https://www.jstor.org/stable/40751576}.

\bibitem[Nor94]{norman1994might}
Donald~A Norman.
\newblock How might people interact with agents.
\newblock {\em Communications of the ACM}, 37(7):68--71, 1994.
\newblock \href {http://dx.doi.org/10.1145/176789.176796}
  {\path{doi:10.1145/176789.176796}}.

\bibitem[NP08]{neville2008presage}
Brendan Neville and Jeremy Pitt.
\newblock {PRESAGE}: A programming environment for the simulation of agent
  societies.
\newblock In {\em International Workshop on Programming Multi-Agent Systems},
  pages 88--103. Springer, 2008.
\newblock \href {http://dx.doi.org/10.1007/978-3-642-03278-3_6}
  {\path{doi:10.1007/978-3-642-03278-3_6}}.

\bibitem[NR{\etalchar{+}}00]{ng2000algorithms}
Andrew~Y Ng, Stuart~J Russell, et~al.
\newblock Algorithms for inverse reinforcement learning.
\newblock In {\em Proceedings of the 17th International Conference onn Machine
  Learning}, pages 663--670, 2000.

\bibitem[NR01]{nisan2001algorithmic}
Noam Nisan and Amir Ronen.
\newblock Algorithmic mechanism design.
\newblock {\em Games and Economic Behavior}, 35(1-2):166--196, 2001.
\newblock \href {http://dx.doi.org/10.1006/game.1999.0790}
  {\path{doi:10.1006/game.1999.0790}}.

\bibitem[NR07]{nisan2007computationally}
Noam Nisan and Amir Ronen.
\newblock Computationally feasible {VCG} mechanisms.
\newblock {\em Journal of Artificial Intelligence Research}, 29:19--47, 2007.
\newblock \href {http://dx.doi.org/10.1613/jair.2046}
  {\path{doi:10.1613/jair.2046}}.

\bibitem[NS98]{nowak1998evolution}
Martin~A Nowak and Karl Sigmund.
\newblock Evolution of indirect reciprocity by image scoring.
\newblock {\em Nature}, 393:573--577, 1998.
\newblock \href {http://dx.doi.org/10.1038/31225} {\path{doi:10.1038/31225}}.

\bibitem[NSA{\etalchar{+}}19]{nassif2019speech}
Ali~Bou Nassif, Ismail Shahin, Imtinan Attili, Mohammad Azzeh, and Khaled
  Shaalan.
\newblock Speech recognition using deep neural networks: A systematic review.
\newblock {\em IEEE Access}, 7:19143--19165, 2019.
\newblock \href {http://dx.doi.org/10.1109/ACCESS.2019.2896880}
  {\path{doi:10.1109/ACCESS.2019.2896880}}.

\bibitem[NWW{\etalchar{+}}09]{north2009violence}
Douglass~C North, John~Joseph Wallis, Barry~R Weingast, et~al.
\newblock {\em Violence and social orders: A conceptual framework for
  interpreting recorded human history}.
\newblock Cambridge University Press, New York, 2009.

\bibitem[NZdS{\etalchar{+}}16]{nallapati2016abstractive}
Ramesh Nallapati, Bowen Zhou, Cícero~Noguiera dos Santos, Çaglar Gülçehre,
  and Bing Xiang.
\newblock Abstractive text summarization using sequence-to-sequence {RNN}s and
  beyond.
\newblock In {\em Proceedings of The 20th SIGNLL Conference on Computational
  Natural Language Learning}, pages 280--290, Berlin, Germany, 2016.
  Association for Computational Linguistics.
\newblock \href {http://dx.doi.org/10.18653/v1/K16-1028}
  {\path{doi:10.18653/v1/K16-1028}}.

\bibitem[ODZ{\etalchar{+}}16]{oord2016wavenet}
Aaron van~den Oord, Sander Dieleman, Heiga Zen, Karen Simonyan, Oriol Vinyals,
  Alex Graves, Nal Kalchbrenner, Andrew Senior, and Koray Kavukcuoglu.
\newblock Wave{N}et: A generative model for raw audio.
\newblock preprint, 2016.
\newblock \href {http://arxiv.org/abs/1609.03499} {\path{arXiv:1609.03499}}.

\bibitem[OJ96]{o1996foundations}
Greg~MP O'Hare and Nicholas~R Jennings, editors.
\newblock {\em Foundations of Distributed Artificial Intelligence}.
\newblock John Wiley \& Sons, New York, 1996.

\bibitem[OKL{\etalchar{+}}19]{omidshafiei2019learning}
Shayegan Omidshafiei, Dong-Ki Kim, Miao Liu, Gerald Tesauro, Matthew Riemer,
  Christopher Amato, Murray Campbell, and Jonathan~P How.
\newblock Learning to teach in cooperative multiagent reinforcement learning.
\newblock In {\em Proceedings of the AAAI Conference on Artificial
  Intelligence}, volume~33, pages 6128--6136, 2019.
\newblock \href {http://dx.doi.org/10.1609/aaai.v33i01.33016128}
  {\path{doi:10.1609/aaai.v33i01.33016128}}.

\bibitem[OKSSP15]{o2015ease}
Cathleen O’Grady, Christian Kliesch, Kenny Smith, and Thomas~C
  Scott-Phillips.
\newblock The ease and extent of recursive mindreading, across implicit and
  explicit tasks.
\newblock {\em Evolution and Human Behavior}, 36(4):313--322, 2015.
\newblock \href {http://dx.doi.org/10.1016/j.evolhumbehav.2015.01.004}
  {\path{doi:10.1016/j.evolhumbehav.2015.01.004}}.

\bibitem[O'L19]{o2019google}
Daniel~E O'Leary.
\newblock Google's duplex: Pretending to be human.
\newblock {\em Intelligent Systems in Accounting, Finance and Management},
  26(1):46--53, 2019.
\newblock \href {http://dx.doi.org/10.1002/isaf.1443}
  {\path{doi:10.1002/isaf.1443}}.

\bibitem[Oli96]{oliver1996machine}
Jim~R Oliver.
\newblock A machine-learning approach to automated negotiation and prospects
  for electronic commerce.
\newblock {\em Journal of management information systems}, 13(3):83--112, 1996.
\newblock \href {http://dx.doi.org/10.1080/07421222.1996.11518135}
  {\path{doi:10.1080/07421222.1996.11518135}}.

\bibitem[Ols82]{10.2307/j.ctt1nprdd}
Mancur Olson.
\newblock {\em The Rise and Decline of Nations: Economic Growth, Stagflation,
  and Social Rigidities}.
\newblock Yale University Press, New Haven, 1982.
\newblock URL: \url{http://www.jstor.org/stable/j.ctt1nprdd}.

\bibitem[Ord86]{ordeshook1986game}
Peter~C Ordeshook.
\newblock {\em Game theory and political theory: An introduction}.
\newblock Cambridge University Press, Cambridge, UK, 1986.

\bibitem[OSFM07]{olfati2007consensus}
Reza Olfati-Saber, J~Alex Fax, and Richard~M Murray.
\newblock Consensus and cooperation in networked multi-agent systems.
\newblock {\em Proceedings of the IEEE}, 95(1):215--233, 2007.
\newblock \href {http://dx.doi.org/10.1109/JPROC.2006.887293}
  {\path{doi:10.1109/JPROC.2006.887293}}.

\bibitem[OSJ{\etalchar{+}}18]{olah2018building}
Chris Olah, Arvind Satyanarayan, Ian Johnson, Shan Carter, Ludwig Schubert,
  Katherine Ye, and Alexander Mordvintsev.
\newblock The building blocks of interpretability.
\newblock {\em Distill}, 2018.
\newblock \href {http://dx.doi.org/10.23915/distill.00010}
  {\path{doi:10.23915/distill.00010}}.

\bibitem[Ost98]{ostrom1998behavioral}
Elinor Ostrom.
\newblock A behavioral approach to the rational choice theory of collective
  action: Presidential address, {A}merican {P}olitical {S}cience {A}ssociation,
  1997.
\newblock {\em American Political Science Review}, 92(1):1--22, 1998.
\newblock \href {http://dx.doi.org/10.2307/2585925}
  {\path{doi:10.2307/2585925}}.

\bibitem[Ost00]{10.2307/2646923}
Elinor Ostrom.
\newblock Collective action and the evolution of social norms.
\newblock {\em The Journal of Economic Perspectives}, 14(3):137--158, 2000.
\newblock URL: \url{http://www.jstor.org/stable/2646923}, \href
  {http://dx.doi.org/10.1257/jep.14.3.137} {\path{doi:10.1257/jep.14.3.137}}.

\bibitem[Par93]{parkhe1993strategic}
Arvind Parkhe.
\newblock Strategic alliance structuring: A game theoretic and transaction cost
  examination of interfirm cooperation.
\newblock {\em Academy of Management Journal}, 36(4):794--829, 1993.
\newblock \href {http://dx.doi.org/10.5465/256759} {\path{doi:10.5465/256759}}.

\bibitem[PBGRLS17]{pareda2017}
María Pereda, Pablo Brañas-Garza, Ismael Rodriguez-Lara, and Angel Sánchez.
\newblock The emergence of altruism as a social norm.
\newblock {\em Scientific Reports}, 7:9684, 08 2017.
\newblock \href {http://dx.doi.org/10.1038/s41598-017-07712-9}
  {\path{doi:10.1038/s41598-017-07712-9}}.

\bibitem[PDB99]{park1999adaptive}
Sunju Park, Edmund~H Durfee, and William~P Birmingham.
\newblock An adaptive agent bidding strategy based on stochastic modeling.
\newblock In {\em Proceedings of the Third Annual Conference on Autonomous
  Agents}, pages 147--153, 1999.
\newblock \href {http://dx.doi.org/10.1145/301136.301181}
  {\path{doi:10.1145/301136.301181}}.

\bibitem[Pes00]{Pesendorfer2000ASO}
Martin Pesendorfer.
\newblock A study of collusion in first-price auctions.
\newblock {\em The Review of Economic Studies}, 67(3):381--411, 2000.
\newblock \href {http://dx.doi.org/10.1111/1467-937X.00136}
  {\path{doi:10.1111/1467-937X.00136}}.

\bibitem[PFR09]{pipattanasomporn2009multi}
Manisa Pipattanasomporn, Hassan Feroze, and Saifur Rahman.
\newblock Multi-agent systems in a distributed smart grid: Design and
  implementation.
\newblock In {\em IEEE/PES Power Systems Conference and Exposition}, pages
  1--8. IEEE, 2009.
\newblock \href {http://dx.doi.org/10.1109/PSCE.2009.4840087}
  {\path{doi:10.1109/PSCE.2009.4840087}}.

\bibitem[PG04]{Pickering2004Toward}
Martin~J. Pickering and Simon Garrod.
\newblock Toward a mechanistic psychology of dialogue.
\newblock {\em Behavioral and Brain Sciences}, 27:169--190, 2004.
\newblock \href {http://dx.doi.org/10.1017/S0140525X04000056}
  {\path{doi:10.1017/S0140525X04000056}}.

\bibitem[PKSA06]{pitt2006voting}
Jeremy Pitt, Lloyd Kamara, Marek Sergot, and Alexander Artikis.
\newblock Voting in multi-agent systems.
\newblock {\em The Computer Journal}, 49(2):156--170, 2006.
\newblock \href {http://dx.doi.org/10.1093/comjnl/bxh164}
  {\path{doi:10.1093/comjnl/bxh164}}.

\bibitem[PL05]{panait2005cooperative}
Liviu Panait and Sean Luke.
\newblock Cooperative multi-agent learning: The state of the art.
\newblock {\em Autonomous Agents and Multi-Agent Systems}, 11:387--434, 2005.
\newblock \href {http://dx.doi.org/10.1007/s10458-005-2631-2}
  {\path{doi:10.1007/s10458-005-2631-2}}.

\bibitem[PLB{\etalchar{+}}19]{paquette2019press}
Philip Paquette, Yuchen Lu, Steven Bocco, Max~O. Smith, Satya Ortiz-Gagne,
  Jonathan~K. Kummerfeld, Satinder Singh, Joelle Pineau, and Aaron Courville.
\newblock No press diplomacy: Modeling multi-agent gameplay.
\newblock preprint, 2019.
\newblock \href {http://arxiv.org/abs/1909.02128} {\path{arXiv:1909.02128}}.

\bibitem[PNC17]{10.1093/esr/jcx072}
Wojtek Przepiorka, Lukas Norbutas, and Rense Corten.
\newblock Order without law: Reputation promotes cooperation in a cryptomarket
  for illegal drugs.
\newblock {\em European Sociological Review}, 33(6):752--764, 10 2017.
\newblock \href {http://dx.doi.org/10.1093/esr/jcx072}
  {\path{doi:10.1093/esr/jcx072}}.

\bibitem[Pol19]{polis}
Polis.
\newblock Polis: Input crowd, output meaning, 2019.
\newblock URL: \url{https://pol.is/home}.

\bibitem[Pom88]{pomerleau1989alvinn}
Dean~A. Pomerleau.
\newblock {ALVINN}: An autonomous land vehicle in a neural network.
\newblock In {\em Advances in Neural Information Processing Systems}, volume~1,
  pages 305--313, 1988.
\newblock URL:
  \url{http://papers.nips.cc/paper/95-alvinn-an-autonomous-land-vehicle-in-a-neural-network.pdf}.

\bibitem[Pow99]{powell1999shadow}
Robert Powell.
\newblock {\em In the shadow of power: States and strategies in international
  politics}.
\newblock Princeton University Press, Princeton, NJ, 1999.

\bibitem[Pow06]{powell2006war}
Robert Powell.
\newblock War as a commitment problem.
\newblock {\em International Organization}, pages 169--203, 2006.
\newblock \href {http://dx.doi.org/10.1017/S0020818306060061}
  {\path{doi:10.1017/S0020818306060061}}.

\bibitem[PPM{\etalchar{+}}08]{paruchuri2008playing}
Praveen Paruchuri, Jonathan~P Pearce, Janusz Marecki, Milind Tambe, Fernando
  Ordonez, and Sarit Kraus.
\newblock Playing games for security: An efficient exact algorithm for solving
  {B}ayesian {S}tackelberg games.
\newblock In {\em Proceedings of the 7th International Joint Conference on
  Autonomous Agents and Multiagent Systems}, volume~2, pages 895--902.
  International Foundation for Autonomous Agents and Multiagent Systems, 2008.
\newblock URL:
  \url{http://ifaamas.org/Proceedings/aamas08/proceedings/pdf/paper/AAMAS08_0057.pdf}.

\bibitem[PSM13]{pinyol2013computational}
Isaac Pinyol and Jordi Sabater-Mir.
\newblock Computational trust and reputation models for open multi-agent
  systems: a review.
\newblock {\em Artificial Intelligence Review}, 40(1):1--25, 2013.
\newblock \href {http://dx.doi.org/10.1007/s10462-011-9277-z}
  {\path{doi:10.1007/s10462-011-9277-z}}.

\bibitem[PSM14]{pennington2014glove}
Jeffrey Pennington, Richard Socher, and Christopher~D Manning.
\newblock Glo{V}e: Global vectors for word representation.
\newblock In {\em Proceedings of the 2014 Conference on Empirical Methods in
  Natural Language Processing (EMNLP)}, pages 1532--1543, 2014.
\newblock \href {http://dx.doi.org/10.3115/v1/D14-1162}
  {\path{doi:10.3115/v1/D14-1162}}.

\bibitem[PW02]{parsons2002game}
Simon Parsons and Michael Wooldridge.
\newblock Game theory and decision theory in multi-agent systems.
\newblock {\em Autonomous Agents and Multi-Agent Systems}, 5(3):243--254, 2002.
\newblock \href {http://dx.doi.org/10.1023/A:1015575522401}
  {\path{doi:10.1023/A:1015575522401}}.

\bibitem[PW05]{padgham2005developing}
Lin Padgham and Michael Winikoff.
\newblock {\em Developing intelligent agent systems: A practical guide}.
\newblock John Wiley \& Sons, West Sussex, 2005.

\bibitem[PW15]{wellman2015economic}
David Parkes and Michael Wellman.
\newblock Economic reasoning and artificial intelligence.
\newblock {\em Science}, 349(6245):267--272, 2015.
\newblock \href {http://dx.doi.org/10.1126/science.aaa8403}
  {\path{doi:10.1126/science.aaa8403}}.

\bibitem[QMG{\etalchar{+}}19]{qin2019adversarial}
Chongli Qin, James Martens, Sven Gowal, Dilip Krishnan, Krishnamurthy
  Dvijotham, Alhussein Fawzi, Soham De, Robert Stanforth, and Pushmeet Kohli.
\newblock Adversarial robustness through local linearization.
\newblock In {\em Advances in Neural Information Processing Systems},
  volume~32, pages 13847--13856, 2019.
\newblock URL:
  \url{https://papers.nips.cc/paper/2019/file/0defd533d51ed0a10c5c9dbf93ee78a5-Paper.pdf}.

\bibitem[Rap66]{rapoport1966taxonomy}
Anatol Rapoport.
\newblock A taxonomy of 2$\times$2 games.
\newblock {\em General Systems}, 11:203--214, 1966.

\bibitem[RD15]{richtel2015google}
Matt Richtel and Conor Dougherty.
\newblock Google’s driverless cars run into problem: Cars with drivers.
\newblock {\em The New York Times}, page~A1, 2015.
\newblock URL:
  \url{https://www.nytimes.com/2015/09/02/technology/personaltech/google-says-its-not-the-driverless-cars-fault-its-other-drivers.html}.

\bibitem[RDT15]{russell2015research}
Stuart Russell, Daniel Dewey, and Max Tegmark.
\newblock Research priorities for robust and beneficial artificial
  intelligence.
\newblock {\em AI Magazine}, 36(4):105--114, 2015.
\newblock \href {http://dx.doi.org/10.1609/aimag.v36i4.2577}
  {\path{doi:10.1609/aimag.v36i4.2577}}.

\bibitem[Rey59]{reynolds1959burning}
Winston~A Reynolds.
\newblock The burning ships of {H}ern{\'a}n {C}ort{\'e}s.
\newblock {\em Hispania}, pages 317--324, 1959.
\newblock \href {http://dx.doi.org/10.2307/335707} {\path{doi:10.2307/335707}}.

\bibitem[Rey87]{reynolds1987flocks}
Craig~W Reynolds.
\newblock Flocks, herds and schools: A distributed behavioral model.
\newblock In {\em Proceedings of the 14th Annual Conference on Computer
  Graphics and Interactive Techniques}, pages 25--34, 1987.
\newblock \href {http://dx.doi.org/10.1145/37401.37406}
  {\path{doi:10.1145/37401.37406}}.

\bibitem[RG05]{robinson2005topology}
David Robinson and David Goforth.
\newblock The topology of the 2x2 games: A new periodic table, 01 2005.
\newblock \href {http://dx.doi.org/10.4324/9780203340271}
  {\path{doi:10.4324/9780203340271}}.

\bibitem[RGN12]{rand2012spontaneous}
David~G Rand, Joshua~D Greene, and Martin~A Nowak.
\newblock Spontaneous giving and calculated greed.
\newblock {\em Nature}, 489:427--430, 2012.
\newblock \href {http://dx.doi.org/10.1038/nature11467}
  {\path{doi:10.1038/nature11467}}.

\bibitem[RHJ04]{ramchurn2004trust}
Sarvapali~D Ramchurn, Dong Huynh, and Nicholas~R Jennings.
\newblock Trust in multi-agent systems.
\newblock {\em The Knowledge Engineering Review}, 19(1):1--25, 2004.
\newblock \href {http://dx.doi.org/10.1017/S026988890400011}
  {\path{doi:10.1017/S026988890400011}}.

\bibitem[RKG{\etalchar{+}}18]{rajeswaran2018learning}
Aravind Rajeswaran, Vikash Kumar, Abhishek Gupta, Giulia Vezzani, John
  Schulman, Emanuel Todorov, and Sergey Levine.
\newblock Learning complex dexterous manipulation with deep reinforcement
  learning and demonstrations.
\newblock In {\em Proceedings of Robotics: Science and Systems}, volume~14,
  2018.
\newblock \href {http://dx.doi.org/10.15607/RSS.2018.XIV.049}
  {\path{doi:10.15607/RSS.2018.XIV.049}}.

\bibitem[RN13]{rand2013human}
David~G Rand and Martin~A Nowak.
\newblock Human cooperation.
\newblock {\em Trends in Cognitive sciences}, 17(8):413--425, 2013.
\newblock \href {http://dx.doi.org/10.1016/j.tics.2013.06.003}
  {\path{doi:10.1016/j.tics.2013.06.003}}.

\bibitem[Rob79]{roberts1979characterization}
Kevin Roberts.
\newblock The characterization of implementable choice rules.
\newblock In Jean-Jacques Laffont, editor, {\em Aggregation and revelation of
  preferences}, pages 321--348. North-Holland, 1979.

\bibitem[Rob92]{robinson1992regulation}
Gene~E Robinson.
\newblock Regulation of division of labor in insect societies.
\newblock {\em Annual Review of Entomology}, 37(1):637--665, 1992.
\newblock \href {http://dx.doi.org/10.1146/annurev.en.37.010192.003225}
  {\path{doi:10.1146/annurev.en.37.010192.003225}}.

\bibitem[Rob04]{robertson2004multi}
David Robertson.
\newblock Multi-agent coordination as distributed logic programming.
\newblock In {\em International Conference on Logic Programming}, pages
  416--430. Springer, 2004.
\newblock \href {http://dx.doi.org/10.1007/978-3-540-27775-0_29}
  {\path{doi:10.1007/978-3-540-27775-0_29}}.

\bibitem[Roc01]{rock2001securities}
Edward Rock.
\newblock Securities regulation as lobster trap: A credible commitment theory
  of mandatory disclosure.
\newblock {\em Cardozo Law Review}, 23:675, 2001.

\bibitem[Rou10]{roughgarden2010algorithmic}
Tim Roughgarden.
\newblock Algorithmic game theory.
\newblock {\em Communications of the ACM}, 53(7):78--86, 2010.
\newblock \href {http://dx.doi.org/10.1145/1785414.1785439}
  {\path{doi:10.1145/1785414.1785439}}.

\bibitem[RPKT{\etalchar{+}}14]{rand2014social}
David~G Rand, Alexander Peysakhovich, Gordon~T Kraft-Todd, George~E Newman,
  Owen Wurzbacher, Martin~A Nowak, and Joshua~D Greene.
\newblock Social heuristics shape intuitive cooperation.
\newblock {\em Nature Communications}, 5:3677, 2014.
\newblock \href {http://dx.doi.org/10.1038/ncomms4677}
  {\path{doi:10.1038/ncomms4677}}.

\bibitem[RS11]{rilling2011neuroscience}
James~K Rilling and Alan~G Sanfey.
\newblock The neuroscience of social decision-making.
\newblock {\em Annual Review of Psychology}, 62:23--48, 2011.
\newblock \href {http://dx.doi.org/10.1146/annurev.psych.121208.131647}
  {\path{doi:10.1146/annurev.psych.121208.131647}}.

\bibitem[RT94]{Traum1994Computational}
David Rood~Traum.
\newblock {\em A Computational Theory of Grounding in Natural Language
  Conversation}.
\newblock PhD thesis, University of Rochester, 1994.
\newblock URL: \url{http://hdl.handle.net/1802/809}.

\bibitem[Rus19]{russell2019human}
Stuart Russell.
\newblock {\em Human compatible: Artificial intelligence and the problem of
  control}.
\newblock Penguin, New York, 2019.

\bibitem[RZ94]{rosenschein1994rules}
Jeffrey~S Rosenschein and Gilad Zlotkin.
\newblock {\em Rules of encounter: designing conventions for automated
  negotiation among computers}.
\newblock MIT Press, Cambridge, MA, 1994.

\bibitem[RZ02]{resnick2002trust}
Paul Resnick and Richard Zeckhauser.
\newblock Trust among strangers in internet transactions: Empirical analysis of
  e{B}ay’s reputation system.
\newblock In Michael~R. Baye, editor, {\em The Economics of the Internet and
  E-commerce}, pages 127--158. Elvesier, 2002.

\bibitem[S{\etalchar{+}}08]{schotter2008economic}
Andrew Schotter et~al.
\newblock {\em The economic theory of social institutions}.
\newblock Cambridge University Press, 2008.
\newblock \href {http://dx.doi.org/10.1017/CBO9780511983863}
  {\path{doi:10.1017/CBO9780511983863}}.

\bibitem[S{\etalchar{+}}18]{sae2018taxonomy}
{SAE On-Road Automated Vehicle Standards Committee} et~al.
\newblock Taxonomy and definitions for terms related to driving automation
  systems for on-road motor vehicles, 2018.
\newblock URL: \url{https://www.sae.org/standards/content/j3016_201806/}.

\bibitem[San96]{sandholm1996limitations}
Tuomas~W Sandholm.
\newblock Limitations of the {V}ickrey auction in computational multiagent
  systems.
\newblock In {\em Proceedings of the Second International Conference on
  Multiagent Systems}, pages 299--306, 1996.

\bibitem[San03]{sandholm2003automated}
Tuomas Sandholm.
\newblock Automated mechanism design: A new application area for search
  algorithms.
\newblock In {\em International Conference on Principles and Practice of
  Constraint Programming}, pages 19--36, 2003.
\newblock \href {http://dx.doi.org/10.1007/978-3-540-45193-8_2}
  {\path{doi:10.1007/978-3-540-45193-8_2}}.

\bibitem[Sat75]{satterthwaite1975strategy}
Mark~Allen Satterthwaite.
\newblock Strategy-proofness and {A}rrow's conditions: Existence and
  correspondence theorems for voting procedures and social welfare functions.
\newblock {\em Journal of Economic Theory}, 10(2):187--217, 1975.
\newblock \href {http://dx.doi.org/10.1016/0022-0531(75)90050-2}
  {\path{doi:10.1016/0022-0531(75)90050-2}}.

\bibitem[SB18]{sutton2020reinforcement}
Richard~S Sutton and Andrew~G Barto.
\newblock {\em Reinforcement learning: An introduction}.
\newblock MIT Press, Cambridge, MA, second edition, 2018.

\bibitem[SBA{\etalchar{+}}03]{sierhuis2003human}
Maarten Sierhuis, Jeffrey~M Bradshaw, Alessandro Acquisti, Ron van Hoof, Renia
  Jeffers, and Andrzej Uszok.
\newblock Human-agent teamwork and adjustable autonomy in practice.
\newblock In {\em Proceedings of the Seventh International Symposium on
  Artificial Intelligence, Robotics and Automation in Space}, 2003.

\bibitem[SC17]{shirado2017locally}
Hirokazu Shirado and Nicholas~A Christakis.
\newblock Locally noisy autonomous agents improve global human coordination in
  network experiments.
\newblock {\em Nature}, 545:370--374, 2017.
\newblock \href {http://dx.doi.org/10.1038/nature22332}
  {\path{doi:10.1038/nature22332}}.

\bibitem[Sch71]{schelling1971dynamic}
Thomas~C Schelling.
\newblock Dynamic models of segregation.
\newblock {\em Journal of Mathematical Sociology}, 1(2):143--186, 1971.
\newblock \href {http://dx.doi.org/10.1080/0022250X.1971.9989794}
  {\path{doi:10.1080/0022250X.1971.9989794}}.

\bibitem[Sch80]{schelling1980strategy}
Thomas~C Schelling.
\newblock {\em The Strategy of Conflict}.
\newblock Harvard University Press, Cambridge, MA, 1980.

\bibitem[Sch99]{schaal1999imitation}
Stefan Schaal.
\newblock Is imitation learning the route to humanoid robots.
\newblock {\em Trends in Cognitive Sciences}, 3:232--242, 1999.
\newblock \href {http://dx.doi.org/10.1016/S1364-6613(99)01327-3}
  {\path{doi:10.1016/S1364-6613(99)01327-3}}.

\bibitem[Sea95]{searle1995construction}
John~R Searle.
\newblock {\em The construction of social reality}.
\newblock Simon and Schuster, New York, 1995.

\bibitem[Sej20]{sejnowski2020unreasonable}
Terrence~J Sejnowski.
\newblock The unreasonable effectiveness of deep learning in artificial
  intelligence.
\newblock {\em Proceedings of the National Academy of Sciences},
  117(48):30033--30038, 2020.
\newblock \href {http://dx.doi.org/10.1073/pnas.1907373117}
  {\path{doi:10.1073/pnas.1907373117}}.

\bibitem[Sen85]{sen1985goals}
Amartya Sen.
\newblock Goals, commitment, and identity.
\newblock {\em Journal of Law, Economics, and Organization}, 1:341, 1985.

\bibitem[Sen13]{sen2013comprehensive}
Sandip Sen.
\newblock A comprehensive approach to trust management.
\newblock In {\em Proceedings of the 2013 International Conference on
  Autonomous Agents and Multiagent Systems}, pages 797--800, 2013.
\newblock URL:
  \url{http://www.ifaamas.org/Proceedings/aamas2013/docs/p797.pdf}.

\bibitem[SFA{\etalchar{+}}20]{shah2020neurips}
Rohin Shah, Pedro Freire, Neel Alex, Rachel Freedman, Dmitrii Krasheninnikov,
  Lawrence Chan, Michael Dennis, Pieter Abbeel, Anca Dragan, and Stuart
  Russell.
\newblock Benefits of assistance over reward learning in cooperative {AI}.
\newblock In {\em NeurIPS Workshop on Cooperative AI}, 2020.

\bibitem[SGD19]{shevlane2019contact}
Toby Shevlane, Ben Garfinkel, and Allan Dafoe.
\newblock Contact tracing apps can help stop coronavirus. but they can hurt
  privacy.
\newblock {\em Washington Post}, April 2019.
\newblock URL:
  \url{https://www.washingtonpost.com/politics/2020/04/28/contact-tracing-apps-can-help-stop-coronavirus-they-can-hurt-privacy/}.

\bibitem[Sha53]{shapley1953value}
Lloyd~S Shapley.
\newblock A value for n-person games.
\newblock In H~W Kuhn and A~W Tucker, editors, {\em Contributions to the Theory
  of Games}, volume~2, pages 307--317. Princeton University Press, Princton,
  NJ, 1953.

\bibitem[SHCH14]{shadiev2014review}
Rustam Shadiev, Wu-Yuin Hwang, Nian-Shing Chen, and Yueh-Min Huang.
\newblock Review of speech-to-text recognition technology for enhancing
  learning.
\newblock {\em Journal of Educational Technology \& Society}, 17(4):65--84,
  2014.
\newblock URL: \url{https://www.jstor.org/stable/jeductechsoci.17.4.65}.

\bibitem[SHM{\etalchar{+}}16]{silver2016mastering}
David Silver, Aja Huang, Chris~J Maddison, Arthur Guez, Laurent Sifre, George
  van~den Driessche, Julian Schrittwieser, Ioannis Antonoglou, Veda
  Panneershelvam, Marc Lanctot, Sander Dieleman, Dominik Grewe, John Nham, Nal
  Kalchbrenner, Ilya Sutskever, Timothy Lillicrap, Madeleine Leach, Koray
  Kavukcuoglu, Thore Graepel, and Demis Hassabis.
\newblock Mastering the game of go with deep neural networks and tree search.
\newblock {\em Nature}, 529:484--489, 2016.
\newblock \href {http://dx.doi.org/10.1038/nature16961}
  {\path{doi:10.1038/nature16961}}.

\bibitem[SHS{\etalchar{+}}17]{silver2017mastering}
David Silver, Thomas Hubert, Julian Schrittwieser, Ioannis Antonoglou, Matthew
  Lai, Arthur Guez, Marc Lanctot, Laurent Sifre, Dharshan Kumaran, Thore
  Graepel, Timothy Lillicrap, Karen Simonyan, and Demis Hassabis.
\newblock Mastering chess and shogi by self-play with a general reinforcement
  learning algorithm.
\newblock preprint, 2017.
\newblock \href {http://arxiv.org/abs/1712.01815} {\path{arXiv:1712.01815}}.

\bibitem[SHS{\etalchar{+}}18]{silver2018general}
David Silver, Thomas Hubert, Julian Schrittwieser, Ioannis Antonoglou, Matthew
  Lai, Arthur Guez, Marc Lanctot, Laurent Sifre, Dharshan Kumaran, Thore
  Graepel, Timothy Lillicrap, Karen Simonyan, and Demis Hassabis.
\newblock A general reinforcement learning algorithm that masters chess, shogi,
  and go through self-play.
\newblock {\em Science}, 362(6419):1140--1144, 2018.
\newblock \href {http://dx.doi.org/10.1126/science.aar6404}
  {\path{doi:10.1126/science.aar6404}}.

\bibitem[SHZ{\etalchar{+}}18]{schmitt18}
Simon Schmitt, Jonathan~J Hudson, Augustin Zidek, Simon Osindero, Carl Doersch,
  Wojciech~M Czarnecki, Joel~Z Leibo, Heinrich Kuttler, Andrew Zisserman, Karen
  Simonyan, and S~M Ali~Eslami.
\newblock Kickstarting deep reinforcement learning.
\newblock preprint, 2018.
\newblock \href {http://arxiv.org/abs/1803.03835} {\path{arXiv:1803.03835}}.

\bibitem[Sid94]{sidner1994artificial}
Candace~L Sidner.
\newblock An artificial discourse language for collaborative negotiation.
\newblock In {\em Proceedings of the Twelfth AAAI National Conference on
  Artificial Intelligence}, pages 814--819, 1994.
\newblock URL: \url{https://www.aaai.org/Papers/AAAI/1994/AAAI94-124.pdf}.

\bibitem[Sin94]{singh2007multiagent}
Munindar~P Singh.
\newblock {\em Multiagent systems}.
\newblock Springer, Berlin, 1994.

\bibitem[SKWPT19]{serrino2019finding}
Jack Serrino, Max Kleiman-Weiner, David~C Parkes, and Josh Tenenbaum.
\newblock Finding friend and foe in multi-agent games.
\newblock In {\em Advances in Neural Information Processing Systems},
  volume~32, pages 1251--1261, 2019.
\newblock URL:
  \url{https://papers.nips.cc/paper/2019/file/912d2b1c7b2826caf99687388d2e8f7c-Paper.pdf}.

\bibitem[SL{\etalchar{+}}95]{sandholm1995issues}
Tuomas Sandholm, Victor~R Lesser, et~al.
\newblock Issues in automated negotiation and electronic commerce: Extending
  the contract net framework.
\newblock In {\em Proceedings of the First Innternational Conference on
  Multiagent Systems}, pages 328--335, 1995.
\newblock URL: \url{https://www.aaai.org/Papers/ICMAS/1995/ICMAS95-044.pdf}.

\bibitem[SLB09]{shoham2008multiagent}
Yoav Shoham and Kevin Leyton-Brown.
\newblock {\em Multiagent systems: Algorithmic, game-theoretic, and logical
  foundations}.
\newblock Cambridge University Press, New York, 2009.

\bibitem[Smi80]{smith1980contract}
Reid~G Smith.
\newblock The contract net protocol: High-level communication and control in a
  distributed problem solver.
\newblock {\em IEEE Transactions on Computers}, C-29(12):1104--1113, 1980.
\newblock \href {http://dx.doi.org/10.1109/TC.1980.1675516}
  {\path{doi:10.1109/TC.1980.1675516}}.

\bibitem[SMT{\etalchar{+}}05]{schurr2005future}
Nathan Schurr, Janusz Marecki, Milind Tambe, Paul Scerri, Nikhil Kasinadhuni,
  and John~P Lewis.
\newblock The future of disaster response: Humans working with multiagent teams
  using {DEFACTO}.
\newblock In {\em AAAI spring symposium: AI technologies for homeland
  security}, pages 9--16, 2005.
\newblock URL:
  \url{https://www.aaai.org/Papers/Symposia/Spring/2005/SS-05-01/SS05-01-002.pdf}.

\bibitem[Sni96]{snijders1996trust}
Chris Snijders.
\newblock {\em Trust and commitments}.
\newblock PhD thesis, University of Utrech, 1996.

\bibitem[SNP13]{sherchan2013survey}
Wanita Sherchan, Surya Nepal, and Cecile Paris.
\newblock A survey of trust in social networks.
\newblock {\em ACM Computing Surveys}, 45(4):1--33, 2013.
\newblock \href {http://dx.doi.org/10.1145/2501654.2501661}
  {\path{doi:10.1145/2501654.2501661}}.

\bibitem[SOW{\etalchar{+}}20]{stiennon2020learning}
Nisan Stiennon, Long Ouyang, Jeff Wu, Daniel~M Ziegler, Ryan Lowe, Chelsea
  Voss, Alec Radford, Dario Amodei, and Paul Christiano.
\newblock Learning to summarize from human feedback.
\newblock preprint, 2020.
\newblock \href {http://arxiv.org/abs/2009.01325} {\path{arXiv:2009.01325}}.

\bibitem[SP14]{scott2014speaking}
Thom Scott-Phillips.
\newblock {\em Speaking our minds: Why human communication is different, and
  how language evolved to make it special}.
\newblock Macmillan, New York, 2014.

\bibitem[SPAM{\etalchar{+}}19]{schwarting2019social}
Wilko Schwarting, Alyssa Pierson, Javier Alonso-Mora, Sertac Karaman, and
  Daniela Rus.
\newblock Social behavior for autonomous vehicles.
\newblock {\em Proceedings of the National Academy of Sciences},
  116(50):24972--24978, 2019.
\newblock \href {http://dx.doi.org/10.1073/pnas.1820676116}
  {\path{doi:10.1073/pnas.1820676116}}.

\bibitem[SPC{\etalchar{+}}16]{shneiderman2016designing}
Ben Shneiderman, Catherine Plaisant, Maxine Cohen, Steven Jacobs, Niklas
  Elmqvist, and Nicholas Diakopoulos.
\newblock {\em Designing the user interface: strategies for effective
  human-computer interaction}.
\newblock Pearson, Essex, 2016.

\bibitem[Spe73]{spense1973job}
Michael Spense.
\newblock Job market signaling.
\newblock {\em The Quarterly Journal of Economics}, 87(3):355--374, 1973.
\newblock \href {http://dx.doi.org/10.2307/1882010}
  {\path{doi:10.2307/1882010}}.

\bibitem[SPG07]{shoham2007if}
Yoav Shoham, Rob Powers, and Trond Grenager.
\newblock If multi-agent learning is the answer, what is the question?
\newblock {\em Artificial Intelligence}, 171(7):365--377, 2007.
\newblock \href {http://dx.doi.org/10.1016/j.artint.2006.02.006}
  {\path{doi:10.1016/j.artint.2006.02.006}}.

\bibitem[SS93]{smith1993origin}
John~Maynard Smith and E\"{o}rs Sz{\'a}thmary.
\newblock The origin of chromosomes {I}. {S}election for linkage.
\newblock {\em Journal of Theoretical Biology}, 164(4):437--446, 1993.
\newblock \href {http://dx.doi.org/10.1006/jtbi.1993.1165}
  {\path{doi:10.1006/jtbi.1993.1165}}.

\bibitem[SS97]{smith1997major}
John~Maynard Smith and E\"{o}rs Sz{\'a}thmary.
\newblock {\em The major transitions in evolution}.
\newblock Oxford University Press, Oxford, UK, 1997.

\bibitem[SS05]{sabater2005review}
Jordi Sabater and Carles Sierra.
\newblock Review on computational trust and reputation models.
\newblock {\em Artificial Intelligence Review}, 24:33--60, 2005.
\newblock \href {http://dx.doi.org/10.1007/s10462-004-0041-5}
  {\path{doi:10.1007/s10462-004-0041-5}}.

\bibitem[SSF16]{sukhbaatar2016learning}
Sainbayar Sukhbaatar, Arthur Szlam, and Rob Fergus.
\newblock Learning multiagent communication with backpropagation.
\newblock In {\em Advances in Neural Information Processing Systems},
  volume~29, pages 2244--2252, 2016.
\newblock URL:
  \url{https://papers.nips.cc/paper/2016/file/55b1927fdafef39c48e5b73b5d61ea60-Paper.pdf}.

\bibitem[SSSD16]{sadigh2016information}
Dorsa Sadigh, S~Shankar Sastry, Sanjit~A Seshia, and Anca Dragan.
\newblock Information gathering actions over human internal state.
\newblock In {\em 2016 IEEE/RSJ International Conference on Intelligent Robots
  and Systems (IROS)}, pages 66--73. IEEE, 2016.
\newblock \href {http://dx.doi.org/10.1109/IROS.2016.7759036}
  {\path{doi:10.1109/IROS.2016.7759036}}.

\bibitem[ST92]{shoham1992synthesis}
Yoav Shoham and Moshe Tennenholtz.
\newblock On the synthesis of useful social laws for artificial agent societies
  (preliminary report).
\newblock In {\em Proceedings of the Tenth National Conference on Artificial
  Intelligence}, pages 276--281, 1992.
\newblock URL: \url{https://www.aaai.org/Papers/AAAI/1992/AAAI92-043.pdf}.

\bibitem[ST95]{shoham1995social}
Yoav Shoham and Moshe Tennenholtz.
\newblock On social laws for artificial agent societies: off-line design.
\newblock {\em Artificial Intelligence}, 73(1-2):231--252, 1995.
\newblock \href {http://dx.doi.org/10.1016/0004-3702(94)00007-N}
  {\path{doi:10.1016/0004-3702(94)00007-N}}.

\bibitem[ST97]{shoham1997emergence}
Yoav Shoham and Moshe Tennenholtz.
\newblock On the emergence of social conventions: modeling, analysis, and
  simulations.
\newblock {\em Artificial Intelligence}, 94(1-2):139--166, 1997.
\newblock \href {http://dx.doi.org/10.1016/S0004-3702(97)00028-3}
  {\path{doi:10.1016/S0004-3702(97)00028-3}}.

\bibitem[Sta02]{stalnaker2002common}
Robert Stalnaker.
\newblock Common ground.
\newblock {\em Linguistics and philosophy}, 25(5/6):701--721, 2002.
\newblock URL: \url{https://www.jstor.org/stable/25001871}.

\bibitem[SV00]{stone2000multiagent}
Peter Stone and Manuela Veloso.
\newblock Multiagent systems: A survey from a machine learning perspective.
\newblock {\em Autonomous Robots}, 8:345--383, 2000.
\newblock \href {http://dx.doi.org/10.1023/A:1008942012299}
  {\path{doi:10.1023/A:1008942012299}}.

\bibitem[Tam11]{tambe2011security}
Milind Tambe.
\newblock {\em Security and game theory: algorithms, deployed systems, lessons
  learned}.
\newblock Cambridge University Press, New York, 2011.

\bibitem[Tan17]{tang2017reinforcement}
Pingzhong Tang.
\newblock Reinforcement mechanism design.
\newblock In {\em Proceedings of the 26th International Joint Conference on
  Artificial Intelligence}, pages 5146--5150, 2017.
\newblock URL: \url{https://www.ijcai.org/Proceedings/2017/0739.pdf}.

\bibitem[TBG{\etalchar{+}}]{Traskstructuredtransparency}
Andrew Trask, Emma Bluemke, Ben Garfinkel, Claudia Ghezzou Cuervas-Mons, and
  Allan. Dafoe.
\newblock Beyond privacy tradeoffs with structured transparency.

\bibitem[Ten04]{tennenholtz2004program}
Moshe Tennenholtz.
\newblock Program equilibrium.
\newblock {\em Games and Economic Behavior}, 49(2):363--373, 2004.
\newblock \href {http://dx.doi.org/10.1016/j.geb.2004.02.002}
  {\path{doi:10.1016/j.geb.2004.02.002}}.

\bibitem[Tes94]{tesauro1994td}
Gerald Tesauro.
\newblock {TD-Gammon}, a self-teaching backgammon program, achieves
  master-level play.
\newblock {\em Neural Computation}, 6(2):215--219, 1994.
\newblock \href {http://dx.doi.org/10.1162/neco.1994.6.2.215}
  {\path{doi:10.1162/neco.1994.6.2.215}}.

\bibitem[THHF18]{tung2018reward}
Hsiao-Yu~Fish Tung, Adam~W Harley, Liang-Kang Huang, and Katerina Fragkiadaki.
\newblock Reward learning from narrated demonstrations.
\newblock In {\em Proceedings of the IEEE Conference on Computer Vision and
  Pattern Recognition}, pages 7004--7013, 2018.
\newblock \href {http://dx.doi.org/10.1109/CVPR.2018.00732}
  {\path{doi:10.1109/CVPR.2018.00732}}.

\bibitem[TK74]{tversky1974judgment}
Amos Tversky and Daniel Kahneman.
\newblock Judgment under uncertainty: Heuristics and biases.
\newblock {\em Science}, 185(4157):1124--1131, 1974.
\newblock \href {http://dx.doi.org/10.1126/science.185.4157.1124}
  {\path{doi:10.1126/science.185.4157.1124}}.

\bibitem[TKNF00]{takeuchi2000cultural}
Yugo Takeuchi, Yasuhiro Katagiri, Clifford Nass, and BJ~Fogg.
\newblock A cultural perspective in social interface.
\newblock In {\em Proceedings of the ACM CHI 2000 Human Factors in Computing
  Systems Conference}, 2000.

\bibitem[Tom09]{tomasello2009we}
Michael Tomasello.
\newblock {\em Why we cooperate}.
\newblock MIT Press, Cambridge, MA, 2009.

\bibitem[Tom12]{tomz2012reputation}
Michael Tomz.
\newblock {\em Reputation and international cooperation: Sovereign debt across
  three centuries}.
\newblock Princeton University Press, Princeton, NJ, 2012.

\bibitem[TPJL06]{teacy2006travos}
WT~Luke Teacy, Jigar Patel, Nicholas~R Jennings, and Michael Luck.
\newblock Travos: Trust and reputation in the context of inaccurate information
  sources.
\newblock {\em Autonomous Agents and Multi-Agent Systems}, 12:183--198, 2006.
\newblock URL: \url{10.1007/s10458-006-5952-x}.

\bibitem[TR10]{thornton2010}
Alex Thornton and Nichola Raihani.
\newblock Identifying teaching in wild animals.
\newblock {\em Learning \& Behavior}, 38:297--309, 08 2010.
\newblock \href {http://dx.doi.org/10.3758/LB.38.3.297}
  {\path{doi:10.3758/LB.38.3.297}}.

\bibitem[TS09]{taylor2009transfer}
Matthew~E Taylor and Peter Stone.
\newblock Transfer learning for reinforcement learning domains: A survey.
\newblock {\em Journal of Machine Learning Research}, 10(56):1633--1685, 2009.
\newblock URL: \url{https://www.jmlr.org/papers/v10/taylor09a.html}.

\bibitem[TSG{\etalchar{+}}19]{tacchetti2019neural}
Andrea Tacchetti, DJ~Strouse, Marta Garnelo, Thore Graepel, and Yoram Bachrach.
\newblock A neural architecture for designing truthful and efficient auctions.
\newblock preprint, 2019.
\newblock \href {http://arxiv.org/abs/1907.05181} {\path{arXiv:1907.05181}}.

\bibitem[TSJ{\etalchar{+}}20]{traeger2020vulnerable}
Margaret~L Traeger, Sarah~Strohkorb Sebo, Malte Jung, Brian Scassellati, and
  Nicholas~A Christakis.
\newblock Vulnerable robots positively shape human conversational dynamics in a
  human--robot team.
\newblock {\em Proceedings of the National Academy of Sciences},
  117(12):6370--6375, 2020.
\newblock \href {http://dx.doi.org/10.1073/pnas.1910402117}
  {\path{doi:10.1073/pnas.1910402117}}.

\bibitem[TVS07]{tanenbaum2007distributed}
Andrew~S Tanenbaum and Maarten Van~Steen.
\newblock {\em Distributed systems: principles and paradigms}.
\newblock Prentice-Hall, Upper Saddle River, NJ, 2007.

\bibitem[VBC{\etalchar{+}}19]{vinyals2019grandmaster}
Oriol Vinyals, Igor Babuschkin, Wojciech~M Czarnecki, Micha{\"e}l Mathieu,
  Andrew Dudzik, Junyoung Chung, David~H Choi, Richard Powell, Timo Ewalds,
  Petko Georgiev, et~al.
\newblock Grandmaster level in {StarCraft II} using multi-agent reinforcement
  learning.
\newblock {\em Nature}, 575(7782):350--354, 2019.
\newblock \href {http://dx.doi.org/10.1038/s41586-019-1724-z}
  {\path{doi:10.1038/s41586-019-1724-z}}.

\bibitem[vdHW02]{van2002tractable}
Wiebe van~der Hoek and Michael Wooldridge.
\newblock Tractable multiagent planning for epistemic goals.
\newblock In {\em Proceedings of the First International Joint Conference on
  Autonomous Agents and Multiagent Systems: part 3}, pages 1167--1174, 2002.
\newblock \href {http://dx.doi.org/10.1145/545056.545095}
  {\path{doi:10.1145/545056.545095}}.

\bibitem[vHLP08]{van2008handbook}
Frank van Harmelen, Vladimir Lifschitz, and Bruce Porter, editors.
\newblock {\em Handbook of knowledge representation}.
\newblock Elsevier, Amsterdam, 2008.

\bibitem[Vic61]{vickrey1961counterspeculation}
William Vickrey.
\newblock Counterspeculation, auctions, and competitive sealed tenders.
\newblock {\em The Journal of Finance}, 16(1):8--37, 1961.
\newblock \href {http://dx.doi.org/10.2307/2977633}
  {\path{doi:10.2307/2977633}}.

\bibitem[VLRY14]{van2014reward}
Paul A~M Van~Lange, Bettina Rockenbach, and Toshio Yamagishi, editors.
\newblock {\em Reward and punishment in social dilemmas}.
\newblock Oxford University Press, New York, 2014.

\bibitem[vNM07]{von2007theory}
John von Neumann and Oskar Morgenstern.
\newblock {\em Theory of games and economic behavior (commemorative edition)}.
\newblock Princeton University Press, Princeton, NJ, 2007.

\bibitem[VSDD05]{vazquez2005organizing}
Javier V{\'a}zquez-Salceda, Virginia Dignum, and Frank Dignum.
\newblock Organizing multiagent systems.
\newblock {\em Autonomous Agents and Multi-Agent Systems}, 11(3):307--360,
  2005.
\newblock \href {http://dx.doi.org/10.1007/s10458-005-1673-9}
  {\path{doi:10.1007/s10458-005-1673-9}}.

\bibitem[VSMS13]{villatoro2013robust}
Daniel Villatoro, Jordi Sabater-Mir, and Sandip Sen.
\newblock Robust convention emergence in social networks through
  self-reinforcing structures dissolution.
\newblock {\em ACM Transactions on Autonomous and Adaptive Systems},
  8(1):1--21, 2013.
\newblock \href {http://dx.doi.org/10.1145/2451248.2451250}
  {\path{doi:10.1145/2451248.2451250}}.

\bibitem[VSP{\etalchar{+}}17]{vaswani2017attention}
Ashish Vaswani, Noam Shazeer, Niki Parmar, Jakob Uszkoreit, Llion Jones,
  Aidan~N. Gomez, Lukasz Kaiser, and Illia Polosukhin.
\newblock Attention is all you need.
\newblock preprint, 2017.
\newblock \href {http://arxiv.org/abs/1706.03762} {\path{arXiv:1706.03762}}.

\bibitem[War18]{warneken2018children}
Felix Warneken.
\newblock How children solve the two challenges of cooperation.
\newblock {\em Annual Review of Psychology}, 69:205--229, 2018.
\newblock \href {http://dx.doi.org/10.1146/annurev-psych-122216-011813}
  {\path{doi:10.1146/annurev-psych-122216-011813}}.

\bibitem[WCL97]{williams1997origins}
John~T Williams, Brian Collins, and Mark~I Lichbach.
\newblock The origins of credible commitment to the market.
\newblock In {\em Annual Meeting of the American Political Science
  Association}, 1997.

\bibitem[Web20]{weber20202020s}
Steven Weber.
\newblock The 2020s political economy of machine translation.
\newblock preprint, 2020.
\newblock \href {http://arxiv.org/abs/2011.01007} {\path{arXiv:2011.01007}}.

\bibitem[Wei83]{10.1145/357980.357991}
Joseph Weizenbaum.
\newblock Eliza — a computer program for the study of natural language
  communication between man and machine.
\newblock {\em Communications of the ACM}, 26(1):23–28, January 1983.
\newblock \href {http://dx.doi.org/10.1145/357980.357991}
  {\path{doi:10.1145/357980.357991}}.

\bibitem[Wei95]{weiss1995adaptation}
Gerhard Wei{\ss}.
\newblock Adaptation and learning in multi-agent systems: Some remarks and a
  bibliography.
\newblock In {\em International Joint Conference on Artificial Intelligence},
  pages 1--21. Springer, 1995.
\newblock \href {http://dx.doi.org/10.1007/3-540-60923-7_16}
  {\path{doi:10.1007/3-540-60923-7_16}}.

\bibitem[Wei99]{weiss1999multiagent}
Gerhard Weiss, editor.
\newblock {\em Multiagent systems: a modern approach to distributed artificial
  intelligence}.
\newblock MIT Press, Cambridge, MA, 1999.

\bibitem[Wel93]{wellman1993market}
Michael~P Wellman.
\newblock A market-oriented programming environment and its application to
  distributed multicommodity flow problems.
\newblock {\em Journal of artificial intelligence research}, 1:1--23, 1993.
\newblock \href {http://dx.doi.org/10.1613/jair.2} {\path{doi:10.1613/jair.2}}.

\bibitem[WFH20]{woodward2020learning}
Mark Woodward, Chelsea Finn, and Karol Hausman.
\newblock Learning to interactively learn and assist.
\newblock In {\em Proceedings of the AAAI Conference on Artificial
  Intelligence}, pages 2535--2543, 2020.
\newblock \href {http://dx.doi.org/10.1609/aaai.v34i03.5636}
  {\path{doi:10.1609/aaai.v34i03.5636}}.

\bibitem[WGSW03]{wellman20032001}
Michael~P Wellman, Amy Greenwald, Peter Stone, and Peter~R Wurman.
\newblock The 2001 trading agent competition.
\newblock {\em Electronic Markets}, 13(1):4--12, 2003.
\newblock \href {http://dx.doi.org/10.1080/1019678032000062212}
  {\path{doi:10.1080/1019678032000062212}}.

\bibitem[Wie05]{wiessner2005norm}
Polly Wiessner.
\newblock Norm enforcement among the {J}u/’hoansi bushmen.
\newblock {\em Human Nature}, 16:115--145, 2005.
\newblock \href {http://dx.doi.org/10.1007/s12110-005-1000-9}
  {\path{doi:10.1007/s12110-005-1000-9}}.

\bibitem[Woo09]{wooldridge2009introduction}
Michael Wooldridge.
\newblock {\em An introduction to multiagent systems}.
\newblock John Wiley \& Sons, West Sussex, 2009.

\bibitem[WP00]{wooldridge2000languages}
Michael Wooldridge and Simon Parsons.
\newblock Languages for negotiation.
\newblock In {\em 14th European Conference on Artificial Intelligence}, pages
  393--397, 2000.
\newblock URL: \url{http://www.frontiersinai.com/ecai/ecai2000/p0393.html}.

\bibitem[WPN14]{werfel2014designing}
Justin Werfel, Kirstin Petersen, and Radhika Nagpal.
\newblock Designing collective behavior in a termite-inspired robot
  construction team.
\newblock {\em Science}, 343(6172):754--758, 2014.
\newblock \href {http://dx.doi.org/10.1126/science.1245842}
  {\path{doi:10.1126/science.1245842}}.

\bibitem[WRUW03]{wagner2003progress}
Kyle Wagner, James~A Reggia, Juan Uriagereka, and Gerald~S Wilkinson.
\newblock Progress in the simulation of emergent communication and language.
\newblock {\em Adaptive Behavior}, 11(1):37--69, 2003.
\newblock \href {http://dx.doi.org/10.1177/10597123030111003}
  {\path{doi:10.1177/10597123030111003}}.

\bibitem[WS07]{wang2007formal}
Yonghong Wang and Munindar~P Singh.
\newblock Formal trust model for multiagent systems.
\newblock In {\em Proceedings of the 20th International Joint Conference on
  Artifical Intelligence}, pages 1551--1556, 2007.
\newblock URL: \url{https://www.aaai.org/Papers/IJCAI/2007/IJCAI07-250.pdf}.

\bibitem[WZ18]{wohrer2018design}
Maximilian W{\"o}hrer and Uwe Zdun.
\newblock Design patterns for smart contracts in the ethereum ecosystem.
\newblock In {\em IEEE International Conference on Internet of Things (iThings)
  and IEEE Green Computing and Communications (GreenCom) and IEEE Cyber,
  Physical and Social Computing (CPSCom) and IEEE Smart Data (SmartData)},
  pages 1513--1520. IEEE, 2018.
\newblock \href {http://dx.doi.org/10.1109/Cybermatics_2018.2018.00255}
  {\path{doi:10.1109/Cybermatics_2018.2018.00255}}.

\bibitem[YCZ{\etalchar{+}}20]{ye2020towards}
Deheng Ye, Guibin Chen, Wen Zhang, Sheng Chen, Bo~Yuan, Bo~Liu, Jia Chen, Zhao
  Liu, Fuhao Qiu, Hongsheng Yu, Yinyuting Yin, Bei Shi, Liang Wang, Tengfei
  Shi, Qiang Fu, Wei Yang, Lanxiao Huang, and Wei Liu.
\newblock Towards playing full {MOBA} games with deep reinforcement learning.
\newblock {\em Advances in Neural Information Processing Systems}, 33, 2020.
\newblock URL:
  \url{https://papers.nips.cc/paper/2020/file/06d5ae105ea1bea4d800bc96491876e9-Paper.pdf}.

\bibitem[YJBK17]{Yim2017Gift}
Junho Yim, Donggyu Joo, Jihoon Bae, and Junmo Kim.
\newblock A gift from knowledge distillation: Fast optimization, network
  minimization and transfer learning.
\newblock In {\em Proceedings of the IEEE Conference on Computer Vision and
  Pattern Recognition}, pages 7130--7138, 2017.
\newblock \href {http://dx.doi.org/10.1109/CVPR.2017.754}
  {\path{doi:10.1109/CVPR.2017.754}}.

\bibitem[YKL{\etalchar{+}}07]{yan2007autonomous}
Jun Yan, Ryszard Kowalczyk, Jian Lin, Mohan~B Chhetri, Suk~Keong Goh, and
  Jianying Zhang.
\newblock Autonomous service level agreement negotiation for service
  composition provision.
\newblock {\em Future Generation Computer Systems}, 23(6):748--759, 2007.
\newblock \href {http://dx.doi.org/10.1016/j.future.2007.02.004}
  {\path{doi:10.1016/j.future.2007.02.004}}.

\bibitem[YLF{\etalchar{+}}20]{yang2020learning}
Jiachen Yang, Ang Li, Mehrdad Farajtabar, Peter Sunehag, Edward Hughes, and
  Hongyuan Zha.
\newblock Learning to incentivize other learning agents.
\newblock preprint, 2020.
\newblock \href {http://arxiv.org/abs/2006.06051} {\path{arXiv:2006.06051}}.

\bibitem[yLLd06]{y2006normative}
Fabiola~L{\'o}pez y~L{\'o}pez, Michael Luck, and Mark d’Inverno.
\newblock A normative framework for agent-based systems.
\newblock {\em Computational \& Mathematical Organization Theory}, 12:227--250,
  2006.
\newblock \href {http://dx.doi.org/10.1007/s10588-006-9545-7}
  {\path{doi:10.1007/s10588-006-9545-7}}.

\bibitem[You96]{young1996economics}
H~Peyton Young.
\newblock The economics of convention.
\newblock {\em Journal of Economic Perspectives}, 10(2):105--122, 1996.
\newblock \href {http://dx.doi.org/10.1257/jep.10.2.105}
  {\path{doi:10.1257/jep.10.2.105}}.

\bibitem[YSL{\etalchar{+}}13]{yu2013survey}
Han Yu, Zhiqi Shen, Cyril Leung, Chunyan Miao, and Victor~R Lesser.
\newblock A survey of multi-agent trust management systems.
\newblock {\em IEEE Access}, 1:35--50, 2013.
\newblock \href {http://dx.doi.org/10.1109/ACCESS.2013.2259892}
  {\path{doi:10.1109/ACCESS.2013.2259892}}.

\bibitem[YZRL13]{yu2013}
Chao Yu, Minjie Zhang, Fenghui Ren, and Xudong Luo.
\newblock Emergence of social norms through collective learning in networked
  agent societies.
\newblock In {\em 12th International Conference on Autonomous Agents and
  Multiagent Systems}, pages 475--482, 2013.
\newblock URL:
  \url{http://www.ifaamas.org/Proceedings/aamas2013/docs/p475.pdf}.

\bibitem[Zah77]{zahavi1977reliability}
Amotz Zahavi.
\newblock Reliability in communication systems and the evolution of altruism.
\newblock In Bernard Stonehouse and Christopher Perrins, editors, {\em
  Evolutionary Ecology}, pages 253--259. MacMillan, London, 1977.

\bibitem[ZJBP07]{zinkevich2008regret}
Martin Zinkevich, Michael Johanson, Michael Bowling, and Carmelo Piccione.
\newblock Regret minimization in games with incomplete information.
\newblock In {\em Advances in Neural Information Processing Systems},
  volume~20, pages 1729--1736, 2007.
\newblock URL:
  \url{https://papers.nips.cc/paper/2007/file/08d98638c6fcd194a4b1e6992063e944-Paper.pdf}.

\bibitem[ZL10]{zhang2010multi}
Chongjie Zhang and Victor~R Lesser.
\newblock Multi-agent learning with policy prediction.
\newblock In {\em Proceedings of the Twenty-Fourth AAAI Conference on
  Artificial Intelligence}, pages 927--934, 2010.
\newblock URL:
  \url{https://www.aaai.org/ocs/index.php/AAAI/AAAI10/paper/view/1885}.

\bibitem[ZM00]{zacharia2000trust}
Giorgos Zacharia and Pattie Maes.
\newblock Trust management through reputation mechanisms.
\newblock {\em Applied Artificial Intelligence}, 14(9):881--907, 2000.
\newblock \href {http://dx.doi.org/10.1080/08839510050144868}
  {\path{doi:10.1080/08839510050144868}}.

\bibitem[ZPR09]{zuckerman2009algorithms}
Michael Zuckerman, Ariel~D Procaccia, and Jeffrey~S Rosenschein.
\newblock Algorithms for the coalitional manipulation problem.
\newblock {\em Artificial Intelligence}, 173(2):392--412, 2009.
\newblock \href {http://dx.doi.org/10.1016/j.artint.2008.11.005}
  {\path{doi:10.1016/j.artint.2008.11.005}}.

\bibitem[ZTS{\etalchar{+}}20]{zheng2020ai}
Stephan Zheng, Alexander Trott, Sunil Srinivasa, Nikhil Naik, Melvin Gruesbeck,
  David~C Parkes, and Richard Socher.
\newblock The {AI} economist: Improving equality and productivity with
  {AI}-driven tax policies.
\newblock preprint, 2020.
\newblock \href {http://arxiv.org/abs/2004.13332} {\path{arXiv:2004.13332}}.

\bibitem[ZVW14]{zimmer2014teacher}
Matthieu Zimmer, Paolo Viappiani, and Paul Weng.
\newblock Teacher-student framework: a reinforcement learning approach, 2014.
\newblock presentation at AAMAS Workshop Autonomous Robots and Multirobot
  Systems.
\newblock URL:
  \url{https://matthieu-zimmer.net/publications/ARMS2014_slides.pdf}.

\bibitem[ZWZ19]{Zhang2019HIBERT}
Xingxing Zhang, Furu Wei, and Ming Zhou.
\newblock {HIBERT}: Document level pre-training of hierarchical bidirectional
  transformers for document summarization.
\newblock In {\em Proceedings of the 57th Annual Meeting of the Association for
  Computational Linguistics}, pages 5059--5069, 2019.
\newblock URL: \url{https://www.aclweb.org/anthology/P19-1499}.

\bibitem[ZYW{\etalchar{+}}19]{zhang2019survey}
Xiang Zhang, Lina Yao, Xianzhi Wang, Jessica Monaghan, David Mcalpine, and
  Yu~Zhang.
\newblock A survey on deep learning based brain computer interface: Recent
  advances and new frontiers.
\newblock preprint, 2019.
\newblock \href {http://arxiv.org/abs/1905.04149} {\path{arXiv:1905.04149}}.

\end{thebibliography}
\bibliographystyle{alphaurl}

\end{document}